\documentclass[10pt]{article} % For LaTeX2e
\usepackage[preprint]{tmlr}   % preprint

\usepackage[utf8]{inputenc} % allow utf-8 input
\usepackage[T1]{fontenc}    % use 8-bit T1 fonts
\usepackage{booktabs}       % professional-quality tables
\usepackage{amsfonts}       % blackboard math symbols
\usepackage{nicefrac}       % compact symbols for 1/2, etc.
\usepackage{microtype}      % microtypography
\usepackage{xcolor}         % colors

\usepackage{subcaption}
\usepackage{multicol}
\usepackage{multirow}
\usepackage{amsmath}       % align
\usepackage{cancel}        % cancelto
\usepackage{xspace}        % \ie, \eg
\usepackage{soul}          % highlight
\usepackage{bbm}           % \mathbbm{1}
\usepackage{graphicx}
\usepackage{tikz}
\usepackage[boxed]{algorithm2e}

\newcommand{\change}[1]{\textcolor{blue}{#1}}
\definecolor{myblue}{rgb}{0.0196, 0.5922, 1}

\definecolor{mygreen}{rgb}{0.3961, 0.8196, 0}

\newcommand\etc{\textit{etc.}\xspace}
\newcommand\ie{\textit{i.e.}\xspace}
\newcommand\eg{\textit{e.g.}\xspace}

\newcommand{\indep}{\perp\!\!\!\perp}
\renewcommand{\mid}{\,\middle|\,}

\DeclareMathSymbol{\shortminus}{\mathbin}{AMSa}{"39}
\newcommand{\bracketlowersymb}[2]{[#1]_{\raisebox{-1.75pt}{$\scriptstyle #2$}}}
\newcommand{\pospart}[1]{\bracketlowersymb{#1}{+}}
\newcommand{\negpart}[1]{\bracketlowersymb{#1}{-}}

\newcommand{\pnot}{\mathbin{\stackrel{{\textcolor{gray}{\scriptscriptstyle \textrm{P}}}}{\lnot}}}
\newcommand{\pand}{\mathbin{\stackrel{{\textcolor{gray}{\scriptscriptstyle \textrm{P}}}}{\land}}}
\newcommand{\por}{\mathbin{\stackrel{{\textcolor{gray}{\scriptscriptstyle \textrm{P}}}}{\lor}}}

\usepackage{amsmath,amsfonts,bm}

\def\eqref#1{equation~\ref{#1}}
\def\1{\bm{1}}

\DeclareMathAlphabet{\mathsfit}{\encodingdefault}{\sfdefault}{m}{sl}
\SetMathAlphabet{\mathsfit}{bold}{\encodingdefault}{\sfdefault}{bx}{n}

\usepackage{hyperref}
\usepackage{url}

\hypersetup{
  colorlinks=true,
  linkcolor=black,
  citecolor=black,     
  urlcolor=black,
}

\title{Neural Logic Networks for Interpretable Classification}

\author{\name Vincent Perreault \email vincent.perreault@polymtl.ca \\
      \addr Department of Mathematics and Industrial Engineering\\
      Polytechnique Montréal
      \AND
      \name Katsumi Inoue \email inoue@nii.ac.jp \\
      \addr National Institute of Informatics, Tokyo
      \AND
      \name Richard Labib \email richard.labib@polymtl.ca\\
      \addr Department of Mathematics and Industrial Engineering\\
      Polytechnique Montréal
      \AND
      \name Alain Hertz \email alain.hertz@gerad.ca\\
      \addr Department of Mathematics and Industrial Engineering\\
      Polytechnique Montréal}

\def\month{06}  % Insert correct month for camera-ready version
\def\year{2026} % Insert correct year for camera-ready version
\def\openreview{\url{https://openreview.net/forum?id=CmJXxeimUk}} % Insert correct link to OpenReview for camera-ready version

\begin{document}

\maketitle

\begin{abstract}
Traditional neural networks have an impressive classification performance, but what they learn cannot be inspected, verified or extracted. Neural Logic Networks on the other hand have an interpretable structure that enables them to learn a logical mechanism relating the inputs and outputs with AND and OR operations. We generalize these networks with NOT operations and biases that take into account unobserved data and develop a rigorous logical and probabilistic modeling in terms of concept combinations to motivate their use. We also propose a novel factorized IF-THEN rule structure for the model as well as a modified learning algorithm. Our method improves the state-of-the-art in Boolean networks discovery and is able to learn relevant, interpretable rules in tabular classification, notably on examples from the medical and industrial fields where interpretability has tangible value.
\end{abstract}

\section{Introduction}

Neural networks have revolutionized Machine Learning (ML) with unprecedented performance in perception tasks, ranging from prediction of complex phenomena to recognition and generation of images, sound, speech and text. However, this impressive performance is accompanied by a lack of explainability of how it is achieved, with neural networks being treated as black-box models due to the opaque nature of their learned parameters. As a result, it has often been claimed that the information that a neural network has learned cannot be inspected, verified or extracted.

As these black box models are increasingly being used to support or automate decision making, transparency has become a critical concern, giving rise to the field of eXplainable Artificial Intelligence (XAI) \citep{Arrieta20, Calegari20}. This is especially important for domains where ethics plays a pivotal role such as medecine, transportation, legal, finance, military \citep{Adadi18} and scientific discovery. In those contexts, a prediction or decision is only useful when it is accompanied by an explanation of how it was obtained, as well as by an assurance that is is not the result of unacceptable biases, such as gender or ethnicity for example.

In parallel with this increasing demand for transparency, research on \emph{neuro-symbolic} methods has become more popular. This family of methods aims to combine neural (also known as connectionist or sub-symbolic) with symbolic (logical) techniques to obtain a best-of-both-world scenario, with the complementary strengths of both paradigms. Neural methods are best at perception tasks and use continuous vector representations to learn a distributed representation from raw data, making them fast, strong at handling unstructured data and robust to noise and errors in the data \citep{Yu23}. On the other hand, symbolic methods are best at cognition tasks and use discrete logical representations to reason deductively about knowledge, making them provably correct, human-intelligible, and with strong generalization ability. Neuro-symbolic methods include many different approaches to unify these two paradigms: neural implementations of logic, logical characterizations of neural systems, and hybrid systems that combine both in more or less equal measures \citep{Besold21}.

One branch of hybrid neuro-symbolic methods defines new types of neural networks where the neurons represent logical AND and OR combinations, as opposed to the linear combination with non-linear activation of the classical perceptron. Such neural AND/OR networks aim to learn a logical mechanism relating the inputs and outputs that involve only AND, OR and NOT operations, resulting in a transparent and interpretable model. Ironically, the very inception of classical neural networks was justified by their ability to model such AND/OR networks \citep{McCulloch43}.

\subsection{Related work}

Neural networks with neurons that explicitly represent AND/OR operations were first defined by \cite{Pedrycz93, Hirota94}. Their learnable AND/OR nodes were defined in fuzzy logic for general t-norms (fuzzy AND) and t-conorms (fuzzy OR), while we adopt the product t-norm and t-conorm which admit a probabilistic interpretation. Moreover, their definition of the AND node used weights in a counterintuitive manner, having the opposite behavior of what we now understand as weights. In their formalism, a weight of 0 meant that its associated input was included in the AND operation while a weight of 1 meant that it was not. %, \ie whether an input is included in the AND operation.
The OR node with general t-norms and t-conorms was also independently defined by \cite{Gupta93, Gupta94}, which included an additional activation function. At the time, the idea of defining new neurons with fuzzy logic to be used in neural networks was an active field, referred to as Hybrid Fuzzy Neural Networks by \cite{Buckley98}. With a pre-processing layer to define fuzzy sets $S_{i,k}$ for each input $x_i$, followed by an AND layer and a final OR layer, a Neuro-Fuzzy classifier \citep{Fuller95} could be developed that would learn IF-THEN rules for each class $Y_k$ like
\begin{align*}
    \textbf{IF }x_1 \in S_{1,k}\text{ AND }\,...\,\text{ AND }x_d \in S_{d,k}\textbf{, THEN }x \in Y_k.
\end{align*}
Furthermore, by working with explicit AND/OR nodes instead of perceptrons, problem-specific expert knowledge can be pre-encoded into the network before the learning to be refined and finally extracted from the learned network. This idea was first tried with normal perceptrons in Knowledge-Based Artificial Neural Networks \citep{Towell94}. The initial knowledge was successfully encoded into the network, but it could not be re-extracted from the network after learning, because of its distributed representation.

AND/OR neural networks with product fuzzy logic were rediscovered in \cite{Payani19, Payani20} under the name Neural Logic Networks (NLN), which we adopt as well. Their formulation was obtained from an expected desiderata of what the weight values should mean (1: included in the AND/OR; 0: not included) rather than a rigorous logical or probabilistic modeling. The same definitions were then reused by \cite{Zhang23} and developed further by \cite{Wang20, Wang21, Wang24} into the Rule-based Representation Learner (RRL). To combat their notorious vanishing gradient \citep{Krieken22}, the RRL uses a new method called gradient grafting to learn the weights, along with approximated definitions of the AND/OR nodes which introduced new hyper-parameters to further improve learning. The RRL also introduced learnable upper and lower bounds to define the pre-processing sets $S_{i,k}$ in the case of continuous inputs $x_i$. Compared to our approach, these AND/OR nodes are limited in three significant ways. Firstly, they cannot consider negated inputs, \ie using the contrary opposite of an input (\eg using $\texttt{is\_off} = \texttt{is\_on}^c$), without doubling the number of weights, whereas we use a single weight to model both cases. Secondly, their modeling assumes that all the relevant data is observed and given in the input features, while our formulation takes into account the impact of other unobserved but relevant data. Thirdly, their approaches relied on data pre-processing techniques to deal with missing values in the observed input features, while our approach has a natural extension for missing values.

Other attempts at AND/OR neural networks were also created by others. \cite{Cingillioglu22} constrained the bias of regular perceptrons to obtain either an AND node or an OR node, depending on a hyper-parameter that is tuned during learning. Like us, their approach also considers both inputs and their contrary with a single weight. However, the magnitude of their weights do not explicitly represent their relative importance, unlike our weights which directly represent probabilities. Moreover, their formulation also fails to take into account unobserved data. \cite{Sato23} cleverly uses a ReLU network with constrained biases to learn a 2-layer AND/OR network, but their approach only works for perfect binary data with no noise or errors. Another type of model that produces similar IF-THEN rules is decision trees, and their generalization decision diagrams \citep{Florio23}. These models are represented by rooted directed graphs in which every node splits the possible values of one or more input $x_i$ in two or more branches, thus dividing up the input space into discrete bins that belong to the same class $Y_k$.

Due to their probabilistic formalization, NLNs also serve as probabilistic models of the target classes/labels when conditioning on the input features. In doing so, they implicitly learn Probabilistic Graphical Models (PGM), which encode the conditional structure of random variables in graph form (see appendix \ref{sec:app_theory_prob_PGM} for the PGM behind NLN's probabilistic modeling). However, NLNs require an additional approximation to ensure tractability in practical settings. With additional assumptions regarding the direction of causality, NLNs can also be viewed as a special form of structural causal model \citep{Peters17} in which the assignment functions have a clear interpretation as AND/OR combinations of binary random variables, some of which may be unobserved. However these stronger causality assumptions are not required to use NLNs, and would only be necessary to produce interventional distributions or counterfactuals. NLNs are also related to probabilistic circuits \citep{Choi20} which study the tractability of probabilistic queries in sum-product networks via structural constraints. In particular, logistic circuits \citep{Liang19} are probabilistic circuits with strong structural constraints that combine structure learning and logistic regression to learn AND/OR networks for interpretable classification. However, unlike NLNs, their final learned AND/OR networks contain non-discretized weights which are not as easily interpretable.

Finally, other approaches with similar names have also been proposed, in reasoning rather than learning tasks. The first method to be named Neural Logic Network was proposed by \cite{Teh95} and reintroduced recently by \cite{Ding18}. It defines neural networks with values $(t,f) \in [0,1]^2$ representing three-valued truth-values (true, false, unknown), with two sets of weights per neuron. Although some arrangements of weights give rise to interpretable AND and OR combinations, a restricted form of learning is required to maintain this interpretability, not making use of the current powerful gradient-based methods. Another approach, called Logical Neural Networks \citep{Riegel20, Sen22}, encodes any first-order logic program template in a neural network. By learning this neural network from a relational database, it aims to discover the best first-order logic rule (on relations) of the form given by the template. In comparison, our approach, by learning from a tabular data set, aims to discover the best propositional logic rule (on features) of any form.

\subsection{Structure and contributions}

To our knowledge, for the first time,
\begin{itemize}
    \item we present a theoretical formalism for the logical and probabilistic modeling behind NLNs and their AND/OR nodes (section \ref{sec:theory});
    
    \item we introduce biases accounting for unobserved data and we generalize the weights to be positive for using a concept and negative for using its contrary (section \ref{sec:theory});

    \item we develop an interpretable structure for NLNs with factorized IF-THEN rule modules and appropriate input pre-processing for binary, categorical and continuous features (section \ref{sec:ML_int_struct});

    \item we propose a modified learning algorithm for NLNs with a rule reset scheme at every epoch and a post-learning simplification  of the model to increase its interpretability (section \ref{sec:ML_learning}); and

    \item we test our method on two classification tasks: Boolean network discovery and interpretable tabular classification with additional examples in medicine and industry (section \ref{sec:exp}).
\end{itemize}

\begin{figure}[t]
    \centering
     \begin{subfigure}[b]{0.45\textwidth}
         \centering
         \includegraphics[width=\textwidth]{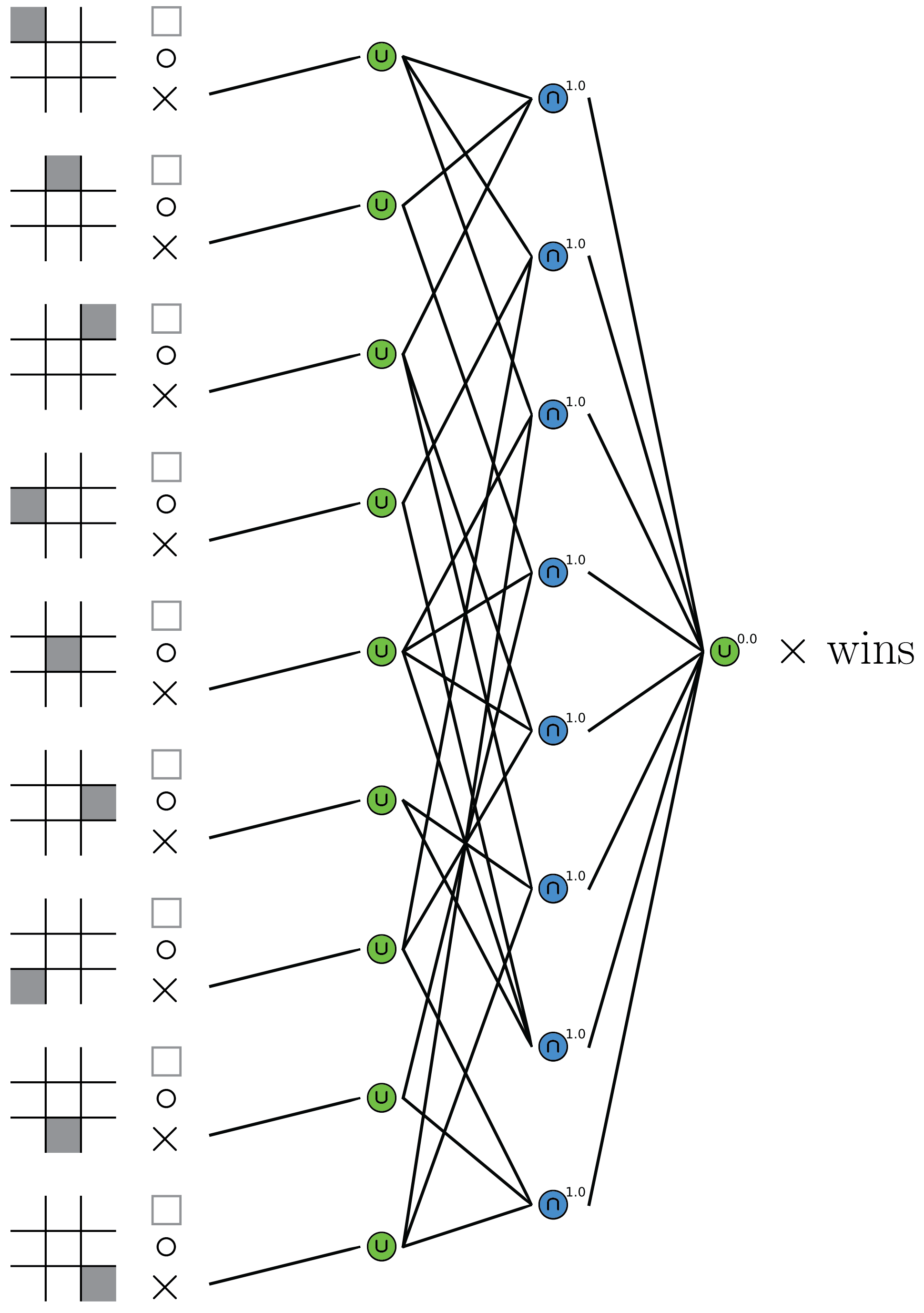}
         \caption{\textbf{NLN}}
         \label{fig:tictactoe_NLN}
     \end{subfigure}
     \hfill
     \begin{subfigure}[b]{0.45\textwidth}
         \centering
         \includegraphics[width=\textwidth]{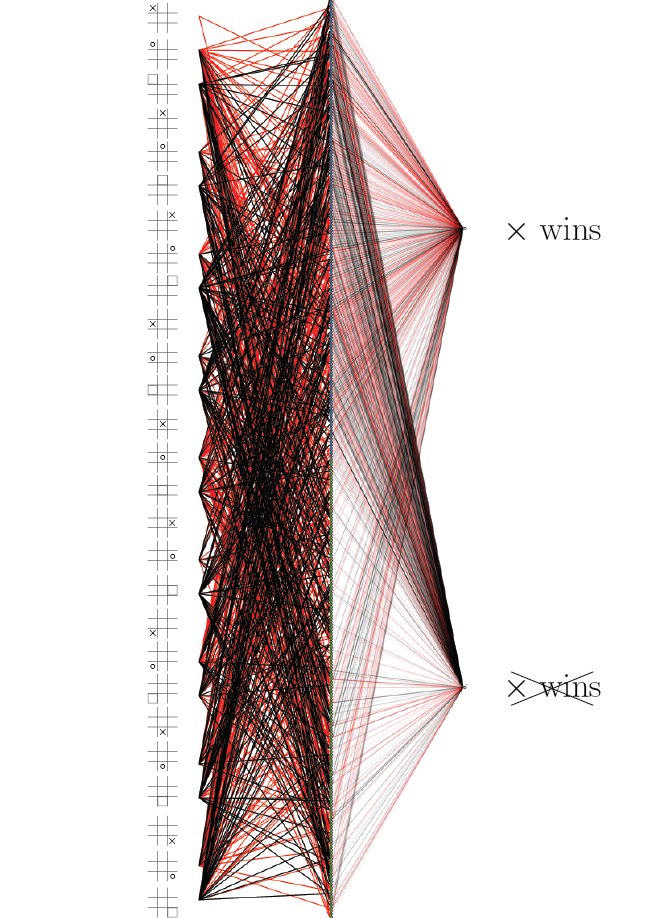}
         \caption{RRL}
         \label{fig:tictactoe_RRL}
     \end{subfigure}
    \caption{Interpretability of the learned AND/OR networks on tic-tac-toe. Our NLN with only AND/OR nodes recovers exactly the rules while the RRL with its output linear layer learns a distributed representation in its hidden layer of AND/OR nodes.}
    \label{fig:tictactoe}
\end{figure}

We will show that our NLN, with its factorized structure and extended modeling, is able to learn sparser and more interpretable rules than its predecessor, the RRL, in tabular classification. For instance, as illustrated in Figure \ref{fig:tictactoe}, our NLN is able to discover the rules of tic-tac-toe, simply by predicting if $\times$ has won from the end-game board configuration.

\section{Theory}
\label{sec:theory}

The modeling behind the AND/OR neurons in NLNs is probabilistic, but it can also be viewed through the lens of fuzzy logic or logic programming. Moreover, we present opportunities and issues in their interpretation.

\subsection{Probabilistic modeling}

We introduce the probabilistic modeling with a toy setting, before moving on to the general setting and addressing the necessary assumptions of independence that were made in order to obtain a tractable model.

\subsubsection{Toy setting}

\begin{figure}[t]
    \centering
     \begin{subfigure}[b]{0.45\textwidth}
         \centering
         \includegraphics[width=\textwidth]{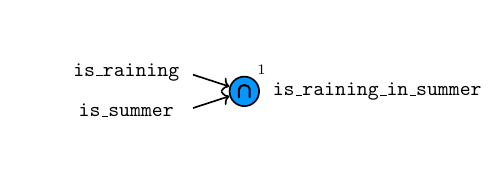}
         \caption{Necessary components of a situation, where 1 indicates that all necessary components are observed}
         \label{fig:interp_AND1}
     \end{subfigure}
     \hfill
     \begin{subfigure}[b]{0.45\textwidth}
         \centering
         \includegraphics[width=\textwidth]{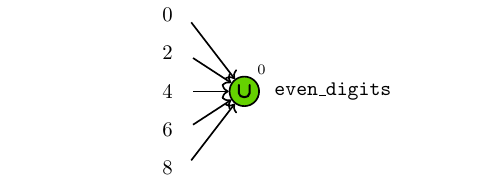}
         \caption{Possible cases of an equivalency class, where 0 indicates that no other case is missing or unobserved}
         \label{fig:interp_OR1}
     \end{subfigure}

     \vspace{12pt}
     \begin{subfigure}[b]{0.45\textwidth}
         \centering
         \includegraphics[width=\textwidth]{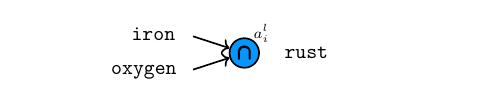}
         \caption{Necessary causal ingredients producing a consequence, where $a_i^l$ represents the probability of other missing ingredients, \eg water or humidity\\~}
         \label{fig:interp_AND2}
     \end{subfigure}
     \hfill
     \begin{subfigure}[b]{0.45\textwidth}
         \centering
         \includegraphics[width=\textwidth]{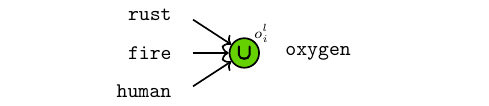}
         \caption{Possible consequences of a causal ingredient, where $o_i^l$ represents the probability that the causal ingredient is still present when none of these consequences are observed}
         \label{fig:interp_OR3}
     \end{subfigure}

     \vspace{12pt}
     \begin{subfigure}[b]{0.45\textwidth}
         \centering
         \includegraphics[width=\textwidth]{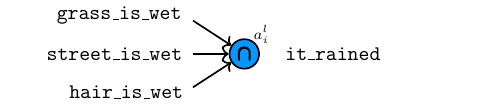}
         \caption{Necessary consequences of a cause, where $a_i^l$ represents the probability that the cause is indeed present when all the consequences are present, \ie that the consequences are not explained by other causes}
         \label{fig:interp_AND3}
     \end{subfigure}
     \hfill
     \begin{subfigure}[b]{0.45\textwidth}
         \centering
         \includegraphics[width=\textwidth]{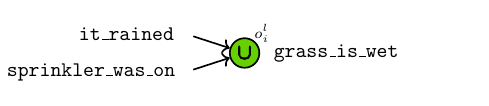}
         \caption{Possible causes of a consequence, where $o_i^l$ represents the probability of other missing causes, \eg a bucket of water was dropped\\~}
         \label{fig:interp_OR2}
     \end{subfigure}

     \vspace{12pt}
     \begin{subfigure}[b]{0.45\textwidth}
         \centering
         \includegraphics[width=\textwidth]{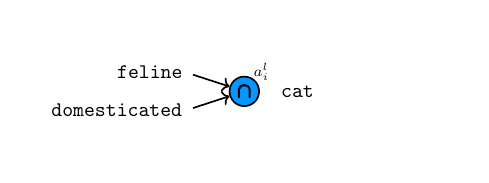}
         \caption{Necessary parent concepts of a sub-concept, where $a_i^l$ represents the probability of the other missing parent concepts of the sub-concept, \eg having partially webbed feet as opposed to the leopard cat, another domesticated feline which has fully webbed feet}
         \label{fig:interp_AND4}
     \end{subfigure}
     \hfill
     \begin{subfigure}[b]{0.45\textwidth}
         \centering
         \includegraphics[width=\textwidth]{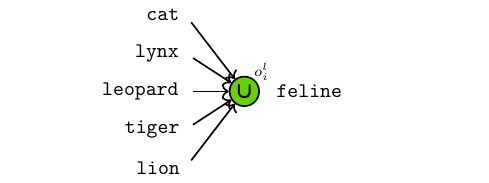}
         \caption{Possible sub-concepts of a parent concept, where $o_i^l$ represents the probability of another missing sub-concept, \eg a cougar or a panther, etc.\\ \\}
         \label{fig:interp_OR4}
     \end{subfigure}
    \caption{Toy examples of interpretations for AND and OR concepts}
    \label{fig:interp_AND_OR}
\end{figure}

We introduce the modeling with a toy example. We are given an object $x$ from which we can derive a number of binary properties $C_i$ about the object $x$, \ie \emph{concepts} $C_i$ that are either present or absent in $x$. For instance, we might know whether it is a \texttt{ball}, $x \in C_1$, where $C_1$ is the set of balls, whether it is \texttt{green}, $x \in C_2$, where $C_2$ is the set of green objects, if it is \texttt{heavy}, $x \in C_3$, with $C_3$ the set of heavy objects, and so on. For instance a light green ball $x$ would satisfy $x \in C_1$, $x \in C_2$, and $x \notin C_3$. From these properties, we are interested in predicting whether the object $x$ is a \texttt{green\_ball}, $x \in Y$, where $Y$ is the set of green balls. Moreover, by learning to predict whether an object $x$ is a green ball $Y$, we wish to discover the definition of a green ball, $Y = C_1 \cap C_2$, \ie that it is the set of objects that are both green and balls. More generally, we are interested in learning an \emph{AND concept}
\begin{align*}
    Y = \bigcap_{i \in \mathcal{N}} C_i,
\end{align*}
where, in this case, its \emph{necessary} concepts are indexed by $\mathcal{N} = \{1,2\}$. Figure \ref{fig:interp_AND_OR} presents examples of AND concepts and their necessary concepts. We learn to estimate which concepts $C_i$ are necessary to $Y$, written $Y \subseteq C_i$ or equivalently $C_i \supseteq Y$, through the statistical learning process of gradient descent. We do this by modeling the random event $x \in Y$ with respect to its necessary concepts $C_i$.
\begin{align*}
    x \in Y \;&=\; \bigcap_{i \in \mathcal{N}} \Big( x \in C_i \Big) & \textrm{[definition of }Y\textrm{]}\\
    &=\; \bigcap_{i} \Big( \big( i \notin \mathcal{N} \big) \,\cup\, \Big(\big( i \in \mathcal{N} \big) \,\cap\, \big(x \in C_i\big)\Big)\Big) & \textrm{[equivalent rewriting]}\\
    &=\; \bigcap_{i} \Big( \big( i \notin \mathcal{N} \big) \,\cup\, \big(x \in C_i\big)\Big) & \textrm{[distributivity of }\cup\textrm{ over }\cap\textrm{]}\\
    &=\; \bigcap_{i} \Big( \big( C_i \not\supseteq Y\big) \,\cup\, \big(x \in C_i\big)\Big) & \textrm{[definition of }\mathcal{N}\textrm{]}
\end{align*}
For instance, an object $x$ is a green ball $x \in Y$ if, for all its input concepts $C_i$, either they are not necessary concepts like heavy $C_3 \not\supseteq Y$, or they are present in $x$ like green $x \in C_2$ and ball $x \in C_1$.

We could otherwise have been interested in a different kind of concept, an \emph{OR concept}. For instance, in a feedforward network where successive layers learn higher-level concepts from lower-level concepts, we could have a first layer of AND concepts that defines concepts such as \mbox{$C_1'$: \texttt{green\_ball}, $C_2'$: \texttt{yellow\_cup}, $C_3'$: \texttt{blue\_stick}} (where $'$ distinguishes from previous examples), and so on and we could be interested in learning in the second layer $Y$: \texttt{green\_ball\_or\_blue\_stick}, \ie $Y = C_1' \cup C_3'$. More generally, we would be learning an OR concept
\begin{align*}
    Y = \bigcup_{i \in \mathcal{S}} C_i',
\end{align*}
where, in this case, its \emph{sufficient} concepts are indexed by $\mathcal{S} = \{1,3\}$. Figure \ref{fig:interp_AND_OR} presents examples of OR concepts and their sufficient concepts. We again use gradient-based statistical learning techniques to estimate which concepts $C_i'$ are sufficient to $Y$, written $C_i' \subseteq Y$, by modeling $x \in Y$ in function of the $C_i'$.
\begin{align*}
    x \in Y \;&=\; \bigcup_{i \in \mathcal{S}} \Big( x \in C_i' \Big) & \textrm{[definition of }Y\textrm{]}\\
    &=\; \bigcup_{i} \Big(\big( i \in \mathcal{S} \big) \,\cap\, \big(x \in C_i'\big)\Big) & \textrm{[equivalent rewriting]}\\
    &=\; \bigcup_{i} \Big( \big( C_i' \subseteq Y \big) \,\cap\, \big(x \in C_i'\big)\Big) & \textrm{[definition of }\mathcal{S}\textrm{]}
\end{align*}
For instance, an object $x$ is a green ball or a blue stick $x \in Y$ if, for any concept in the previous layer $C_i'$, it is both a sufficient concept like green ball $C_1' \subseteq Y$ or blue stick $C_3' \subseteq Y$ and it is present in $x$ like a green ball $x \in C_1'$ or a blue stick $x \in C_3'$.

\subsubsection{General setting}

In general, we are given as \emph{input} a random variable $x=(x_1,x_2,...,x_m)$ that is defined by $m$ measures, where each \emph{input feature} $x_i$ can be binary, categorical or continuous. We are interested in predicting whether certain \emph{target} concepts $Y_k$ are present in the input $x$, formalized as the random event $x \in Y_k$. These targets can be classes in binary or multi-class classification, or labels in multi-label classification. We use a network of concepts $C_i^l$ arranged in layers $l \in \{0,1,...,L\}$ of size $n^l$ with $i \in \{1,...,n^l\}$, where the network's output is $Y_k = C_k^L$. The input layer $l=0$ is made up of concepts $C_i^0$ that can be directly extracted from the input $x$. In other words, given an input $x$, we know for each concept $C_i^0$ whether it is present $x \in C_i^0$ or not $x \notin C_i^0$ (see details in Section \ref{sec:ML_int_struct_input}). The subsequent layers of AND and OR concepts $C_i^l$ will try to learn relevant representations that relate \emph{logically} the input concepts $C_i^0$ to the target concepts $Y_k$ through combinations of AND and OR operations.

In practice, we predict the labels by modeling their conditional probability $\mathbb{P}\!\left[x \in Y_k\mid x\right]$ given the input $x$. We do so by modeling for each concept $C_i^l$ its conditional probability $c_i^l(x) = \mathbb{P}\!\left[x \in C_i^l\mid x\right]$, starting from the input concepts $C_i^0$ for which we already know their probabilities $c_i^0(x) = \mathbb{P}\!\left[x \in C_i^0\mid x\right]$. These input probabilities $c_i^0(x)$ can take any value in $[0,1]$, with binary values $\{0,1\}$ representing certain knowledge about $x$. In a feedforward structure, an AND (resp. OR) concept $C_i^l$ in layer $l$ takes its necessary (resp. sufficient) concepts from the previous layer $l-1$. Unlike in the previous toy examples, we consider that a concept $C_j^{l\shortminus 1}$ or its \emph{contrary opposite} $(C_j^{l\shortminus 1})^c$ can be a necessary or sufficient concept. For instance, a \texttt{ball\_that\_is\_not\_green} would have as necessary concepts \texttt{ball} and $\texttt{green}^c$, \ie we would have
\begin{align*}
    \texttt{ball} \supseteq \texttt{ball\_that\_is\_not\_green}, \qquad \textrm{and} \qquad \texttt{green}^c \supseteq \texttt{ball\_that\_is\_not\_green}.
\end{align*}

\paragraph{Weights}

We introduce a matrix of \emph{weights} $A_{i,j}^l \in [-1,1]$ for AND concepts and $O_{i,j}^l \in [-1,1]$ for OR concepts to learn these necessary and sufficient relations, such that
\begin{align*}
    A_{i,j}^{l} = \underbrace{\mathbb{P}\!\left[C_j^{l\shortminus 1} \supseteq C_i^l\right]}_{\bracketlowersymb{A_{i,j}^{l}}{\scriptscriptstyle +}} - \underbrace{\mathbb{P}\!\left[(C_j^{l\shortminus 1})^c \supseteq C_i^l\right]}_{\bracketlowersymb{A_{i,j}^{l}}{\scriptscriptstyle -}}, \quad \textrm{and} \quad O_{i,j}^{l} = \underbrace{\mathbb{P}\!\left[C_j^{l\shortminus 1} \subseteq C_i^l\right]}_{\bracketlowersymb{O_{i,j}^{l}}{\scriptscriptstyle +}} - \underbrace{\mathbb{P}\!\left[(C_j^{l\shortminus 1})^c \subseteq C_i^l\right]}_{\bracketlowersymb{O_{i,j}^{l}}{\scriptscriptstyle -}}.
\end{align*}
It is important to note that the set inclusion relations $\subseteq$ are in opposite directions in necessary and sufficient relations. If a concept $C_j^{l\shortminus 1}$ is necessary to concept $C_i^l$, then whenever we have $x \in C_i^l$, we must also have $x \in C_j^{l\shortminus 1}$ since it is necessary, hence $C_j^{l\shortminus 1} \supseteq C_i^l$. On the other hand, if a concept $C_j^{l\shortminus 1}$ is sufficient to concept $C_i^l$, then whenever we have $x \in C_j^{l\shortminus 1}$, we must also have $x \in C_i^l$ since $C_j^{l\shortminus 1}$ is sufficient to $C_i^l$, hence $C_j^{l\shortminus 1} \subseteq C_i^l$. When $A_{i,j}^l > 0$ or $O_{i,j}^l > 0$, the concept $C_j^{l\shortminus 1}$ is believed to be necessary or sufficient to $C_i^l$ with probability $A_{i,j}^l$ or $O_{i,j}^l$. When $A_{i,j}^l < 0$ or $O_{i,j}^l < 0$, the absence of the concept $(C_j^{l\shortminus 1})^c$ is believed to be necessary or sufficient with probability $|A_{i,j}^l|$ or $|O_{i,j}^l|$. This modeling allows a single parameter to learn both possibilities simultaneously, since they are contradictory. However, doing so also assumes that at all times at least one of those probabilities \eg $\mathbb{P}\!\left[C_j^{l\shortminus 1} \supseteq C_i^l\right]$, $\mathbb{P}\!\left[(C_j^{l\shortminus 1})^c \supseteq C_i^l\right]$ is zero, with the rest of the probability mass distributed between the other and $\mathbb{P}\!\left[ C_j^{l\shortminus 1} \not\supseteq C_i^l, (C_j^{l\shortminus 1})^c \not\supseteq C_i^l\right]$. In other words, this modeling introduces a cognitive bias in the model %that ``jumps to conclusions'' regarding the sign of a causal role in the sense that it presumes only one sign is possible at once. 
such that it cannot simultaneously consider both contradictory roles at the same time.
It must consider one option fully, for instance $C_j^{l\shortminus 1} \supseteq C_i^l$ with $A_{i,j}^l > 0$, before reaching $A_{i,j}^l = 0$ and being able to consider the other option $(C_j^{l\shortminus 1})^c \supseteq C_i^l$ with $A_{i,j}^l < 0$.

\paragraph{Biases}

Moreover, we consider the possibility of missing or \emph{unobserved} data $u$ being part of the full \emph{relevant} data $\omega = (x,u)$, where $\omega$ is the concatenation of the observed data $x$ and unobserved data $u$. For instance, we could be trying to predict whether an object $\omega=(x,u)$ is a \texttt{green\_ball} without having any information in $x$ about the color of the object, only that it is a ball. In that case we would have to estimate statistically the probability that a ball is green $\mathbb{P}\!\left[u \in \texttt{green}\mid x \in \texttt{ball}\right]$ given the distribution of objects $\omega$ that we have seen. We now consider all the relevant data $\omega$ to model which concepts $C_i^l$ are present and with what probability $c_i^l(x) = \mathbb{P}\!\left[\omega \in C_i^l\mid x\right]$. Only for $l=0$, we have the input concepts $C_i^0$ which depend only on the input $x$, and for which we are always given the probabilities $c_i^0(x) = \mathbb{P}\!\left[x \in C_i^0\mid x\right]$. We use $\mathcal{X}$ and $\mathcal{U}$ to denote the space of possible $x$ and $u$ values respectively, so that we have $\omega \in \mathcal{X} \!\times \mathcal{U}$ and $C_i^l \subseteq \mathcal{X} \!\times \mathcal{U}$. With this final extension, an AND concept $C_i^l$ would be defined as 
\begin{align*}
    C_i^l &= \Bigg(\bigcap_{j \in \mathcal{N}_+} C_j^{l\shortminus 1}\Bigg) \cap \Bigg(\bigcap_{j \in \mathcal{N}_-} (C_j^{l\shortminus 1})^c\Bigg) \cap \Bigg(\mathcal{X} \!\times \underbrace{\bigcap_{z \in \tilde{\mathcal{N}}} \tilde{C}_z}_{\displaystyle \tilde{N}_i^l}\Bigg),
\end{align*}
where $\mathcal{N}_+$ and $\mathcal{N}_-$ are its \emph{observed} necessary concepts in the previous layer, either using directly $C_j^{l\shortminus 1}$ or using its opposite $(C_j^{l\shortminus 1})^c$, and the $\tilde{C}_z$ represent unobserved concepts that depend only on the unobserved data $u$ and which are also necessary concepts of $C_i^l$. We define $\tilde{N}_i^l$ as being the intersection of all these necessary but unobserved concepts $\tilde{C}_z,~ \forall\, z \in \tilde{\mathcal{N}}$, \ie we have $u \in \tilde{N}_i^l$ iff we have $u \in \tilde{C}_z,~ \forall\, z \in \tilde{\mathcal{N}}$. The random event of whether the AND concept is present $\omega \in C_i^l$ is then given by
\begin{align*}
    \omega \in C_i^l &= (u \in \tilde{N}_{i}^{l}) \,\cap\, \bigcap_{C \in \mathcal{C}_{\pm}^{l\shortminus 1}} \Big(\big( C \not\supseteq C_i^l \big) \,\cup\, \big(\omega \in C\big) \Big), \tag{D-AND}
\end{align*}
where we define the concepts of the previous layer and their opposites $\mathcal{C}_{\pm}^{l\shortminus 1} = \{C_1^{l\shortminus 1},(C_1^{l\shortminus 1})^c,...,\allowbreak C_{n^{l\shortminus 1}}^{l\shortminus 1},(C_{n^{l\shortminus 1}}^{l\shortminus 1})^c\}$. Equivalently, for an OR concept $C_i^l$, we would have
\begin{align*}
    C_i^l = \Bigg(\bigcup_{j \in \mathcal{S}_+} C_j^{l\shortminus 1}\Bigg) \cup \Bigg(\bigcup_{j \in \mathcal{S}_-} (C_j^{l\shortminus 1})^c\Bigg) \cup \Bigg(\mathcal{X} \!\times \underbrace{\bigcup_{z \in \tilde{\mathcal{S}}} \tilde{C}_z}_{\displaystyle \tilde{S}_i^l}\Bigg),
\end{align*}
where $\tilde{S}_i^l$ is the union of all its sufficient but unobserved concepts $\tilde{C}_z,~ \forall\, z \in \tilde{\mathcal{S}}$ and
\begin{align*}
    \omega \in C_i^l = (u \in \tilde{S}_{i}^{l}) \,\cup\, \bigcup_{C \in \mathcal{C}_{\pm}^{l\shortminus 1}} \Big(\big(C \subseteq C_i^l \big) \,\cap\, \big(\omega \in C\big) \Big). \tag{D-OR}
\end{align*}
Modeling these necessary/sufficient unobserved concepts introduces \emph{biases} $a_{i}^l \in [0,1]$ for AND concepts and $o_{i}^l \in [0,1]$ for OR concepts, defined as the conditional probabilities
\begin{align*}
    a_{i}^{l} &= \mathbb{P}\!\left[u \in \tilde{N}_{i}^{l} \,\middle|\, \bigcap_{C \in \mathcal{C}_{\pm}^{l\shortminus 1}} \Big(\big( C \not\supseteq C_i^l \big) \,\cup\, \big(\omega \in C\big) \Big)\right], \qquad
    o_{i}^{l} = \mathbb{P}\!\left[u \in \tilde{S}_{i}^{l} \,\middle|\,\left(\bigcup_{C \in \mathcal{C}_{\pm}^{l\shortminus 1}} \Big(\big(C \subseteq C_i^l \big) \,\cap\, \big(\omega \in C\big) \Big) \right)^{\!\!\!c~}\right].
\end{align*}
For an AND concept, $a_{i}^{l}$ is the probability that all the unobserved necessary concepts are present $u \in \tilde{N}_{i}^{l}$ when all the observed necessary concepts are present. This indicates how often this AND concept is indeed activated when the input $x$ suggests that it should. If $a_{i}^{l}=0$, then this AND concept is never activated and it becomes useless in the modeling. For an OR concept, $o_{i}^{l}$ is the probability that any unobserved sufficient concept is present $u \in \tilde{S}_{i}^{l}$ when no observed sufficient concept is present. This indicates how often this OR concept is activated purely by unobserved concepts. If $o_{j}^{l}=1$, then this OR concept is always trivially activated and it is also useless in the modeling. For both types of concepts, these probabilities are measures of how much relevant information we are missing in the input data $x$ to fully model this concept. Although these unobserved concepts depend only on the unobserved data $u$, they are modeled with respect to $\omega$ since $u$ and $x$ are likely correlated in practice. Figure \ref{fig:interp_AND_OR} presents examples of such biases from unobserved concepts for AND and OR concepts.

\subsubsection{Tractable modeling and assumptions}

Our modeling contains two distinct types of probabilities, which represent different types of uncertainty. If we knew exactly how the concepts in our network were related logically, \ie if we knew the exact structure of the ground-truth network with its exact weights $A_{i,j}^l$ and $O_{i,j}^l$, then the probabilities of presence of concepts  $c_i^l(x)$, $a_{i}^{l}$ and $o_{i}^{l}$ would all be strictly \emph{aleatoric} probabilities. For instance, in our green ball example, if we knew the definition of a green ball $C_i^l$, then the probability of an object with observed features $x$ being a ball $c_j^{l\shortminus 1}(x)$, being a green ball $c_i^{l}(x)$ and the probability of any ball being green $a_i^l=\mathbb{P}\!\left[u \in \texttt{green}\mid x \in \texttt{ball}\right]$ are all aleatoric. They are statistical quantities that depend only on the distribution of specific realizations $\omega = (x, u)$ that the network has seen in training. In contrast, the beliefs in the concepts' roles as necessary or sufficient to other concepts $A_{i,j}^l$ and $O_{i,j}^l$ are \emph{epistemic} probabilities. For instance, the probability that a green ball is necessarily green $A_{i,j}^l = \mathbb{P}\!\left[\texttt{ball} \supseteq \texttt{green\_ball}\right]$ is epistemic. They represent a priori \emph{beliefs} in the general causal mechanisms that underlie the random phenomenon that generated $\omega$, and are independent of any such particular realization $\omega$. In practice, since $c_i^l(x)$, $a_{i}^{l}$ and $o_{i}^{l}$ are defined with respect to the believed roles, these probabilities model both aleatoric and epistemic uncertainty.

We obtain a tractable probabilistic modeling of the NLNs by making three assumptions of independence, to which we will return shortly. They allow the probabilities $c_i^l(x)$ to be easily computed in a parallelizable fashion (the full derivation is given in appendix \ref{sec:app_theory_prob_deriv_with_3}).
\begin{align*}
    c_i^{l}(x) &= a_i^{l} \prod_{j=1}^{n^{l\shortminus1}} \Big( 1 - \pospart{A_{i,j}^{l}} \Big( 1 - c_j^{l\shortminus1}(x) \Big) \Big)\Big( 1 - \negpart{A_{i,j}^{l}} \, c_j^{l\shortminus1}(x) \Big) \tag{P-AND}\\
    c_{i}^{l}(x) &= 1 - \big(1- o_i^{l}\big) \prod_{j=1}^{n^{l\shortminus1}} \Big( 1 - \pospart{O_{i,j}^{l}} \; c_j^{l\shortminus1}(x)\Big) \Big(1 - \negpart{O_{i,j}^{l}} \Big( 1 - c_j^{l\shortminus1}(x) \Big) \Big) \tag{P-OR}
\end{align*}

The first two assumptions of independence that are required to obtain (P-AND) and (P-OR) are modeling choices, while the third one is an approximation. The first assumption is %\emph{independence between the presence of concepts in some $\omega$ and their general roles in the next layer as necessary or sufficient}
independence between \emph{the presence in some $\omega$} of concepts in the same layer and \emph{their general roles} in the next layer as \emph{necessary or sufficient}
\begin{align*}
    \omega \in C &\indep C' \supseteq C_i^{l+1}, &\textrm{for all input concepts } C,C' \in \mathcal{C}_{\pm}^{l} \textrm{ of AND concept } C_i^{l+1},\\
    \omega \in C &\indep C' \subseteq C_i^{l+1}, &\textrm{for all input concepts } C,C' \in \mathcal{C}_{\pm}^{l} \textrm{ of OR concept } C_i^{l+1}.\hphantom{\textrm{N}}
\end{align*}
We compute the probabilities $c_i^l(x) = \mathbb{P}\!\left[\omega \in C_i^l \mid x\right]$ using epistemic beliefs $A_{i',j'}^{l'}$, $O_{i',j'}^{l'}$ of the previous layers $l' \leq l$. The presence of a concept is thus certainly not independent of the roles that form its own definition in the previous layers. However, this assumption of independence is between the presence of concepts in a layer $l$ and their roles in the next layer $l+1$. In other words, we assume that the presence of concepts in a particular realization $\omega$ does not give any information regarding their general roles in higher-level concepts, and vice versa. For instance, the probability that some particular object is a ball $\omega \in \texttt{ball}$ is independent from the probability that all green balls are necessarily balls $\texttt{ball} \supseteq \texttt{green\_ball}$ or are necessarily heavy $\texttt{heavy} \supseteq \texttt{green\_ball}$. This independence would not hold if we were conditioning on the presence of the concepts in the next layer, \eg
\begin{align*}
    \left.\omega \in C \not\indep C \supseteq C_i^{l+1} ~~\middle|~~ \omega \in C_i^{l+1}\right. .
\end{align*}
If, for instance, we have for some $\omega$ that $C$ is absent ($\omega \notin \texttt{heavy}$) but $C_i^{l}$ is present ($\omega \in \texttt{green\_ball}$), then it is impossible that $C$ could ever be a necessary concept of $C_i^{l}$ ($\texttt{heavy} \not\supseteq \texttt{green\_ball}$). Without conditioning however, it is conceivable that knowing whether we have $\omega \in C$ for some $\omega$ gives by itself no information on its general roles in the next layer, or on the roles of other concepts in the same layer.

The second assumption is %\emph{independence between the necessary/sufficient concepts of a concept}
independence between \emph{the necessary/sufficient concepts} of a concept
\begin{align*}
    C \supseteq C_i^{l+1} &\indep C' \supseteq C_i^{l+1}, &\textrm{for all distinct input concepts } C,C' \in \mathcal{C}_{\pm}^{l} \textrm{ of AND concept } C_i^{l+1},\\
    C \subseteq C_i^{l+1} &\indep C' \subseteq C_i^{l+1}, &\textrm{for all distinct input concepts } C,C' \in \mathcal{C}_{\pm}^{l} \textrm{ of OR concept } C_i^{l+1}.\hphantom{\textrm{N}}
\end{align*}
where, by distinct, we mean that $C' \notin \{C, \lnot C\}$. This assumption would be false for an observer who has previous knowledge and understanding about the concepts he is manipulating. For instance, if an AND concept already has \texttt{green} as a necessary concept, then it could not also have \texttt{red} or another incompatible color as necessary concepts. Moreover, an AND concept that already has \texttt{green} and \texttt{ball} would also be more likely to have \texttt{presence\_of\_a\_tennis\_racket}. Although this assumption does not hold for an observer with previous understanding, it would hold for an observer who has absolutely no idea or previous understanding of what concepts he is manipulating, like an agent in an alien environment or the operator in Searle's ``Chinese room'' thought experiment \citep{Searle80}. However, for an observer with previous knowledge, this assumption would amount to a cognitive bias of ``total open-mindedness'' that considers every combination of concepts to be equiprobable, irrespective of how related or incongruous they might be. We summarize the probabilistic modeling up to this point, including the first two assumptions of independence, in a PGM presented in appendix \ref{sec:app_theory_prob_PGM}.

The third assumption which we use as an approximation is %\emph{conditional independence between concepts in the same layer, given the input}
conditional independence between \emph{concepts in the same layer}, given the input
\begin{align*}
    \omega \in C &\indep \left. \omega \in C' \vphantom{C_i^l}~~\middle|~~ x,\right. &\textrm{for all distinct concepts in the same layer } C,C' \in \mathcal{C}_{+}^{l},
\end{align*}
where we defined the concepts of a layer $\mathcal{C}_{+}^{l} = \{C_1^{l},...,C_{n^{l}}^{l}\}$. This assumption is the least realistic, being false in most cases. For instance, $\omega \in \texttt{heavy\_green\_ball}$ and $\omega \in \texttt{bouncy\_green\_ball}$ would be considered independent given $x \in \texttt{ball}$, even though they both depend on $\omega \in \texttt{green\_ball}$ in a previous layer, which uses unobserved information $u \in \texttt{green}$. This assumption is trivially satisfied by the input features $x \in C_i^0$ since they only depend on $x$. However, for the other concepts defined in layers $l \geq 1$, this assumption is only true if no two concepts share a common necessary/sufficient concept in the previous layer, a property known as decomposability in the probabilistic circuits literature \citep{Choi20}. This is very unlikely for non trivial NLNs, and it becomes progressively less likely as $L$ increases. Although this assumption is incorrect in most practical cases, avoiding it results in a combinatorial explosion of computations for the forward pass alone (see appendix 
\ref{sec:app_theory_prob_deriv_without_3} for the full derivation without it). As a result, the method would become intractable and unusable in practice without this assumption. As a purely probabilistic model, this issue is catastrophic for NLNs. However, as a ML method, this approximation can be justified. First of all, in the case where all inputs are binary (\ie with binary or categorical features), where all the weights are given from the set $\{-1,0,1\}$ and where the biases are full with $a_{\cdot}^l = 1$, $o_{\cdot}^{l'}=0$, then this assumption is not necessary. In fact, none of the assumptions are necessary in that case because the probabilistic formulation (P-AND) and (P-OR) coincide exactly with the logical definition of the AND and OR nodes (see section \ref{sec:theory_fuzzy}). In general, we always need the weights to be integers for maximum interpretability and, as such, we discretize them during post-processing when learning a NLN (see section \ref{sec:ML_post}). Moreover, for any network that contains continuous features, with enough pre-processing nodes (see section \ref{sec:ML_int_struct_input}), the signal can become arbitrarily close to binary thanks to appropriately scaled sigmoid curves. In practice, we only relax this binary constraint through the possible non-binary biases, as well as by limiting the number of pre-processing nodes for continuous features, so that values away from 0 and 1 can be obtained close to the learned boundaries (see section \ref{sec:ML_int_struct_input}). Therefore, we keep this approximation leading to (P-AND) and (P-OR) and, in practice, this modeling is still able to obtain promising predictive performance and interpretable rule discovery, even in non-binary cases.

\subsection{Fuzzy logic equivalency}
\label{sec:theory_fuzzy}

A fuzzy logic is a real-valued generalization of classical logic, where instead of a statement like $x \in Y_k$ being either \emph{true} or \emph{false}, it can have a \emph{degree of truth} between 0 (false) and 1 (true). The different fuzzy logics  differ by their choices of how to generalize the classical logic operators
\begin{align*}
	\land:~\textrm{AND}, \qquad\qquad \lor:~\textrm{OR}, \qquad\qquad \lnot:~\textrm{NOT},
\end{align*}
\etc For instance, product fuzzy logic is the fuzzy logic that coincides with probability distributions in which all random events are independent of each other. They thus define their fuzzy logic operators as
\begin{align*}
	\stackrel{{\textcolor{gray}{\scriptscriptstyle \textrm{P}}}}{\bigwedge_i} v_i := \prod_i v_i, \qquad\qquad \stackrel{{\textcolor{gray}{\scriptscriptstyle \textrm{P}}}}{\bigvee_i} v_i := 1 - \prod_i (1-v_i), \qquad\qquad \pnot v := 1-v,
\end{align*}
by using the product t-norm ($\pand$), the probabilistic sum t-conorm ($\por$), and the strong negation ($\pnot$) \citep{Klement10}.

We can rewrite (P-AND) and (P-OR) with the product fuzzy logic operators
\begin{align*}
    c_i^{l}(x) &= a_i^{l} \ \pand\; \stackrel{{\textcolor{gray}{\scriptscriptstyle \textrm{P}}}}{\bigwedge_{j \in \{1, ..., n^{l\shortminus1}\}}} \Big( \pnot \pospart{A_{i,j}^{l}} \;\por\; c_j^{l\shortminus1}(x) \Big) \;\pand\; \Big( \pnot \negpart{A_{i,j}^{l}} \;\por\; \pnot\, c_j^{l\shortminus1}(x) \Big), \tag{F-AND}\\
    c_{i}^{l}(x) &= o_i^{l} \ \por\; \stackrel{{\textcolor{gray}{\scriptscriptstyle \textrm{P}}}}{\bigvee_{j \in \{1, ..., n^{l\shortminus1}\}}} \Big( \pospart{O_{i,j}^{l}} \;\pand\; c_j^{l\shortminus1}(x)\Big) \;\por\; \Big(\negpart{O_{i,j}^{l}} \;\pand\; \pnot\, c_j^{l\shortminus1}(x) \Big), \tag{F-OR}
\end{align*}
By doing so,  we obtain the product fuzzy logic generalization of the logical definition of our operators
\begin{align*}
    \omega \in C_i^l &= (u \in \tilde{N}_{i}^{l}) \,\land\, \bigwedge_{C \in \mathcal{C}_{\pm}^{l\shortminus 1}} \Big(\lnot \big( C \supseteq C_i^l \big) \,\lor\, \big(\omega \in C\big) \Big), \tag{L-AND}\\
    \omega \in C_i^l &= (u \in \tilde{S}_{i}^{l}) \,\lor\, \bigvee_{C \in \mathcal{C}_{\pm}^{l\shortminus 1}} \Big(\big(C \subseteq C_i^l \big) \,\land\, \big(\omega \in C\big) \Big). \tag{L-OR}
\end{align*}
This logical definition is equivalent to our previous probabilistic definition (D-AND) and (D-OR). Only the symbols have changed from one formalism to another. The intersections $\cap$ of random events have become conjunctions $\land$ (AND) of truth values, the unions $\cup$ have become disjunctions $\lor$ (OR) and the complement $\cdot^c$ implied in $\big( C \not\supseteq C_i^l \big)=\big( C \supseteq C_i^l \big)^c$ has become negation $\lnot$ (NOT). It is interesting to note that we could rewrite (F-AND) and (F-OR) with a different fuzzy logic with differentiable t-norm and t-conorm such as Łukasiewicz logic and we would still obtain a learnable logic network. Doing so would keep the underlying logical modeling of our method with (L-AND) and (L-OR), but we would lose its probabilistic interpretation.

Previous formalizations of NLNs omitted any probabilistic or logical modeling of the AND and OR concepts and instead directly used fuzzy logic formulations like (F-AND) and (F-OR) \citep{Payani19, Payani20, Wang20, Wang21, Wang24, Zhang23}. They obtained similar formulations to our own by designing them according to a desiderata of the expected behavior of AND/OR nodes. However, their formulations were more restricted, being special cases of our own. They did not consider negated concepts in their formalism, \ie they assumed $A_{i,j}^l, O_{i,j}^l \geq 0$. To consider negated concepts in practice, they would instead duplicate the inputs as hardcoded negated versions of themselves, hence also doubling the number of weights. Moreover, none of the previous formalizations considered the effect of unobserved concepts $\tilde{C}_z$ and data $u$. They assumed no unobserved necessary or sufficient concepts with $a_i^l = 1$ and $o_j^l = 0$. In addition to considering unobserved concepts, our probabilistic modeling also avoids having to assume independence between these unobserved concepts and the rest of the definitions. It also avoids assuming independence between a concept's causal role and its contrary's causal role. These assumptions would have been implied by a strictly product fuzzy logic modeling, such as in previous formalizations.

\subsection{Logical perspective}

AND, OR and NOT are fundamental operators in classical logic and some of their properties are relevant to our approach. Firstly, just like intersection, union and complement, we have that conjunction (AND), disjunction (OR) and negation (NOT) are related by De Morgan's laws and distributivity. Our formalism is compatible with De Morgan's laws, as can be seen from (P-AND)-(P-OR), (F-AND)-(F-OR), (L-AND)-(L-OR) and (D-AND)-(D-OR) (proofs in appendix \ref{sec:app_theory_log_deMorgan}). As a consequence, any AND concept can be converted to an OR concept, and vice versa, by flipping the signs of its incoming and outgoing weights and by taking the complement of its bias. However, since product fuzzy logic operators are not idempotent \citep{Krieken22}
\begin{align*}
    a \pand a &= a \cdot a \neq a, &a \por a &= 1 - (1 - a)\cdot(1-a) \neq a,
\end{align*}
distributivity can not be applied in our formalism, unless all probabilities $c_{i}^l(x)$, $|A_{i,j}^l|$, $|O_{i,j}^l|$, $a_i^l$, $o_i^l$ are binary.

Classical logic also studies logical formulas, which combine AND, OR and NOT operations. Using De Morgan's laws and distributivity, complex logical formulas can be rewritten in many equivalent formulations. Two notable forms are the Disjunctive Normal Form (DNF), a disjunction (OR) of conjunctions (AND) with negation, \eg
\begin{align*}
    (A_1 \land \lnot A_2 \land A_7) \lor (A_3) \lor (A_2 \land A_4 \land \lnot A_5),\tag{DNF}
\end{align*}
and the Conjunctive Normal Form  (CNF), a conjunction (AND) of disjunctions (OR) with negation, \eg
\begin{align*}
    (A_1 \lor A_2 \lor A_3) \land (\lnot A_2 \lor A_3 \lor A_4) \land (\lnot A_2 \lor A_3 \lor \lnot A_5) \land (A_2 \lor A_3 \lor A_7),\tag{CNF}
\end{align*}
equivalent to the (DNF) above. As this example shows, depending on the logical formula, one form might be much simpler than the other. In our setting, we want to describe when a target concept is present $\omega \in Y_k$ with a logical formula that combines the input features $x \in C_i^0$ and unobserved concepts $u \in \tilde{N}_i^l$, $u \in \tilde{S}_i^l$. In this setting, we use the DNF to describe a sufficient condition for the target concept to be present. In that case, the DNF can be written in the implicational form and interpreted as a (normal) logic program, which is a finite set of rules of the form
\begin{align*}
   B_1 \land ... \land B_{m} \land \lnot B_{m+1} \land ... \land \lnot B_n &\rightarrow H
\end{align*}
where $H$ is called the head and the left-hand side is called the body. The usual notation is written from right to left, but we adopt the opposite convention here to coincide with the graphical representation of neural networks. Such a rule says that if the condition in the body is true, then the head $H$ is true. For instance, suppose that B is implied by the formula (DNF) above, then it can be described by the following logic program
\begin{align*}
    A_1 \land \lnot A_2 \land A_7 &\rightarrow B, \\
    A_3 &\rightarrow B, \\
    A_2 \land A_4 \land \lnot A_5 &\rightarrow B.
\end{align*}
If any of these 3 rules is activated by the left-hand side AND combination being true, then $B$ is true, and otherwise it is false. The OR is implied by the fact that any activated rule is sufficient for $B$ to be true. In Section \ref{sec:ML_int_struct}, we will encode this logic program formulation into the structure of our NLNs in order to be able to interpret what a NLN has learned as a set of rules.

\subsection{Interpretation}

Although a NLN attempts to learn the underlying causal mechanism relating the inputs to the outputs, it cannot uniquely determine the causal structure or the direction of causality. An AND concept or an OR concept can each represent many different cases. Some possible interpretations are given in Table \ref{tab:interpretations} with toy examples for each pictured in Figure \ref{fig:interp_AND_OR}.

\begin{table}[h!]%[t]
  \centering
  \caption{Some possible interpretations of the AND and OR concepts}
  \label{tab:interpretations}
  \scalebox{.9}{\begin{tabular}{rclcrcl}
    \toprule
    % \multicolumn{3}{c}{AND} & $\!\!\!\!$ & \multicolumn{3}{c}{OR} \\
    % \cmidrule(r){1-3}\cmidrule(r){5-7}
    Necessary concepts & of an & AND concept & $\!\!$ & Sufficient concepts & of an & OR concept \\\midrule
    necessary components & of a & situation & $\!\!$ & possible cases & of an & equivalency class \\
    necessary causal ingredients & $\!\!\!$producing a$\!\!\!$ & consequence & $\!\!$ & possible consequences & of a & causal ingredient \\
    necessary consequences & of a & cause & $\!\!$ & possible causes & of a & consequence \\
    necessary parent concepts & of a & sub-concept & $\!\!\!\!$ & possible sub-concepts & of a & parent concept \\
    \bottomrule
  \end{tabular}}
\end{table}

Moreover, since a finite combination of AND (resp. OR) concepts can be represented by a single AND (resp. OR) concept, each type can represent an infinite number of cases. We illustrate some intuitive and counter-intuitive examples of such causal structures in appendix \ref{sec:app_theory_interp}. This multiplicity of cases is also exacerbated by the fact that a logical formula can be rewritten in many equivalent ways through De Morgan's laws and distributivity. This makes the interpretation of a NLN very difficult without additional expert knowledge. However, the equivalence of AND/OR concepts through De Morgan's laws can also be used as an advantage in the learning of deep NLNs. A layer of concepts using negation that is followed by another layer allowing negation can be learned with an arbitrary type and then be interpreted a posteriori, once the learning is done, by experts.

\section{Machine learning pipeline}
\label{sec:ML}

We propose a specific NLN structure and an associated learning method that can learn a set of simple rules (logic program) to classify each input.

\subsection{Interpretable structure}
\label{sec:ML_int_struct}

\begin{figure}[h!]
    \centering
    \includegraphics[width=.6\textwidth]{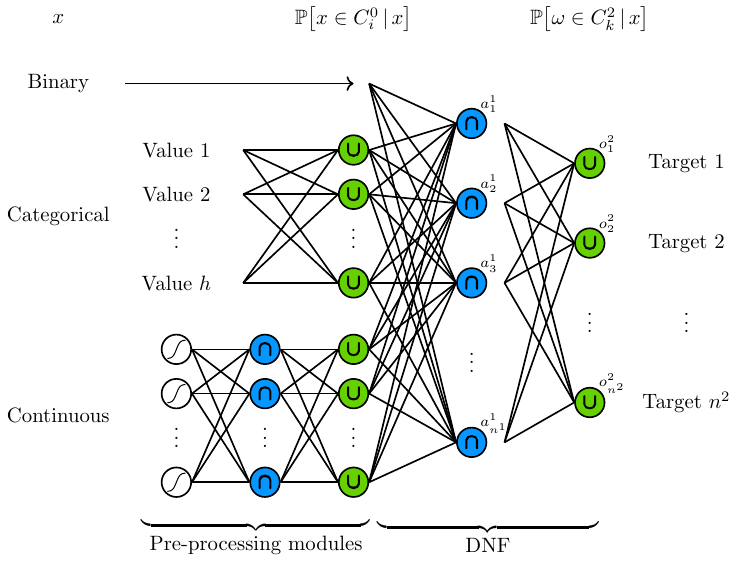}
    \caption{Structure of an interpretable NLN. Features have adapted pre-processing modules that can learn: for categorical features, (OR) equivalency classes of values with the same behavior; and, for continuous features, (OR) collections of (AND) intervals with the same behavior. These representations are then combined into (AND) rules that make up each (OR) logic program associated to each target class/label $Y_k$.}
    \label{fig:nln_structure}
\end{figure}

By interpretability, we mean the ability to provide its meaning in human-understandable terms \citep{Arrieta20}. In this sense, the AND and OR concepts that we have defined are interpretable so long as their inputs are themselves interpretable. To ensure that the full NLN model is interpretable, we can impose inductive biases in its structure to ensure that each AND/OR concept learns a meaningful concept. We propose the structure pictured in Figure \ref{fig:nln_structure}. It contains two fully-connected layers arranged in DNF, \ie an AND layer with negation followed by an OR layer without negation, in order to learn a logic program for each target (section \ref{sec:ML_int_struct_DNF}). The input features that are not binary are pre-processed with appropriate input modules, one for categorical features, and another for continuous features (section \ref{sec:ML_int_struct_input}).

\subsubsection{Fully-connected DNF layers}
\label{sec:ML_int_struct_DNF}

NLNs, as we have defined them, could be constructed with arbitrary depth. However, deep NLNs pose challenges regarding learning and intepretability. NLNs in general are difficult to learn because of vanishing gradients \citep{Payani19, Wang20, Wang21, Wang24, Krieken22} and depth exacerbates this issue. Deeper networks are also harder to interpret. The first fully-connected layer of concepts can be easily interpreted because it is directly defined in terms of the input features, which are usually interpretable. However, the next layers are defined by combining these higher-level concepts and their resulting definitions become increasingly harder to interpret. Their definitions in terms of the input features are not only more indirect, their activation patterns with respect to those features are also much more complex.

To avoid these issues, we restrict ourselves to two fully connected layers arranged in DNF. The hidden layer is made up of AND concepts that allow negation and the output layer is made up of OR concepts that do not allow negation. As mentioned previously, this DNF can learn any logical formula and can be interpreted as a logic program. Each target output concept (OR) is implied by any one of multiple rules (AND), which are each activated by a specific combination of values from the input features. Although the OR output concepts do not allow negation, they remain just as expressive. Moreover, this way, the AND rules in the hidden layer cannot be re-interpreted as OR concepts (through De Morgan's laws) and, therefore, they always remain interpretable as rules.

\subsubsection{Input pre-processing modules}
\label{sec:ML_int_struct_input}

Our previous modeling assumed that the input concepts $x \in C_i^0$ are binary or represent probabilities of random events $c_i^0(x) = \mathbb{P}\!\left[x \in C_i^0\mid x\right]$. Hence, \emph{binary} input features can directly be used as input concepts, but \emph{categorical} (one value out of a finite set of possible values) or \emph{continuous} (in a subset of $\mathbb{R}$) features must be pre-processed. We use a different pre-processing module for each to convert them into interpretable probabilities.

\paragraph{Categorical features}

Categorical features can be directly converted to binary variables with a one-hot encoding. However, feeding these one-hot encodings to the fully-connected DNF layers would needlessly multiply the number of rules in a model whenever multiple values of a category behave the same way in some circumstance. We instead introduce their own layer of OR concepts without negation and without unobserved sufficient conditions (since we can observe every possible value). These OR concepts learn equivalency classes of categorical values that have the same effect. In addition to reducing the number of duplicate rules for each related value, this encoding is also interpretable and results in a limited form of \emph{predicate invention}. Again, these OR concepts remain just as expressive even without allowing negation. The choice to not allow negation here is to help orient their learning away from trivial cases, such as when two or more values have a weight of $O_{i,j}^{0}=-1$.

\paragraph{Continuous features}

Features that are continuous need to be discretized in order to be manipulated by AND and OR concepts. \cite{Wang21} introduced the idea of learning upper and lower bounds for each continuous feature, noting that in the following layers these bounds could be combined into intervals (AND) and then arbitrary collections of such intervals (OR). Since our framework can take advantage of negation, only upper bounds $\mathcal{B}_{i,k} \in \mathbb{R}$ are needed, and we additionally learn a sharpness parameter $\alpha_{i,k}>0$ that controls how sharp or fuzzy is the transition at the boundary, resulting in a fuzzy discretization. We call these concepts \emph{fuzzy dichotomies}, defined for a continuous feature $x_k$ by
\begin{align*}
    \sigma\Big(\alpha_{i,k}(x_k - \mathcal{B}_{i,k})\Big), \tag{FD}
\end{align*}
where $\sigma(\cdot)$ is the sigmoid function. For each continuous feature, we use a number of these fuzzy dichotomies which are fed to their own DNF (without unobserved concepts), in order to learn arbitrary collections of fuzzy intervals. For instance, in a task that uses a continuous feature $x_k$ representing \texttt{weight}, one rule might hold only for very light objects or somewhat heavy objects such as $x_k \in [0,0.1]\cup[10,15]$ which would be learned as $(x_k < 0.1) \cup \big((x_k > 10) \cap (x_k < 15)\big)$. The resulting collections of fuzzy intervals are then used as input to the fully-connected DNF (with unobserved concepts) that  can learn the final rules with all the features. The fuzzy dichotomies are learned conjointly with all the AND/OR concepts in the NLN. These fuzzy dichotomies should not be confused with prior notions of ``dichotomies'' in fuzzy logic, such as in the clustering of a binary classification problem \citep{Ruspini70} or in measuring the ``dichotomousness'' of a fuzzy set \citep{Kitainik87}.

\subsubsection{Input encodings and rule modules}

\begin{figure}[t]
    \centering
     \begin{subfigure}[b]{0.36\textwidth}
         \centering
         \includegraphics[width=\textwidth]{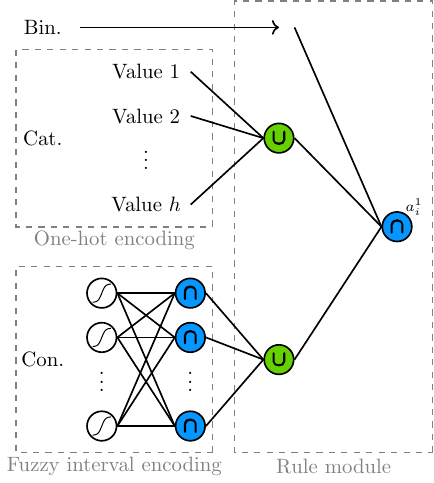}
         \caption{Input encodings and rule modules}
         \label{fig:NLN_rule_module_rule_module}
     \end{subfigure}
     \qquad\qquad
     \begin{subfigure}[b]{0.5\textwidth}
         \centering
         \includegraphics[width=\textwidth]{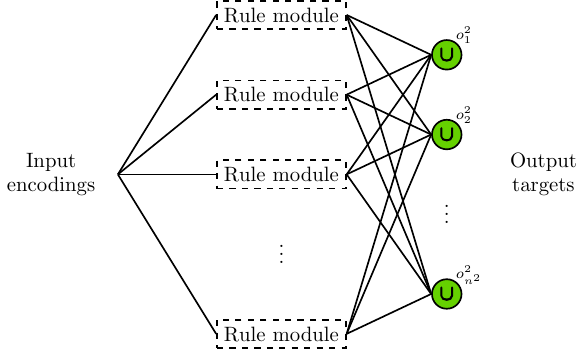}
         \vspace{.6cm}
         \caption{NLN with input encodings and rule modules}
         \label{fig:NLN_rule_module_NLN}
     \end{subfigure}
    \caption{NLN structure used for learning. The (AND) rules are combined with its own pre-processed representations into independent rule modules that learn on the same input encodings: for categorical features, the classical one-hot encoding; and, for continuous features, the fuzzy interval encoding made up of (AND) intervals of fuzzy dichotomies.}
    \label{fig:NLN_rule_module}
\end{figure}

To help the learning process, we do not learn the NLN directly with the structure in Figure \ref{fig:nln_structure}. We instead disentangle the learning of each AND rule by introducing separate rule modules and shared input encodings that are used by all rule modules, pictured in Figures \ref{fig:NLN_rule_module}(a) and \ref{fig:NLN_rule_module}(b).
Each rule module contains a single AND rule that takes its inputs from (1) the binary features, (2) its own OR equivalency classes, one for each categorical feature, and (3) its own OR collections of fuzzy intervals, one for each continuous feature. In turn, binary features are used directly, but categorical features have a shared one-hot encoding, and continuous features are encoded with shared fuzzy dichotomies and AND fuzzy intervals. This factorization of the structure allows each rule to learn more independently of the others and reduces the number of parameters in the model. However, the fuzzy interval encodings of the continuous features are still learned conjointly for the whole NLN. For categorical features, this factorization is equivalent in terms of representability. Since the one-hot encoding is binary, an AND rule that combines multiple OR equivalency classes for the same categorical feature can always be rewritten with a single OR equivalency class when the weights are integers. The same is not true for the fuzzy interval encoding of continuous features since they are never binary. However, in terms of interpretability, a single OR collection of fuzzy intervals is much simpler to interpret in an AND rule, which may already involve many other features.

\subsection{Learning}
\label{sec:ML_learning}

Learning a NLN is done in two stages: (1) training, and (2) post-processing, which includes weight discretization, continuous parameter retraining, and pruning.

\subsubsection{Training}
\label{sec:ML_train}

NLNs are trained with typical gradient descent algorithms from a random initialization with an additional rule reset mechanism.

\paragraph{Objective and regularization}

We use the ADAM optimizer \citep{Diederik15} to minimize the $L_2$ loss. Its minimizer is $\mathbb{P}\!\left[\omega \in Y_k \mid x \right]$ which is precisely what we want our NLN's outputs $c_k^2(x) = \mathbb{P}\!\left[\omega \in C_k^2 \mid x \right]$ to model. To help the learning process we regularize NLNs in two different ways. First, to combat the tendency of unadapted concepts to become trivial, we regularize the AND and OR concepts to have non-empty definitions. For instance, we consider the definition of an AND concept $C_{i}^l$ to be non-empty if
\begin{align*}
    \displaystyle\sum_{j} \left|A_{i,j}^l\right| \geq 1,
\end{align*}
in other words, we consider a concept $C_{i}^l$ to be non-empty if it attributes a probability mass of at least 1 across all of its possible input concepts $C_{j}^{l\shortminus1}$. We force non-empty definitions in all AND and OR concepts by penalizing
\begin{align*}
    \mathcal{L}_{\text{non-empty}} = \sum_{\textrm{AND weights } A_{i,\cdot}^l} \left\|\left[1 - \displaystyle\sum_{j} \left|A_{i,j}^l\right|\right]_{\!\!+\,}\right\|_2^2 + \sum_{\textrm{OR weights } O_{i, \cdot}^l} \left\|\left[1 - \displaystyle\sum_{j} \left|O_{i,j}^l\right|\right]_{\!\!+\,}\right\|_2^2,
\end{align*}
which is only active when a concept $C_{i}^l$'s definition attributes a probability mass less than 1. In that case, the penalty will increase the weights of all of its input concepts $C_{j}^{l\shortminus1}$ uniformly until a probability mass of at least 1 is attributed. Moreover, in order to encourage sparser, more interpretable solutions, we also penalize the $L_1$ norm of all weights in the network. The full loss function is then given by
% \begin{align*}
%     \mathcal{L}(y, c^L(x)) = \left\|y - c^L(x)\right\|_2^2 + \lambda_{\text{non-empty}} \cdot \mathcal{L}_{\text{non-empty}} + \lambda_{\text{sparsity}} \left( \sum_{\textrm{AND weights } A_{i, \cdot}^l} \left\| A_{i, \cdot}^l\right\|_1 + \sum_{\textrm{OR weights } O_{i, \cdot}^l} \left\| O_{i, \cdot}^l\right\|_1\right),
% \end{align*}

\nopagebreak
\hspace{0.075cm}\scalebox{0.94}{\parbox{1.\textwidth}{%
\begin{align*}
    \mathcal{L}(y, c^L(x)) = \left\|y - c^L(x)\right\|_2^2 + \lambda_{\text{non-empty}} \cdot \mathcal{L}_{\text{non-empty}} + \lambda_{\text{sparsity}} \left( \sum_{\textrm{AND weights } A_{i, \cdot}^l} \left\| A_{i, \cdot}^l\right\|_1 + \sum_{\textrm{OR weights } O_{i, \cdot}^l} \left\| O_{i, \cdot}^l\right\|_1\right),
\end{align*}
}}

\noindent where $\lambda_{\text{non-empty}}, \lambda_{\text{sparsity}}  > 0$ are the regularization coefficients of the non-empty penalty and the sparsity penalty respectively. We minimize the expectation of this loss
\begin{align*}
    \mathbb{E}_{(X,Y) \sim \mathcal{D}}\left[\mathcal{L}(Y, c^L(X))\right]
\end{align*}
over the training data set $\mathcal{D}$, subject to the domain constraints of the weights $A_{i,j}^l \in [-1,1]$, $O_{i',j'}^{l'} \in [0,1]$, the biases $a_i^l, o_{i'}^{l'} \in [0,1]$, and the parameters of the fuzzy dichotomies $\mathcal{B}_{i'',k} \in \mathbb{R}$, $\alpha_{i'',k} > 0$ for appropriate indices $(i,j,l)$, $(i',j',l')$ and $(i'',k)$ according to the NLN's structure.

\paragraph{Initialization}

\begin{figure}[h!]
    \centering
     \begin{subfigure}[c]{0.5\textwidth}
         \centering
         \includegraphics[width=\textwidth]{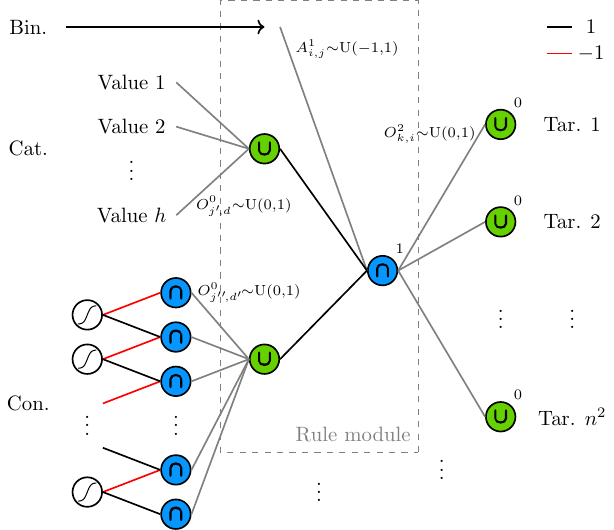}
         \caption{Rule module and input encoding initialization}
         \label{fig:init_NLN}
     \end{subfigure}
     \qquad\qquad
     \begin{subfigure}[c]{0.36\textwidth}
         \centering
         \includegraphics[width=\textwidth]{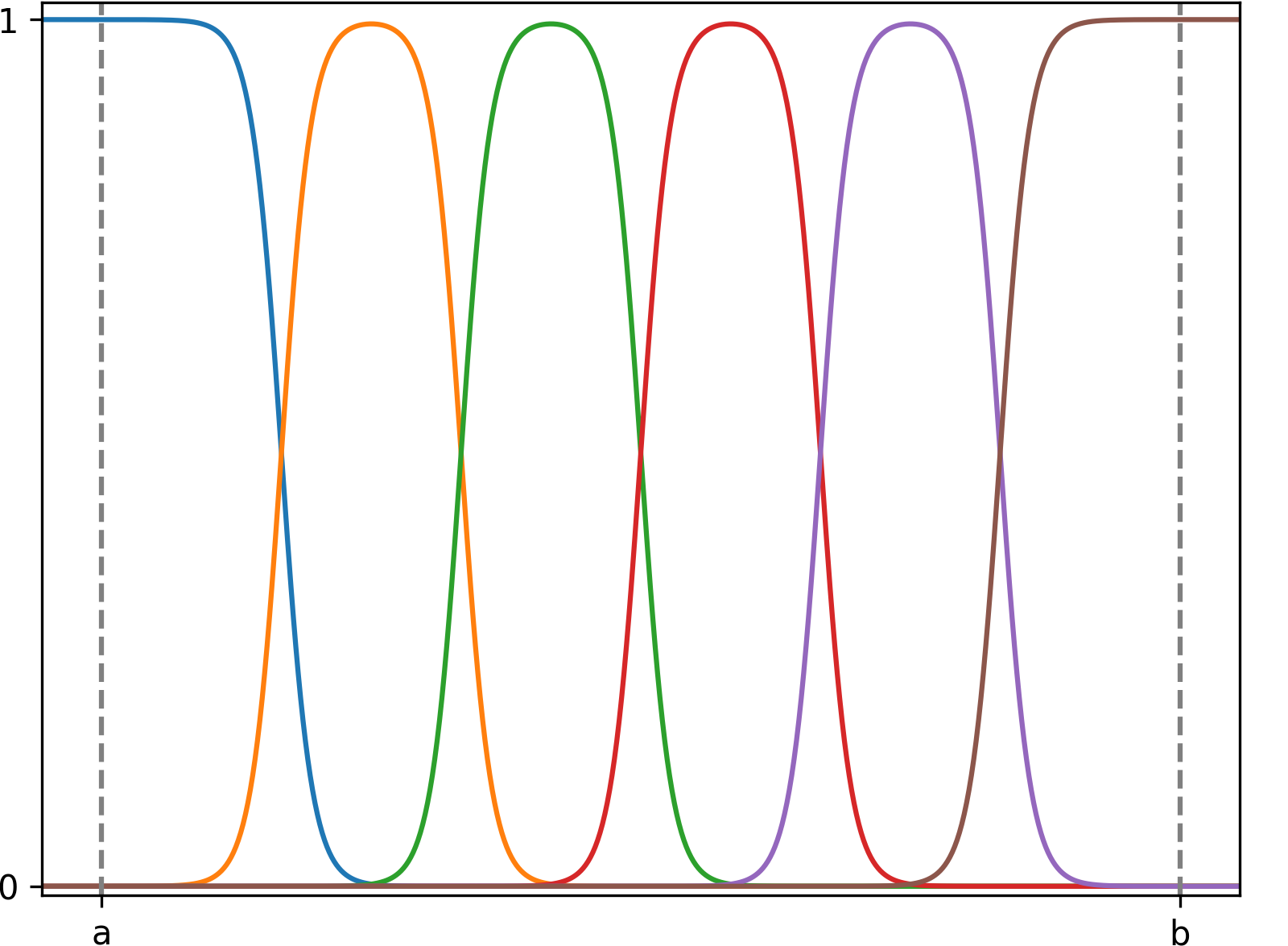}
         \caption{Fuzzy interval encoding initialization for an example with 5 fuzzy dichotomies}
         \label{fig:init_con}
     \end{subfigure}
    \caption{NLN initialization. The weights are sampled from a uniform distribution on their range of possible values; the biases model no effect from unobserved concepts; the fuzzy interval encoding are regularly distributed intervals over the known range of the associated continuous feature.}
    \label{fig:NLN_initialization}
\end{figure}

In our experimentations, the initialization of the NLN was a very important factor in its ability to learn. We have found the best combination to have (1) uniformly random weights, (2) fully observed concepts, and (3) regularly distributed fuzzy interval encodings for continuous features, pictured in Figure \ref{fig:NLN_initialization}. The random weights increase our chances to find potential rules that can be further massaged towards relevant rules, with respect to the target concepts. We begin with full binary biases $a_i^1 = 1$ and $o_j^2 = 0$, \ie without unobserved effects. These ensure that the initial gradients are as strong as possible since, in general, their magnitudes are proportional to $a_i^l$ for AND nodes and to $(1-o_i^l)$ for OR nodes. This is especially important to combat the effect of the vanishing gradients. For binary features and categorical features with one-hot encoding, the random weights can be learned because they receive clean 0-1 signal that is also interpretable. In order to obtain a similarly clean and interpretable signal from continuous features, we initialize the fuzzy interval encoding to regularly distributed intervals with appropriately scaled sharpness. This way, an input $x$ will initially only activate a single fuzzy interval per continuous feature, hence producing a clean, interpretable signal.

\paragraph{Rule reset}

In practice, we have observed that rules that are not helpful in the model quickly become ``dead'' concepts. For such an AND rule $C_i^1$, either its bias $a_i^1$ goes to $0$ or their weights $O_{\cdot,i}^2$ in the next layer all go to $0$. In both cases, they stop receiving signal in the back-propagation and stop learning. We solve this issue by re-initializing dead rule modules. To do so, we randomly re-initialize their bias to $a_i^1 = 1$ and their weights to $A_{i,j}^1 \sim \textrm{U}(-1,1)$ for the binary features $j$ and $O_{\cdot, j'}^0 \sim \textrm{U}(0,1)$ for the categorical/continuous features $j'$. To ensure that the resets do not affect the loss, we also set their outgoing weights in the output layer to $O_{\cdot,i}^2 = 0$. We do this at the end of every epoch by checking for dead rule modules and re-initializing them. By having a big number of rule modules in the network, we can try many random rules at each epoch and keep only those that have potential to be learned further.

\subsubsection{Post-processing}
\label{sec:ML_post}

After training, NLNs are post-processed to obtain an interpretable form as a simple set of rules. This post-processing includes weight discretization, followed by retraining of the continuous parameters, pruning of the weights and bias adjustment.

\paragraph{Weight discretization}

At this point, the NLN has been learned but its weights are still probabilities which are hard to interpret, especially in conjunction with one another. For instance, a simple AND rule that would have weights of $(0.80, -0.65, 0.15)$ for respectively \texttt{ball}, \texttt{green} and \texttt{heavy} is difficult to interpret. It represents a concept that is likely a type of ball, but not necessarily; a concept that is probably not a green object; and a concept that might be a heavy object although it is unlikely; all simultaneously. This is not easily interpretable, unlike the same weights after \emph{discretization} which might be $(1,-1,0)$ that would represent the concept of a \texttt{ball\_that\_is\_not\_green}. By discretizing the weights to values of either $0$, $1$ or $-1$, we obtain instantly understandable concepts that still retain a probabilistic bias in $[0,1]$, indicating if we are missing other unobserved concepts in its definition and how often they appear. To discretize, previous methods would either threshold the weights at above or below $0.5$ \citep{Wang20, Zhang23} or use a modified learning algorithm to learn the discretized weights directly \citep{Wang21, Wang24}. We have instead developed 4 simple greedy algorithms to discretize the weights one at a time. The weights are discretized according to their effect on the loss of the full training set (including the validation set). \change{One of} %Our experimentations suggest that 
our most effective and reliable approach is Algorithm 1. Our other discretization algorithms are presented in appendix \ref{sec:app_ML_post_discretiz}.

\begin{figure}[h!]
    \centering
\begin{algorithm}[H]
\SetKwIF{If}{ElseIf}{Else}{If}{$\!$,}{Else if}{Else}{}%
\SetKwFor{For}{For}{$\!$,}{}%
\SetKwFor{While}{While}{$\!$}{}%
\SetAlgoVlined
\DontPrintSemicolon
\For{{\upshape each layer }$l \in \{1,..,L\}${\upshape , starting from the last layer $L$}}{
    \For{{\upshape each weight }$A_{i,j}^l${\upshape ~(resp. }$O_{i,j}^l${\upshape ), in decreasing likeliness }$\left|A_{i,j}^l\right|${\upshape ~(resp. }$\left|O_{i,j}^l\right|${\upshape )}}{
        \If{{\upshape it is non-zero}}{
            Compare the loss when we fix $A_{i,j}^l \in \left\{0,\, \textrm{sign}\left(A_{i,j}^l\right)\right\}$ (resp. $O_{i,j}^l \in \left\{0,\, \textrm{sign}\left(O_{i,j}^l\right)\right\}$).\;
            Commit to the best discretized value.\;
        }
    }
}
Do the same for the category and continuous input modules, one at a time.\;
\end{algorithm}
    \caption*{Algorithm 1: Descending selection discretization algorithm}
    \label{fig:enter-label}
\end{figure}

\paragraph{Continuous parameter retraining}

Once the weights are discretized, we retrain briefly the continuous parameters of the model with respect to the new weights. These continuous parameters are the model's parameters which are interpretable for continuous values, \ie the biases $a_{i,j}^l, o_{i,j}^l \in [0,1]$ as well as the boundaries in $\mathcal{B}_{i,k} \in \mathbb{R}$ and sharpnesses in $\alpha_{i,k} \in \mathbb{R}_+^*$ of the fuzzy dichotomies used to pre-process the continuous features $x_k$.

\paragraph{Pruning}

Finally, inspired by \cite{Payani19b}, we prune unnecessary weights and simplify the NLN accordingly. To do a pruning pass, starting from the output layer, we consider pruning each weight one at a time and, if it improves or does not affect the loss on the full training set (including the validation set), we permanently prune the weight and, otherwise, we restore its previous value. We keep doing pruning passes until the NLN stops changing.

\paragraph{Data set coverage analysis and bias adjustment}

At this point, the rules of each output targets are finalized. Hence, we can analyze over the full training data set which data points are covered by which rule and whether these rules classify them correctly. This analysis is interesting by itself, but it can furthermore be used to adjust the final biases. Indeed, since the biases of the rules $a_{\cdot}^1$ and of the output targets $o_{\cdot}^2$ have probabilistic definitions, they can be estimated statistically with this analysis. The rule biases $a_{\cdot}^1$ represent the probability that the associated output target is true when its observed definition is fulfilled. This is exactly the proportion of data points covered by this rule that are correctly classified. The target output biases $o_{\cdot}^2$ represent the probability that an output target is true when none of its rules are activated. This is exactly the proportion of data points that are not classified as this target that should be. However, these simple proportions are complicated by the fact that the activation of a rule or classification as a target is actually a probability of presence $c_i^l(x) \in [0,1]$. To address this, we estimate these proportions as the proportions of their respective total probability mass, summed over the full training data set $\mathcal{D}$. The rule bias $a_i^1$ of a rule $C_i^1$ associated to output target $C_k^2$, and its own output target bias $o_k^2$ are respectively adjusted to
\begin{align*}
    a_i^1 = \frac{\displaystyle \sum_{\substack{(x,y) \in \mathcal{D}:\\y_k = 1}} \tilde{c}_i^1(x)}{\displaystyle \sum_{(x,y) \in \mathcal{D}} \tilde{c}_i^1(x)}, \qquad \qquad\qquad\qquad o_k^2 = \frac{\displaystyle \sum_{\substack{(x,y) \in \mathcal{D}:\\y_k = 1}} \Big( 1 - \tilde{c}_k^2(x) \Big)}{\displaystyle \sum_{(x,y) \in \mathcal{D}} \Big( 1 - \tilde{c}_k^2(x) \Big)},
\end{align*}
where $\tilde{c}_i^1(x)$ and $\tilde{c}_k^2(x)$ are respectively the probabilities of the AND node $C_i^1$ and OR node $C_k^2$ without the contribution of their biases $a_i^1$ and $o_k^2$. Since the rule biases $a_{\cdot}^1$ affect the output targets observed probabilities $\tilde{c}_k^2(x)$ used in the computation of the output targets biases $o_{\cdot}^2$, biases are adjusted for rules before output targets. A similar data set coverage analysis is then finally used to eliminate redundant rules, in terms of coverage inclusion (see appendix \ref{sec:app_ML_post_inclusion} for details).

\paragraph{Missing values}

We deal with missing values in a simple manner. In the case of categorical features, we consider an additional value \texttt{?} with index $j$ to which we assign weights $O_{\cdot,j}^0$ like every other categorical value. These special weights are trained just like all the other weights, but they are not discretized since we use them to model a probability. Instead, they are retrained with the biases and fuzzy dichotomies. For each rule, this weight learns the probability that a missing value was in fact one of the categorical values that make up the rule. The case of binary features with missing values are then viewed as categorical features with missing values. Finally, continuous features with missing values are treated analogously with additional weights $O_{\cdot,j}^0$ for the missing values that are added to each OR node. Again, for each rule, this weight learns the probability that a missing value was in fact part of the collection of fuzzy intervals that make up the rule. In previous approaches to NLNs \citep{Payani19, Payani20, Wang20, Wang21, Wang24, Zhang23}, missing values required data pre-processing to artificially replace them with the average or most probable values. Our natural extension can instead work directly with the missing values and not lose any information. In fact, it can learn additional information about the probabilistic distribution of these missing values.

\section{Experiments}
\label{sec:exp}

We evaluate our approach on two different tasks: discovery of boolean networks, and tabular classification, \ie classification from structured data. In both cases we use a NLN with 128 rule modules with regularization coefficients of $\lambda_{\text{non-empty}} = 10^{-1}$ for the non-empty definitions regularization and of $\lambda_{\text{sparsity}} = 10^{-3}$ for the sparsity regularization on the $L_1$ norm of the weights. In cases of multi-class classification, the target class with the highest probability is outputted. In binary or multi-label classification, the threshold with the highest score is used for the binary output (grid search with $0.01$ step). We implement our method in Pytorch \citep{pytorch} using the ADAM optimizer.

\subsection{Boolean networks discovery}

Our method improves the state-of-the-art in Boolean networks discovery. With as little as 16 \% of the data, our method achieves over 97 \% prediction accuracy in all four considered data sets. Moreover, with only 40 \% of the data, it achieves perfect accuracy and correctly identifies the rules of the ground-truth boolean networks. See appendix \ref{sec:app_exp_bool} for details.

\subsection{Interpretable tabular classification}

We test the more general case of classification from tabular data on 7 UCI data sets, 6 of which are often used to test model interpretability. We add an eighth data set by converting the continuous features of the balance data set to categorical features, since they only take 5 possible values. These represent two distinct types of data sets: those that can be represented by a ground-truth logic program (tic-tac-toe \citep{UCItictactoe}, chess KRKPA7 \citep{UCIchess}, and Monk's 2nd problem \citep{UCImonk2}), and those that cannot (chronic kidney disease \citep{UCIckd}, wine \citep{UCIwine}, adult \citep{UCIadult}, and balance \citep{UCIbalance}). However, while balance does not have an easily interpretable logical representation, its ground-truth is still very interpretable. It represents  a balance scale which leans to one side or is in balance depending on which side has the highest product $\texttt{weight}\cdot\texttt{distance}$. However, such a type of rule cannot be encoded in a NLN without trivally encoding all 625 possible cases. This data set exemplifies a limit to what types of interpretable structures a NLN can discover. The characteristics of these data sets are presented in Table \ref{tab:tab_charac}. Since these data sets are mostly unbalanced, we use the F1 score to evaluate the prediction performance of the models with five-fold cross-validation.

\begin{table}[h!]
    \centering
    \caption{Data set characteristics and logic network model capacities (number of Pre-Processing nodes (PP), of Fully-Connected nodes (FC) and Total number of Parameters (TP))}
    \label{tab:tab_charac}
    \scalebox{0.8}{\begin{tabular}{lccccc|cccccc}
        \toprule
        & \multicolumn{3}{c}{Inputs} & Outputs & & \multicolumn{3}{c}{\textbf{NLN}} & \multicolumn{3}{c}{RRL} \\ \cmidrule(r){2-4} \cmidrule(r){5-5} \cmidrule(r){7-9} \cmidrule(r){10-12}
        Data sets & Bin. & Cat. & Con. & Classes & Samples & PP & FC & TP & PP & FC & TP\\
        \midrule
        tic-tac-toe & 0 & 9 & 0 & 2 & 958 & 1\,152 & 128, 1 & 4\,865 & 0 & 1024, 2 & 28\,672\\
        chess & 35 & 1 & 0 & 2 & 3\,196 & 128 & 128, 1 & 5\,249 & 0 & 1024, 2 & 39\,936\\
        monk2 & 0 & 6 & 0 & 2 & 432 & 768 & 128, 1 & 3\,201 & 0 & 1024, 2 & 18\,432\\
        \midrule
        kidney & 0 & 13 & 11 & 2 & 400 & 3\,787 & 128, 1 & 68\,513 & 220 & 1024, 2 & 277\,724\\
        wine & 0 & 0 & 13 & 3 & 178 & 2\,509 & 128, 3 & 71\,651 & 260 & 1024, 3 & 268\,036\\
        adult & 1 & 7 & 6 & 2 & 32\,561 & 2\,054 & 128, 1 & 46\,913 & 120 & 4096, 4096, 2 & 17\,686\,648\\
        balance (con.) & 0 & 0 & 4 & 3 & 625 & 772 & 128, 3 & 22\,403 & 80 & 1024, 3 & 83\,536\\
        balance (cat.) & 0 & 4 & 0 & 3 & 625 & 512 & 128, 3 & 3\,587 & 0 & 1024, 3 & 22\,016\\
        \bottomrule
    \end{tabular}}
\end{table}

For this task, we do not expect in general that there are actual logic programs that can predict perfectly these data sets. As such, we opt to split the training set into training (80 \%) and validation (20 \%) sets to select the best model in early stopping, in order to avoid overfitting. In the post-processing phase, we use the full training set including the validation set, which is especially important for these data sets because of their limited size. We compare our approach with two similar models. The first is RRL \citep{Wang24} which uses the same modeling for the AND/OR nodes, with the exception of the missing bias and the need to double the weights to consider negated concepts (see section \ref{sec:theory_fuzzy} for their special case). It uses an approximated version of this modeling however by introducing three hyper-parameters $(\alpha, \beta, \gamma)$ to reduce the vanishing gradients problem inherent to this approach. Its structure is also different with (1) no input pre-processing except for learnable upper and lower bounds for the continuous features, (2) between 1 and 4 logical layers which are each made up of half AND nodes and half OR nodes, and (3) its output layer is linear. It thus introduces many more hyper-parameters to tune than our approach which has none in practice, since we use by default 32 fuzzy dichotomies with 33 fuzzy interval encodings per continuous feature. We follow their instructions to learn the models and tune their hyper-parameters with the final selected structures presented side-by-side with our own in Table \ref{tab:tab_charac}. The final model is the generalization of decision trees, the Optimal Decision Diagram (ODD) \citep{Florio23}. This approach uses a Mixed Integer Programming formulation to find the optimal decision diagram given a data set, a graph topology and a sparsity constraint. We follow their instructions for which topologies and sparsity constraints to test, and select the best performing one. We only consider univariate splits which have maximum interpretability and produce rules of comparable form to NLN and RRL. We also add a final model that is not interpretable to show the level of prediction performance that is attainable for each data set, XGBoost (XGB) \citep{XGB1, XGB2}. The results are presented in Table \ref{tab:tab_res}.

\begin{table}[h!]
    \centering
    \caption{Comparison of five-fold cross-validation f1-score (\%)}
    \label{tab:tab_res}
    \begin{tabular}{l|lll|lllll}
        \toprule
         Models & tic-tac-toe\!\!\! & chess~~~~~\, & monk2~~~\, & kidney~~~ & wine~~~~~~ & adult~~~~~ & bal.(con.)\! & bal.(cat.)\\
        \midrule
        \textbf{NLN} & \textbf{100} & 99.31 & 79.22 & 98.08 & 94.44 & 65.38 & 57.54 & 53.75\\
        RRL & \textbf{100} & \textbf{99.47} & \textbf{95.26} & \textbf{98.91} & 95.58 & \textbf{80.31} & \textbf{73.86} & \textbf{78.84}\\
        ODD & 83.85 & 97.89 & 66.67 & 97.67 & \textbf{95.87} & 60.45 & 53.86 & 53.28\\
        \midrule
        XGB & 99.91 & 99.34 & 87.85 & 99.03 & 98.22 & 70.47 & 66.86 & 66.86\\
        \bottomrule
    \end{tabular}
\end{table}

As expected, we can see that our method performs better in the data sets with underlying ground-truth logic programs (tic-tac-toe, chess and monk2) than those that do not. We can also see that, with the exception of wine, our method has always the best or second-best classification performance among the 3 comparable models. The RRL consistently has the best performance or very close to the best, although this performance comes at the cost of interpretability as we will see later on. On 4 data sets however (tic-tac-toe, chess, kidney and wine), our NLN has a very close performance. On the other data sets, the NLN's performance is 15 to \mbox{25 \%} below the performance of the RRL. The following investigation into interpretability will shed some light on why this is the case. Finally, we can see that the balance data sets in particular are very difficult for these rule-based systems. Only the RRL with its linear output layer is able to achieve a performance well over \mbox{50 \%}. This additional linear combination of the rules can more easily represent decision regions like those of the balance data sets which depend on numerical comparisons.

However, the main advantage of these methods is their interpretability and, by using much more nodes and a linear output layer, the RRL runs the risk of losing this capacity. To compare the interpretability of the models, we use two traditional measures of interpretability for logic programs: the average number of rules and their average rule size. We consider the size of a rule to be the equivalent number of nodes in the input layer that are used, \ie (i) for binary features, 1 if the feature is used and 0 otherwise; (ii) for categorical features, the number of its values that are used (in the one-hot encoding); and (iii) for continuous features, the number of boundary nodes or splits that are used (\ie the number of fuzzy dichotomies in our case). The results are given in Table \ref{tab:tab_res_interp}.

\begin{table}[h!]
    \centering
    \caption{Comparison of interpretability in average number of rules and average rule size}
    \label{tab:tab_res_interp}
    \begin{tabular}{lclllllllll}
        \toprule
         & &  \multicolumn{3}{c}{\textbf{NLN}} & \multicolumn{3}{c}{RRL} & \multicolumn{3}{c}{ODD} \\\cmidrule(r){3-5} \cmidrule(r){6-8} \cmidrule(r){9-11}
        Data sets & (ground truth) & f1 (\%) & nbr. & size & f1 (\%) & nbr. & size & f1 (\%) & nbr. & size \\
        \midrule
        tic-tac-toe & (8 rules of size 3) & \textbf{100} & \textbf{8.0} & 3.00 & \textbf{100} & 313.6 & 2.02 & 83.85 & 29.8 & 4.88 \\
        chess & (? rules) & 99.31 & \textbf{9.2} & 5.56 & \textbf{99.47} & 461.8 & 1.55 & 97.89 & 18.8 & 4.56 \\
        monk2 & (15 rules of size 6) & 79.22 & 18.6 & 5.65 & \textbf{95.26} & 101.0 & 2.07 & 66.67 & \textbf{16.6} & 4.70 \\
        \midrule
        kidney & & 98.08 & 6.8 & 2.49 & \textbf{98.91} & 35.0 & 1.05 & 97.67 & \textbf{3.0} & 1.67\\
        wine & & 94.44 & 29.4 & 5.53 & 95.58 & 38.6 & 1.09 & \textbf{95.87} & \textbf{4.2} & 2.32 \\
        adult & & 65.38 & 69.2 & 22.53 & \textbf{80.31} & 2855.2 & 4.02 & 60.45 & \textbf{5.6} & 2.55 \\
        bal.(con.) & & 57.54 & 94.2 & 12.14 & \textbf{73.86} & 691.6 & 2.61 & 53.86 & \textbf{20.4} & 4.36 \\
        bal.(cat.) & & 53.75 & 71.8 & 5.07 & \textbf{78.84} & 169.2 & 1.26 & 53.28 & \textbf{22.6} & 4.87 \\
        \bottomrule
    \end{tabular}
\end{table}

We can immediately see that, in most cases, RRL's impressive performance requires an uninterpretable amount of rules. In the 3 data sets where NLN and RRL have comparable performance (tic-tac-toe, chess and kidney), the NLN needs less than 30 rules on average while the RRL needs from 5 to 50 times as many. This could be a consequence of the RRL’s linear output layer. By imposing a rigid interpretable structure in NLNs, the learning is more difficult, but the final learned model is directly interpretable. In contrast, the RRL's linear output layer makes the interpretation of its learned rules much less straightforward since it produces hundreds of very small rules, many of them involving a single feature value. This suggests that the actual decision rules learned by the RRL are contained in the distributed representation of its linear output layer, which is much harder to interpret. In comparison to ODD, for a comparable classification performance, the NLN needs fewer rules than ODD for data sets with a ground-truth logic program (tic-tac-toe, chess, monk2) and needs more rules than ODD for data sets that do not. For the two data sets for which the ground-truth logic programs are known (tic-tac-toe, monk2), only the NLN finds rules of roughly the expected size. Table \ref{tab:tab_res_ground_truth} shows the fraction of the ground-truth rules that are correctly discovered by each model, as well as the proportion of excess non-ground-truth rules relative to the actual number of rules in the ground truth.

\begin{table}[h!]
    \centering
    \caption{Comparison of recovered ground-truth in fraction of recovered rules and excess rules proportion}
    \label{tab:tab_res_ground_truth}
    \begin{tabular}{lllll}
        \toprule
         & \multicolumn{2}{c}{tic-tac-toe} & \multicolumn{2}{c}{monk2}\\\cmidrule(r){2-3} \cmidrule(r){4-5}
        Models~~~~~~ & recovered (\%) & excess (\%) & recovered (\%) & excess (\%) \\
        \midrule
        \textbf{NLN} & \textbf{100} & \textbf{0} & \textbf{53.3} & \textbf{24.0}\\
        RRL & 75.0 & 3820.0 & 0 & 573.33 \\
        ODD & 0 & 360.0 & 20.0 & 86.7 \\
        \bottomrule
    \end{tabular}
\end{table}

The NLN has exactly discovered the ground truth on the tic-tac-toe data set and has recovered more than half of the ground-truth rules on monk2. In comparison, RRL with its many hundreds of rules in each data set has not been able to recover all 8 rules in tic-tac-toe or a single rule in monk2. In both cases, it contains over 500 \% of excess rules. Figure \ref{fig:tictactoe} presents networks found by NLN and RRL with perfect classification on tic-tac-toe (for the RRL, the smallest one in the five-fold cross-validation was selected). The NLN network is minimal and describes each of the 8 possible rules that would make $\times$ win, \ie when the $\times$s form any of the 3 rows, 3 columns and 2 diagonals. In contrast, the smallest network found by RRL requires 318 rules, making it very difficult to interpret. ODD on the other hand was not able to recover a single rule in tic-tac-toe and only recovered 20 \% on monk2. In comparison to decision diagrams% (and decision trees)
, the NLN like the RRL learn each rule independently of each other, in the sense that each rule only uses the features that it needs. This is not the case with decision diagrams %(and trees) 
which all begin from the same root split node and share many internal split nodes. This can create rules that are more specific than necessary and makes it harder to recover exactly the ground-truth rules.

\begin{figure}[t]
    \centering
    \parbox[b][.51\textheight][c]{0.495\textwidth}{
        \begin{subfigure}[b]{0.495\textwidth}
             \centering
             \includegraphics[width=\textwidth]{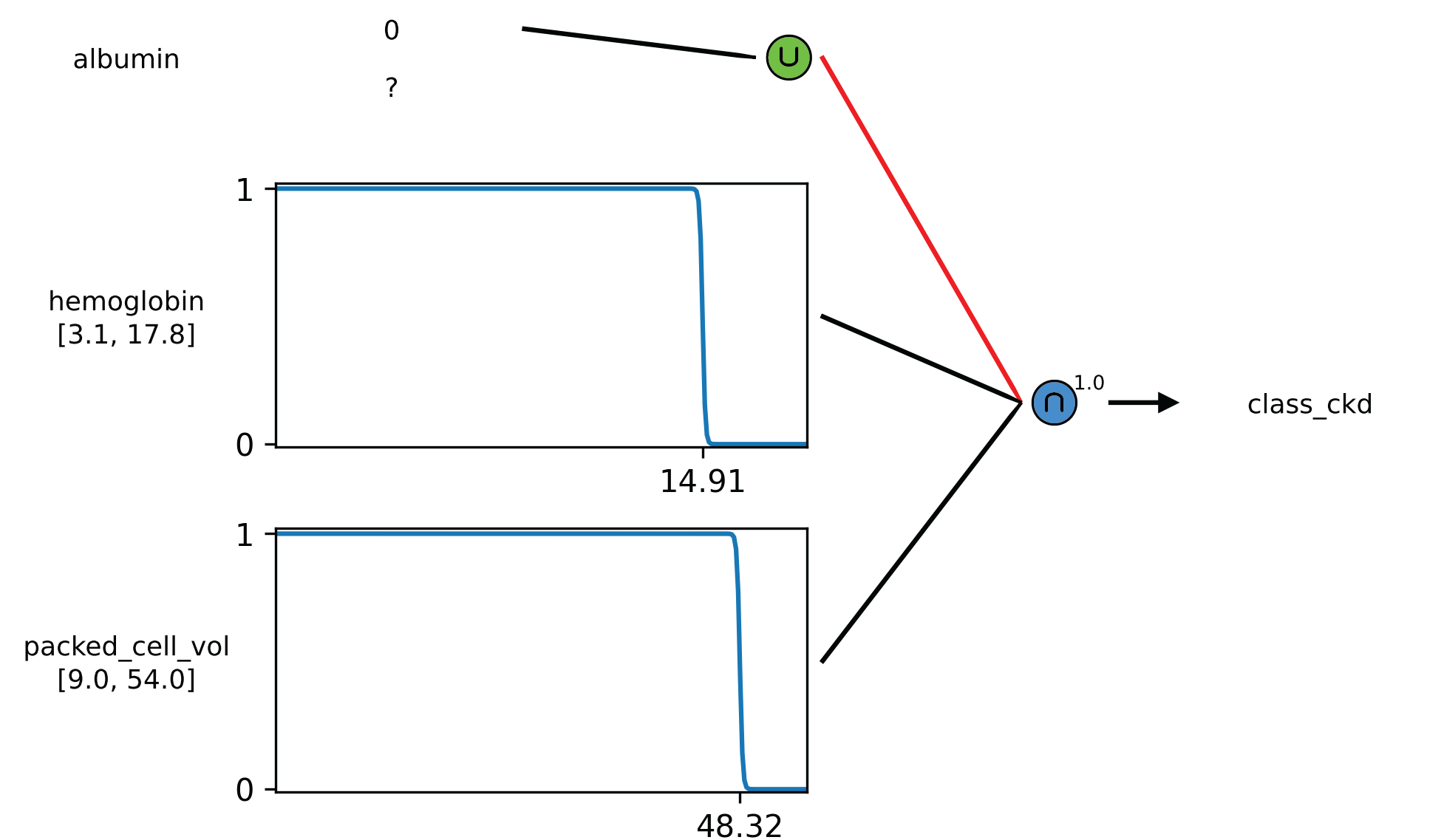}
             \caption{1st rule (covers 76 \% of positive data points)}
             \label{fig:ckd_4rules_1}
        \end{subfigure}

        \vfill
        \begin{subfigure}[b]{0.495\textwidth}
             \centering
             \includegraphics[width=\textwidth]{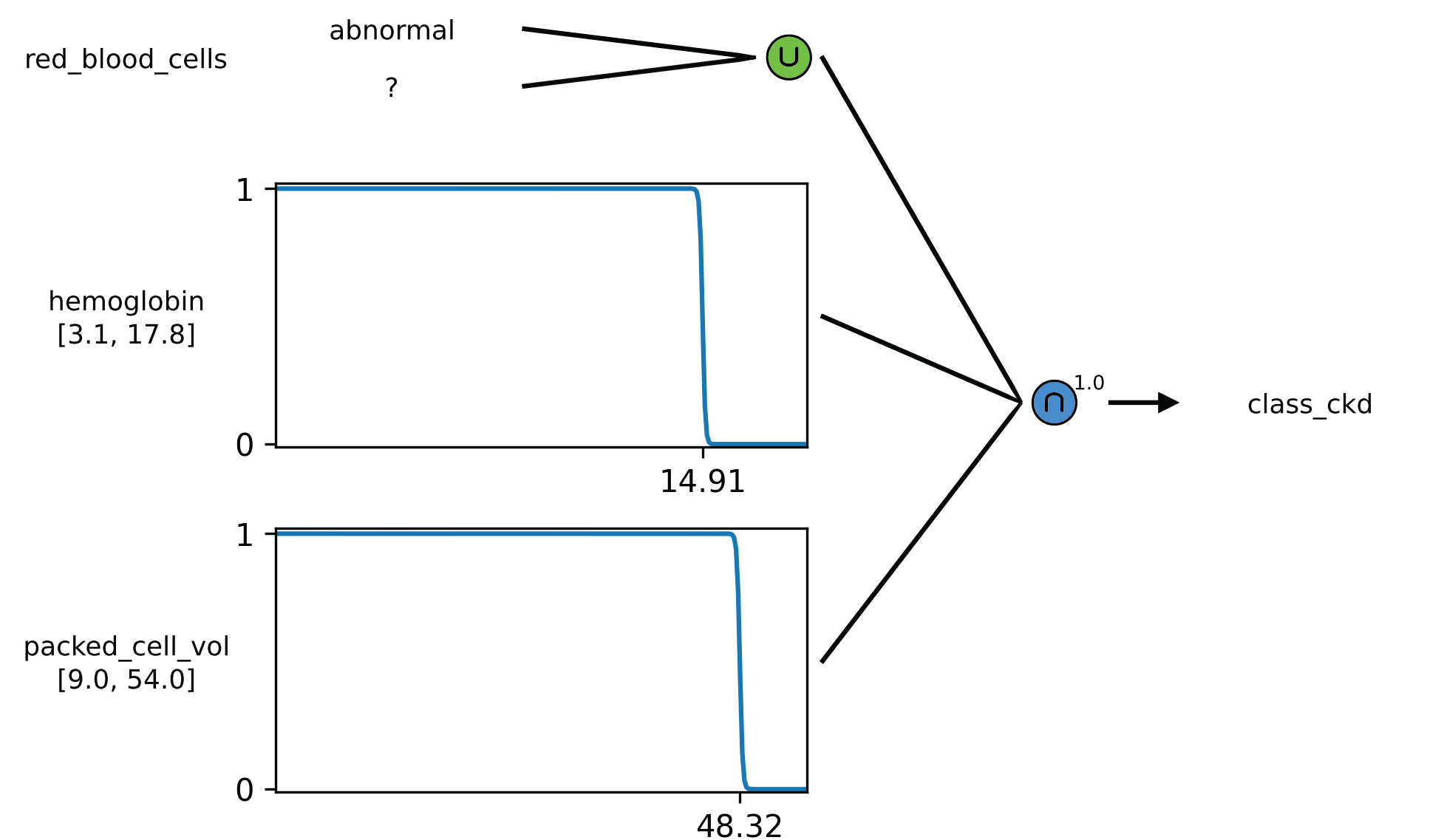}
             \caption{2nd rule (covers 69 \% of positive data points)}
             \label{fig:ckd_4rules_2}
        \end{subfigure}
    }\hfill\parbox[b][.51\textheight][c]{0.48\textwidth}{
        \vfill
        \begin{subfigure}[b]{0.495\textwidth}
             \centering
             \includegraphics[width=\textwidth]{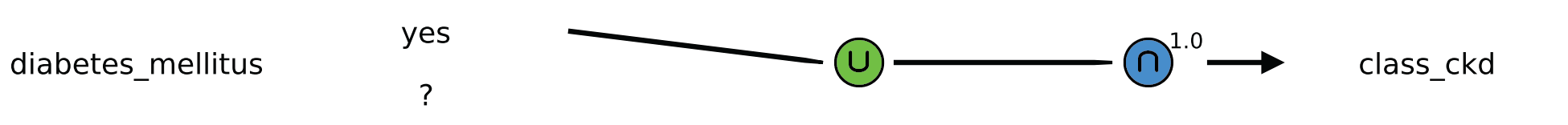}
             \caption{3rd rule (covers 55 \% of positive data points)}
             \label{fig:ckd_4rules_3}
        \end{subfigure}

        \vfill
        \vfill
        \begin{subfigure}[b]{0.495\textwidth}
             \centering
             \includegraphics[width=\textwidth]{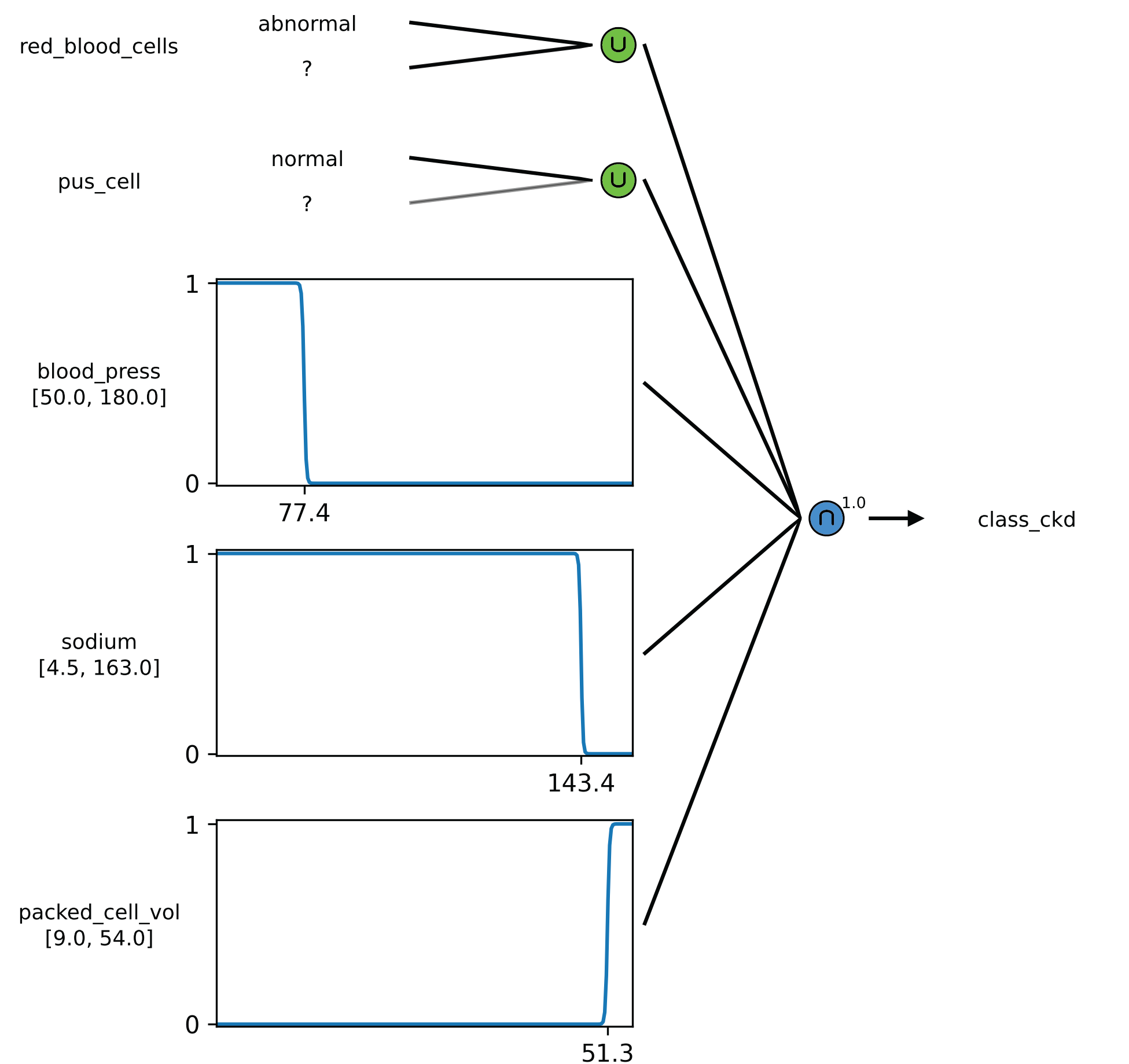}
             \caption{4th rule (covers 8 \% of positive data points)}
             \label{fig:ckd_4rules_4}
        \end{subfigure}
    }
    
    \caption{Merged NLN's 4-rule sub-model with perfect classification on the kidney data set}
    \label{fig:ckd_4rules}
\end{figure}

However, the NLN in its current form still has severe limitations. Its learning is made more difficult by its vanishing gradients. It does not make use of the RRL's approximated formalism with additional hyper-parameters that improve on this issue. The RRL's linear layer is also very helpful to propagate the gradients inside the network, but at the cost of a distributed representation that is not as easily interpretable. Moreover since our search algorithm is stochastic, many different models with equivalent predictive performance can be obtained during learning, resulting in differing sets of rules. For instance, in preliminary tests, a model with perfect accuracy on tic-tac-toe was found that discovered the 3 rows and 3 columns of $\times$, but instead of the 2 diagonals of $\times$, it discovered 2 equivalent rules that each say that, if a diagonal has no $\circ$ anywhere on it, then $\times$ wins. Such equivalent sets of rules can provide additional insight as in this case, although this variability can also be detrimental in other cases where different users may obtain different insights from the same data. Finally, in terms of computing time, the NLN takes much more time than the other two methods, mainly because of the weight discretization step which requires $O\!\left(w\cdot |\mathcal{D}|\right)$ evaluations where $w$ is the number of weights.

\subsubsection{Medical application example}

The kidney data set presents an interesting case of a ML medical diagnosis tool that can support human decision-making. In that setting, interpretability is as much a concern as robust classification performance in order to achieve a trustworthy medical AI \citep{Band23, Ciobanu24}. Interpretable rule-based models like NLNs can be very useful in this context.

Another advantage of rule-based models like NLNs is that they can be merged together. For instance, we have merged the 5 models obtained from the 5-fold cross-validation on this data set into a single model. We have then pruned this merged model again on the full data set to obtain a final model, for which we have again re-adjusted its biases after data set coverage analysis. This final model contains 11 rules and classifies the full data set perfectly. However, not all of its 11 rules are necessary for perfect classification. With only 2 rules, the 2 best sub-combinations of rules obtain an f1 score of 98.99 \% on the full data set. With only 4 rules, the 4 best sub-combinations of rules classify the full data set perfectly. The 4 different sub-combinations differ only on a single rule, which has slightly different boundaries for the continuous features. We have selected the sub-combination with the differing rule that has the highest coverage of positive data points, \ie that covers the most patients with chronic kidney disease. Figure \ref{fig:ckd_4rules} presents its 4 rules individually.

In text form, these 4 rules become

\vspace{-0.2cm}
% \hspace{-0.6cm}\scalebox{0.765}{\parbox{1.\textwidth}{%
% \begin{align*}
%     \textrm{\textbf{IF~}}\texttt{diabetes\_mellitus}=\texttt{yes}\textrm{\textbf{,~}} &\textrm{\textbf{THEN~}}\texttt{class} = \texttt{ckd};\\[1pt]
%     \textrm{\textbf{IF~}}\texttt{albumin}\neq 0\text{ AND }\texttt{hemoglobin}<14.9\textrm{ gms}\text{ AND }\texttt{packed\_cell\_vol}<48.3\textrm{\textbf{,~}} &\textrm{\textbf{THEN~}}\texttt{class} = \texttt{ckd};\\[1pt]
%     \textrm{\textbf{IF~}}\texttt{red\_blood\_cells}=\texttt{abnormal}\text{ AND }\texttt{hemoglobin}<14.9\textrm{ gms}\text{ AND }\texttt{packed\_cell\_vol}<48.3\textrm{\textbf{,~}} &\textrm{\textbf{THEN~}}\texttt{class} = \texttt{ckd};\\[1pt]
%     \textrm{\textbf{IF~}}\texttt{red\_blood\_cells}=\texttt{abnormal}\text{ AND }\texttt{pus\_cell}=\texttt{normal}\text{ AND }\texttt{blood\_press}<77.4\textrm{ mm/Hg} \hphantom{\,\;} &\\
%     {}\text{ AND }\texttt{sodium}<143\textrm{ mEq/L}\text{ AND }\texttt{packed\_cell\_vol}>51.3\textrm{\textbf{,~}} &\textrm{\textbf{THEN~}}\texttt{class} = \texttt{ckd}.\\
% \end{align*}
% }}
\begin{center}
\scalebox{0.8}{\parbox{1.\textwidth}{%
\begin{align*}
    &\textrm{(1)~~~}\textbf{IF~}\,\texttt{albumin}\neq 0\,\textbf{,}\\[-1.5pt]
    &\hphantom{\textrm{(1)~~~}\textbf{IF~}\,}\texttt{hemoglobin}<14.9\textrm{ gms}\,\textbf{,}\\[-1.5pt]
    &\hphantom{\textrm{(1)~~~}\textbf{IF~}\,}\textrm{AND }\,\texttt{packed\_cell\_vol}<48.3\,\textbf{,}&\!\!\!\!\!\!\!\!\!\!\!\!\!\!\!\!\!\!\!\!\!\!\!\!\!\!\!\!\!\!\!\!\!\!\!\!\!\!\!\!\textbf{THEN~}\,\texttt{class} = \texttt{ckd};\hphantom{\texttt{ii}~}\\[7pt]
    &\textrm{(2)~~~}\textbf{IF~}\,\texttt{red\_blood\_cells}=\texttt{abnormal}\,\textbf{,}\\[-1.5pt]
    &\hphantom{\textrm{(2)~~~}\textbf{IF~}\,}\texttt{hemoglobin}<14.9\textrm{ gms}\,\textbf{,}\\[-1.5pt]
    &\hphantom{\textrm{(2)~~~}\textbf{IF~}\,}\textrm{AND }\,\texttt{packed\_cell\_vol}<48.3\,\textbf{,~}\, &\!\!\!\!\!\!\!\!\!\!\!\!\!\!\!\!\!\!\!\!\!\!\!\!\!\!\!\!\!\!\!\!\!\!\!\!\!\!\!\!\textbf{THEN~}\,\texttt{class} = \texttt{ckd};\hphantom{\texttt{ii}~}\\[7pt]
    &\textrm{(3)~~~}\textbf{IF~}\,\texttt{diabetes\_mellitus}=\texttt{yes}\,\textbf{,~} &\!\!\!\!\!\!\!\!\!\!\!\!\!\!\!\!\!\!\!\!\!\!\!\!\!\!\!\!\!\!\!\!\!\!\!\!\!\!\!\!\textbf{THEN~}\,\texttt{class} = \texttt{ckd};\hphantom{\texttt{ii}~}\\[7pt]
    &\textrm{(4)~~~}\textbf{IF~}\,\texttt{red\_blood\_cells}=\texttt{abnormal}\,\textbf{,}\\[-1.5pt]
    &\hphantom{\textrm{(4)~~~}\textbf{IF~}\,}\texttt{pus\_cell}=\texttt{normal}\,\textbf{,}\\[-1.5pt]
    &\hphantom{\textrm{(4)~~~}\textbf{IF~}\,}\texttt{blood\_press}<77.4\textrm{ mm/Hg}\,\textbf{,}\\[-1.5pt]
    &\hphantom{\textrm{(4)~~~}\textbf{IF~}\,}\texttt{sodium}<143\textrm{ mEq/L}\,\textbf{,}\\[-1.5pt]
    &\hphantom{\textrm{(4)~~~}\textbf{IF~}\,}\textrm{AND }\,\texttt{packed\_cell\_vol}>51.3\,\textbf{,~}\, &\!\!\!\!\!\!\!\!\!\!\!\!\!\!\!\!\!\!\!\!\!\!\!\!\!\!\!\!\!\!\!\!\!\!\!\!\!\!\!\!\textbf{THEN~}\,\texttt{class} = \texttt{ckd};\hphantom{\texttt{ii}~}\\[7pt]
    &\textrm{(*)~~~}\textbf{OTHERWISE,~} &\!\!\!\!\!\!\!\!\!\!\!\!\!\!\!\!\!\!\!\!\!\!\!\!\!\!\!\!\!\!\!\!\!\!\!\!\!\!\!\!\textbf{THEN~}\,\texttt{class} = \texttt{no\_ckd}.\\
\end{align*}
}}
\end{center}

\vspace{-0.5cm}
\noindent These rules only use 8 of the 24 input features, providing an example of the automated feature selection that takes place in the discretization and pruning steps of the NLN's learning. However, although these rules classify perfectly the data set, this data set is very small (400 samples) and this perfect classification should not be construed as a proof that these rules are exact. This is especially true given the random nature of how they are learned from the data set. Instead, these rules merely suggest new interesting avenues for research to be explored further by experts in the field.

We followed the same recipe to merge the 5 RRL models into a single model (see details in appendix \ref{sec:app_exp_med_RRL}). The smallest model we found that classified the data set perfectly contained 13 rules, displayed in Figure \ref{fig:ckd_RRL}.

\begin{figure}[h!]
    \centering
    \scalebox{0.8}{\begin{tabular}{ccc}
        \textcolor{red}{$-$0.69} & $\texttt{albumin} \neq 0 \,\text{ OR }\,\texttt{red\_blood\_cells} \neq \texttt{normal}$ & +0.69\\[0.6pt]
        \textcolor{red}{$-$0.50} & $\texttt{red\_blood\_cell\_cnt} \leq 4.73\textrm{ millions/cmm}$ & +0.50\\[0.6pt]
        \textcolor{red}{$-$0.30} & $\texttt{diabetes\_mellitus} = \texttt{yes}$ & +0.31\\[0.6pt]
        \textcolor{red}{$-$0.30} & $\texttt{red\_blood\_cells} \neq \texttt{normal}$ & +0.30\\[0.6pt]
        \textcolor{red}{$-$0.22} & $\texttt{hemoglobin} \leq 13.0\textrm{ gms}$ & +0.22\\[0.6pt]
        +0.24 & $\texttt{hemoglobin} > 12.5\textrm{ gms}$ & \textcolor{red}{$-$0.23}\\[0.6pt]
        +0.29 & $\texttt{albumin} = 0$ & \textcolor{red}{$-$0.29}\\[0.6pt]
        +0.31 & $\texttt{hypertension} \neq \texttt{yes}$ & \textcolor{red}{$-$0.31}\\[0.6pt]
        +0.85 & $\texttt{blood\_gluc\_rand} \leq 160\textrm{ mgs/dl}$ & \textcolor{red}{$-$0.83}\\[0.6pt]
        +1.00 & $\texttt{serum\_creatinine} \leq 1.27\textrm{ mgs/dl}$ & \textcolor{red}{$-$1.00}\\
        \textcolor{gray}{$-$1.14} & & \textcolor{gray}{+1.08}\\\cmidrule(r){1-1} \cmidrule(r){3-3}
        \multicolumn{1}{c}{\texttt{no\_ckd}} & $\stackrel{?}{<}$ & \multicolumn{1}{c}{\texttt{ckd}} \\
    \end{tabular}}
    
    \caption{Merged RRL's 13-rule sub-model with perfect classification on the kidney data set}
    \label{fig:ckd_RRL}
\end{figure}

These rules, like for the NLN, also use only 8 of the 24 input features, albeit a slightly different selection. The biggest difference is that most of these rules are of size 1, representing general individual risk factors with associated weights. In contrast, the NLN finds interactions between the different factors that are always associated to chronic kidney disease. Depending on the use case, the individual risk factors or the interactions between the factors might be more valuable to the end user. The other big difference is that using the RRL model requires a small computation for every new input, while the NLN model only requires to read the rules. However, these two representations of the data set both provide valuable insight about its class distribution, and do so in complementary ways to one another.

Since our discretization, pruning and coverage analysis steps can each remove unnecessary rules, one may wonder why the merged NLN model kept 11 rules when only 4 rules were necessary to classify the data set perfectly. This can happen when continuous features are used or when a binary (resp. categorical) feature is given a probability (resp. probability distribution) as input. In both cases, the corresponding input probabilities of presence $c_i^0(x)$ that are outputted by the pre-processing modules will not be discrete, \ie $c_{i}^0(x) \notin \{0,1\}$. In turn, the NLN's outputs $c_{\cdot}^2(x)$ also will not be discrete in most cases (specifically unless these outputs are activated by a rule that uses purely discrete inputs $c_{i'}^0(x) \in \{0,1\}$), irrespective of the bias values $a_{\cdot}^1$ and $o_{\cdot}^2$. The discretization and pruning steps optimize the NLN's loss $\mathcal{L}$, which at that point is mostly concerned with the $L_2$ distance between its outputs $c_{k}^2(x) \in [0,1]$ and the labels $y_{k} \in \{0,1\}$. When the outputs are systematically not discrete, the NLN will prioritize having more redundant rules than necessary to make the values as close to 1 as possible for positive data points. The coverage analysis's removal of included rules works differently, but, similarly, if a single data point $(x,y) \in \mathcal{D}$ is such that $\tilde{c}_i^1(x) > \tilde{c}_j^1(x)$, then rule $i$ can never be considered included in rule $j$ (see appendix \ref{sec:app_ML_post_inclusion} for details). Especially when continuous features are involved, a slight mismatch of the boundary $\mathcal{B}_{i,k}$ and sharpness $\alpha_{i,k}$ values between different fuzzy dichotomies can easily produce such an exception.

\subsubsection{Industrial application example}

Rule-based models like NLNs are related to Logical Analysis of Data (LAD). LAD is a framework which uses a MILP formulation to find the logic program (in DNF) that best classifies a given data set. LAD has had success in industrial applications such as fault detection and prognosis thanks to the interpretable nature of the rules that they find \citep{Lejeune19}. These rules can then help the experts in the field to understand the root causes of the faults. One difficulty of this framework however is its scalability since its rule generation step depends on solving a NP-hard problem, which becomes a problem for bigger data sets. Methods like NLNs that are based on neural networks could eventually bypass this issue, especially with better training methods and more efficient weight discretization methods.

An example of a typical use of LAD is anomaly detection in cybersecurity. For instance, LAD has been applied \citep{Kumar23} to the NSL-KDD data set that was developed for this setting \citep{Tavallaee09}. This data set aims to predict whether a connection is \texttt{normal} or is an attack (\texttt{anomaly}), depending on the 40 input features that describe its properties (6 binary, 4 categorical and 31 continuous). It contains two data sets, one used for training with 23 different types of attacks and another used for testing with 16 additional unseen types of attacks. It also exists in multiple versions depending on the level of detail in the target labels. The binary version has only two classes: \texttt{normal} or \texttt{anomaly}. A multiclass version also exists with 5 classes, each new class detailing a general category of attack : Denial of Service (\texttt{DoS}), \texttt{Probe}, User to Root (\texttt{U2R}), and Remote to Local (\texttt{R2L}). Their characteristics and class distribution are presented in Table \ref{tab:tab_charac_kdd}.

\begin{table}[h!]
    \centering
    \caption{NSL-KDD data set characteristics and class distribution}
    \label{tab:tab_charac_kdd}
    \begin{tabular}{llllllllll}
        \toprule
        & & & \multicolumn{2}{c}{binary} & \multicolumn{5}{c}{multiclass}\\\cmidrule(r){4-5} \cmidrule(r){6-10}
        Data set & Samples & Attacks & \texttt{normal} & \texttt{anomaly} & \texttt{normal} & \texttt{DoS} & \texttt{Probe} & \texttt{U2R} & \texttt{R2L}\\
        \midrule
        training & $125\,973$ & 23 & 53 \% & 47 \% & 53 \% & 36 \% & 9 \% & 0.04 \% & 0.8 \% \\
        test & $22\,544$ & 39 & 43 \% & 56 \% & 43 \% & 33 \% & 11 \% & 0.9 \% & 12 \% \\
        \bottomrule
    \end{tabular}
\end{table}

We have learned a NLN on the binary training data set, resulting in a network with 24 rules (see appendix \ref{sec:app_exp_KDD} for the rules). In contrast, LAD found 37 rules on this data set. Their classification performance are compared in Table \ref{tab:tab_res_kdd}.

\begin{table}[h!]
    \centering
    \caption{Classification performance after learning on the binary data set and transfer on the multiclass data set (through repruning and readjusting)}
    \label{tab:tab_res_kdd}
    \begin{tabular}{lllllllll}
        \toprule
        & \multicolumn{4}{c}{binary (learned)} & \multicolumn{4}{c}{multiclass (transferred)}\\\cmidrule(r){2-5} \cmidrule(r){6-9}
         &  & \multicolumn{3}{c}{f1 (\%)} &  & \multicolumn{3}{c}{f1 (\%)}\\ \cmidrule(r){3-5} \cmidrule(r){7-9}
        Model & \#rules & training & test & full & \#rules & training & test & full\\
        \midrule
        \textbf{NLN} & \textbf{24} & \textbf{99.73} & 81.37 & \textbf{96.94} & 17 & 88.97 & 60.86 & 84.70 \\
        LAD & 37 & 95.94 & \textbf{84.78} & 94.25 & --- & --- & --- & ---  \\
        \bottomrule
    \end{tabular}
\end{table}

The NLN performs better than LAD on the training data set by about 4 \% with 13 fewer rules, representing the data set with a more accurate and more sparse logic program. However, the NLN also performs worse than LAD on the test data set by about the same amount. This suggests that the additional rules found in LAD help to cover more unseen cases in the test data set. Given, the nature of this test data set with many unseen types of attacks (especially in \texttt{R2L}), we cannot necessarily speak of overfitting here, although it is possible. Also, the comparison in number of rules is not exactly fair since the rules in NLNs are more expressive than the simple ANDs in LAD. The rule modules in NLNs have a richer structure with ORs over categorical values (equivalency classes) and DNFs over continuous boundaries (collections of intervals). For instance, a single rule module in NLN with two equivalency classes of 4 categorical values each would need 16 LAD rules to produce an equivalent logic program.

\begin{figure}[h!]
    \centering
    \begin{subfigure}[b]{\linewidth}
         \centering
         \includegraphics[width=0.8\textwidth]{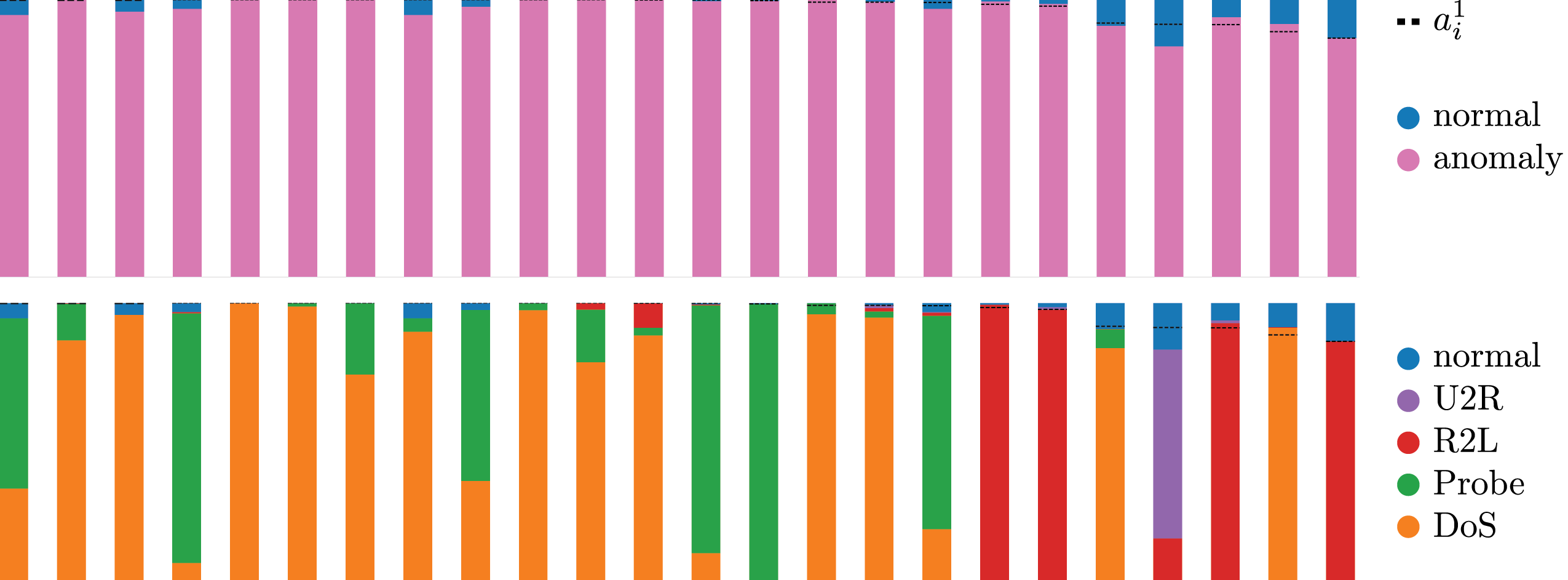}
         \caption{Initial rules learned on the binary training data set}
         \label{fig:kdd_coverage_anomaly}
    \end{subfigure}

    \vspace{12pt}
    \begin{subfigure}[b]{\linewidth}
         \centering
          \includegraphics[width=0.8\textwidth]{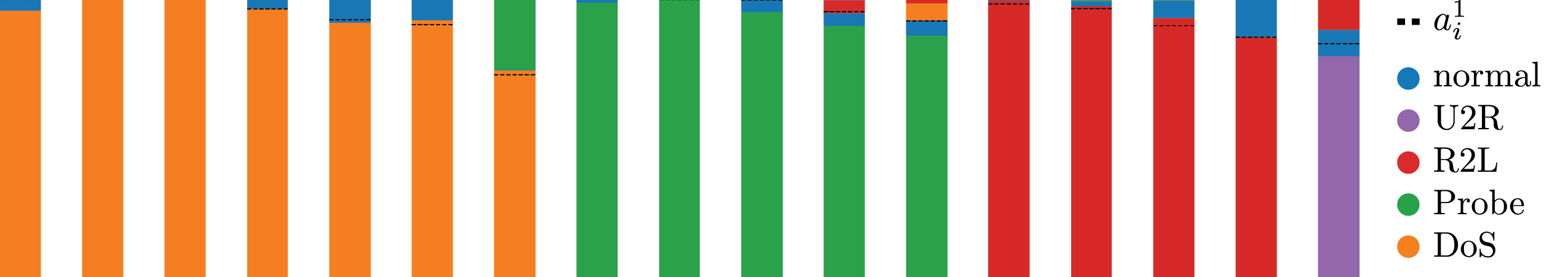}
         \caption{Transferred rules after repruning and readjusting on the multiclass training data set}
         \label{fig:kdd_coverage_subtypes}
    \end{subfigure}
    
    \caption{Breakdown of rule coverage by class on the full data set (training + test)}
    \label{fig:kdd_coverage}
\end{figure}

Another advantage of rule-based neural methods like NLNs is that the rules that we have learned in the binary task can be reused in the multiclass setting. Indeed, by looking at the breakdown of the coverage of each rule in terms of target classes, shown in Figure \ref{fig:kdd_coverage}(a), we can see that many of them cover almost exclusively a single category of attacks. We could thus reuse these rules as a starting point for a new NLN to learn on the multiclass task. Or we could directly transfer these rules to the multiclass task by (1) copying each of them for each of the new target classes; (2) adding an unobserved bias to the \texttt{normal} class with no rules of, for instance, $o_\texttt{normal}^2 = 0.3$; (3) re-pruning the NLN on the multiclass training data set, and (4) re-adjusting all the biases by coverage analysis. By doing this, the NLN keeps only 17 rules, some of which are slightly simplified to accommodate more the selected target class. We can see the coverage of these final rules in Figure \ref{fig:kdd_coverage}(b) and their classification performance in Table \ref{tab:tab_res_kdd}. Having never trained on the multiclass data set, this transferred NLN can still achieve an f1 score of almost 90 \% on the training data set. It does not do as well on the test data set with only about 60 \%, which is not very surprising since this test data set has 16 additional unseen types of attacks and significantly more representation of the \texttt{U2R} and \texttt{R2L} classes.

\subsection{Ablations and statistical validation}

Our weight discretization algorithm beats the previous threshold approach in all datasets, gaining on average around 1 \% of f1 score. However, these gains come at the cost of a very time-intensive algorithm, especially as the model and/or the dataset grows in size. Whether these gains are worth the additional wait time and compute resources will be at the discretion of its users. Our rule reset mechanism improves performance systematically, although this difference is minimal. It is not necessary, but it can help marginally at no significant additional cost in computation. See appendix \ref{sec:app_exp_abl} for details.

\section{Conclusion}

NLNs are a powerful learning and modeling tool for situations that can be described by logic programs, \ie when the output classes/labels $Y_k$ can be described by a set of IF-THEN rules on the input $x$ of the form
\begin{align*}
    \textbf{IF }x_1 \in S_{1,k}\text{ AND }\,...\,\text{ AND }x_d \in S_{d,k}\textbf{, THEN }x \in Y_k.
\end{align*}
By learning to predict the output classes/labels from the input, a NLN can discover this underlying causal structure. Such problems can arise often in practical settings. For instance, a NLN could easily be implemented in a Logical Analysis of Data \citep{Lejeune19} framework for industrial operations and maintenance.

The probabilistic modeling behind NLN's derivation of (P-AND) and (P-OR) supposes three independence assumptions. The first one is between aleatoric and epistemic quantities, \ie specific realizations versus general roles as being necessary or sufficient to other concepts. The second assumption is equivalent to a cognitive bias of ``total open-mindedness'' that considers every combination of concepts to be equiprobable, irrespective of how related or incongruous they might be. The third assumption is an approximation that considers every concept in the same layer to be conditionally independent given the input.

We proposed a factorized rule module structure with pre-processing modules that ensure that the learned rules are easily interpretable to a domain expert. We also proposed a modified learning algorithm with a rule reset scheme to tackle the NLN's notorious vanishing gradient problem. We have evaluated this ML pipeline on tabular classification tasks and, in some cases, the NLNs have successfully learned simple interpretable rules that obtain good classification performance, notably in medical and industrial applications. In one application, we have demonstrated how different NLN models learned on the same data set can be merged together and simplified by pruning and readjusting the biases. The merged model went from the 34 rules of the five initial models to 11 rules, although only 4 rules were necessary to perfectly classify the whole data set. In the other application, we have again used pruning and bias adjustment to transfer a NLN model from a binary data set to a more precise version of the same data set in multiclass form. From the 24 rules of the initial binary model, 17 rules represented cases that were close to being from a single subcategory. In this case, a more thorough approach would have been to use these 24 binary rules as a starting point to learn the multiclass model.

However, in general, our learning strategy does not seem to be sufficient to learn good predictive NLNs, especially in cases where there is no underlying ground-truth logic program. More work needs to be done on this issue to unlock the full modeling capabilities of NLNs. One possible direction is the RRL's modified gradient descent algorithm and approximated AND/OR nodes that introduce hyper-parameters to improve their learning properties. Furthermore, the weight discretization step is currently a severe limitation in terms of computational time and more efficient discretization methods should be explored that do not require as many evaluations.

Interpretable tabular classification is a starting point for NLNs. Future research could explore how they can be adapted to more complex tasks by using different neural network structures. For instance, convolutional NLNs could leverage the AND/OR nodes to tackle interpretable image classification. With convolutional AND kernels and pooling OR layers, these networks could produce interpretable representations of higher-level concepts from 2D arrangements of lower-level concepts. Another example is recurrent NLNs for multi-step reasoning. The tabular NLN presented here can only do single-step reasoning, but by chaining this reasoning through multiple steps, a recurrent NLN could produce multi-step reasoning and solve problems like Sudoku. Again, like in tabular NLN, by learning to predict the finished Sudoku puzzle, the NLN could also discover the rules of Sudoku. Finally, graph NLNs, by working on input graphs with entities and relations, would introduce features of first-order logic by generalizing AND/OR nodes to define universal quantification $\forall$ and existential quantification $\exists$. Moreover, in doing so, graph NLNs could learn to not only predict missing edges or attributes, but also discover underlying relational rules, also known as rule mining in knowledge graphs.

\iftrue
\subsubsection*{Code Availability}
Our implementation is available at \url{https://github.com/VincentPerreault0/NeuralLogicNetworks}.

\subsubsection*{Broader Impact Statement}
This paper presents a new framework for transparent classification. With this framework, a classifier can be learned on any tabular dataset, outputting, for each target class, a set of transparent rules that give a probability of belonging to that class. The rules are transparent in the sense that they directly involve the input features of the dataset, and they mention in which combination of values the rule should be activated.

This can pose a significant risk. Any biases in the dataset will be made transparent in the outputted rules, for instance depicting stereotypical or harmful representations related to protected characteristics like race, religion, gender, and sexuality. As an example, the ``adult'' dataset used in benchmarking aims to classify if an individual has a high income given many attributes, including personal attributes like ethnicity, nationality and gender. Some of the rules that are learned on this outdated dataset from 1995 thus reflect biases in its data in a transparent way. For instance, in some situations, the gender or country of origin will determine whether the individual is classified as having a high income or not. If the owners of such a dataset use a transparent framework like this one to analyze their data, they can discover such discriminatory biases with apparently good classification performance. If these discriminatory rules become the policy (\eg to decide if a loan is granted or not), there would be a significant ethical problem. A way to mitigate this would be to publicly release any such transparent classifier and its learned rules, such that they can be audited by the population it is subjected to. In general, whether the learned rules are discriminatory or not, all the rules should be audited by domain experts. In the best cases, they could lead to an investigation that provides new foundational knowledge on the phenomena in question. In the worst cases, they could prevent the use of rules that are purely the result of spurious correlation, because of a dataset that is either too small or non-representative of its application setting.
% \hl{In this optional section, TMLR encourages authors to discuss possible repercussions of their work, notably any potential negative impact that a user of this research should be aware of. Authors should consult the TMLR Ethics Guidelines available on the TMLR website for guidance on how to approach this subject.}

\subsubsection*{Author Contributions}
Conceptualization: V.P.;~ Formal Analysis: V.P., R.L.;~ Investigation: V.P., K.I., R.L.;~ Methodology: V.P.;~ Resources: A.H.;~ Software: V.P.;~ Supervision: R.L., A.H.;~ Validation: V.P.;~ Visualization: V.P.;~ Writing - original draft: V.P.;~ Writing - review \& editing: All authors.
% \hl{If you'd like to, you may include a section for author contributions as is done in many journals. This is optional and at the discretion of the authors. Only add this information once your submission is accepted and deanonymized.}

\subsubsection*{Acknowledgments}
We thank Lucas Zins, Philippe St-Aubin, Akihiro Takemura, Alex F. Spies, Federico Leuze, Kexin Gu Baugh, Nelson Higuera, and Naïma Francisque for helpful discussions and feedback. Multiple anonymous reviewers provided feedback that significantly improved the manuscript. This work has been supported by JST CREST Grant Number JPMJCR22D3.
% reviewers for enhancing the paper with their comments
%\hl{Use unnumbered third level headings for the acknowledgments. All acknowledgments, including those to funding agencies, go at the end of the paper. Only add this information once your submission is accepted and deanonymized.}
\fi

\bibliography{main}
\bibliographystyle{tmlr}

\newpage
\appendix
% \section{Appendix}
% You may include other additional sections here.

\renewcommand\thesection{\Alph{section}}
\renewcommand\theequation{\thesection.\arabic{equation}}
\renewcommand\thefigure{\thesection.\arabic{figure}}
\renewcommand\thetable{\thesection.\arabic{table}}
\setcounter{section}{0}

\setcounter{equation}{0}
\setcounter{figure}{0}
\setcounter{table}{0}

\section{Symbols and notation}

\begin{table}[h!]
    % \caption{Symbols and notation used throughout the paper}
    \label{tab:symbols}
    \hspace{0.cm}\scalebox{1}{\begin{tabular}{rcp{8cm}}
    \textbf{input data} & $x$ & an input to the network $x=(x_1,...,x_m)$ \\[5pt]
    \textbf{unobserved data} & $u$ & the values of relevant but unobserved variables\\[5pt]
    \textbf{relevant data} & $\omega = (x,u)$ & a complete realization of the modeled random phenomenon \\[15pt]
    \textbf{concept} & $C_i^l$ & concept $i$ of layer $l$\\[5pt]
    \textbf{input concept} & $C_i^0$ & concept $i$ of input layer $0$, known for input $x$\\[5pt]
    \textbf{output concept} & $C_k^L$ & concept $k$ of last layer $L$, models target concept $Y_k$\\[5pt]
    \textbf{layer $l$ $+$} & $\mathcal{C}_{+}^l$ & the  set of all the concepts $C_i^l$ of layer $l$ \\[5pt]
    \textbf{layer $l$ $\pm$} & $\mathcal{C}_{\pm}^l$ & the  set of all the concepts $C_i^l$ of layer $l$ and their opposites $(C_i^l)^c$ \\[5pt]
    \textbf{presence (rand. event)} & $\omega \in C_i^l$ & random event that the concept $C_i^l$ is present in the realization $\omega$ \\[5pt] % the realization 
    \textbf{presence (cond. prob.)} & $c_i^l(x)$ & approximated probability in $[0,1]$ that the concept $C_i^l$ is present in $\omega$ given $x$, \ie $c_i^l(x) \approx \mathbb{P}\!\left[\omega \in C_i^l \,|\, x\right]$ \\[15pt]
    \textbf{necessary concept} & $C \supseteq C_i^l$ & a necessary concept $C$ of AND concept $C_i^l$ \\[5pt]
    \textbf{weights AND} & $A_{i,j}^l$ & if $>0$, probability that $C_j^{l\shortminus 1} \supseteq C_i^l$; \newline if $<0$, $-$probability that $(C_j^{l\shortminus 1})^c \supseteq C_i^l$ \\[5pt]
    \textbf{unobserved necessary concepts} & $u \in \tilde{N}_i^l$ & random event that the unobserved necessary concepts of AND concept $C_i^l$ are present in $\omega$ \\[5pt]
    \textbf{bias AND} & $a_{i}^l$ & probability that the unobserved necessary concepts $\tilde{N}_i^l$ are present when the observed necessary concepts are present \\[15pt]
    \textbf{sufficient concept} & $C \subseteq C_i^l$ & a sufficient concept $C$ of OR concept $C_i^l$ \\[5pt]
    \textbf{weights OR} & $O_{i,j}^l$ & if $>0$, probability that $C_j^{l\shortminus 1} \subseteq C_i^l$; \newline if $<0$, $-$probability that $(C_j^{l\shortminus 1})^c \subseteq C_i^l$ \\[5pt]
    \textbf{unobserved sufficient concepts} & $u \in \tilde{S}_i^l$ & random event that the unobserved sufficient concepts of OR concept $C_i^l$ are present in $\omega$ \\[5pt]
    \textbf{bias OR} & $o_{i}^l$ & probability that the unobserved sufficient concepts $\tilde{S}_i^l$ are present when the observed sufficient concepts are not \\[15pt]
    \textbf{independence} & $A \indep B$ & random events $A$ and $B$ are independent\\[5pt]
    \textbf{set difference} & $A \setminus B$ & set of elements in $A$ that are not in $B$,\newline \ie $A \setminus B = \left\{ a \mid \forall a \in A \textrm{ s.t. } a \notin B \right\}$\\[5pt]
    \textbf{indicator function} & $\mathbbm{1}(\Phi)$ & equals 1 if its argument $\Phi$ is true and 0 otherwise\\[5pt]
    \textbf{positive part} & $\pospart{\lambda}$ & the positive part of $\lambda \in \mathbb{R}$, \ie $\pospart{\lambda} = \max\{0, \lambda\}$\\[5pt]
    \textbf{negative part} & $\negpart{\lambda}$ & the negative part of $\lambda \in \mathbb{R}$, \ie $\negpart{\lambda} = \max\{0, -\lambda\}$\\
\end{tabular}}
\end{table}

\newpage
\section{Theory}

\subsection{Probabilistic modeling}

\subsubsection{Graphical summary of the first two assumptions of independence}
\label{sec:app_theory_prob_PGM}

The first two assumptions of independence that we make in our probabilistic modeling of NLNs are summarized in the PGM below, where we only specify the structure for an AND concept $C_i^l$ since an OR concept would have the same structure.

\begin{figure}[h!]
    \centering
    \includegraphics[width=\textwidth]{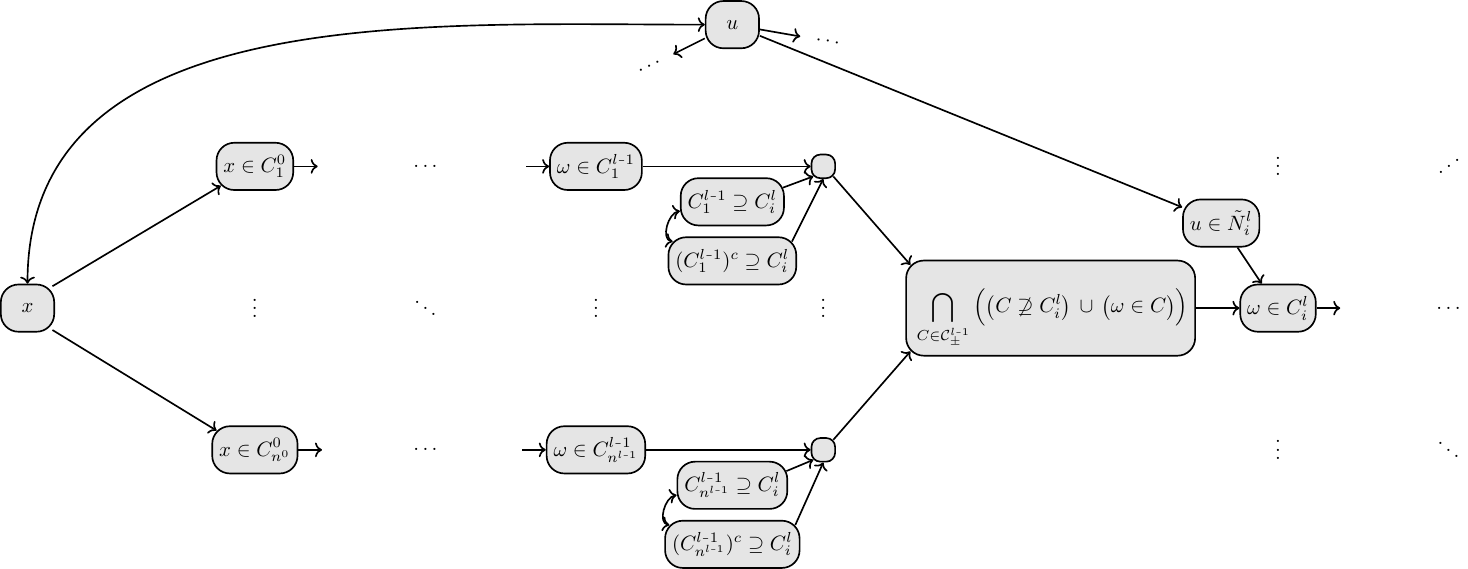}
    \caption{Probabilistic Graphical Model of NLN's probabilistic modeling (rectangles: ran-\newline\hphantom{Figure B.1:} dom variables, arrows: dependencies, such that two sets of random variables\newline\hphantom{Figure B.1:} are conditionally independent with respect to a third set of random variables if\newline\hphantom{Figure B.1:} they are d-separated by it)}
    \label{fig:PGM}
\end{figure}

\begin{itemize}
    \item The observed and unobserved data $x$ and $u$ can be correlated;

    \item The zero-th layer containing the input features $x \in C_i^0$ depends only on the input $x$;

    \item The unobserved necessary/sufficient concepts $u \in \tilde{N}_i^l$ and $u \in \tilde{S}_i^l$ depend only on the unobserved $u$;

    \item \textbf{[1$^\textbf{st}$ assumption]} The role of a concept $C_j^{l\shortminus 1}$ as necessary/sufficient to a concept in the next layer $C_j^{l\shortminus 1} \supseteq C_i^l$ or $C_j^{l\shortminus 1} \subseteq C_i^l$ is independent of every presence of concept $\omega \in C_{i'}^{l'}$ in the layers $l' \leq l - 1$;

    \item \textbf{[2$^\textbf{nd}$ assumption]} The role of a concept $C_j^{l\shortminus 1}$ as necessary/sufficient to a concept in the next layer $C_j^{l\shortminus 1} \supseteq C_i^l$ or $C_j^{l\shortminus 1} \subseteq C_i^l$ is independent of every other roles in the network except for $(C_j^{l\shortminus 1})^c \supseteq C_i^l$ or $(C_j^{l\shortminus 1})^c \subseteq C_i^l$, depending on $C_i^l$'s node type.
\end{itemize}

Critically, we can see that the
\begin{itemize}
    \item \textbf{[3$^\textbf{rd}$ assumption]} The presence of concepts in the same layer $\omega \in C_i^l$ and $\omega \in C_j^l$ (with $i \neq j$) are conditionally independent given the input $x$.
\end{itemize}
is incorrect whenever two concepts share a common necessary/sufficient concept in the previous layers. In that case, this common ``ancestor'' creates a fork that d-connects the two concepts, even when conditioning on $x$.

\subsubsection{Derivation with the 3$^\textrm{rd}$ assumption}
\label{sec:app_theory_prob_deriv_with_3}

\paragraph{AND (conjunction)}
\label{sec:app_prob_indep_AND}

We will derive the following formula for the probability of an AND node $C_i^l$ given $x$

\hspace{0.075cm}\scalebox{0.88}{\parbox{1.\textwidth}{%
\begin{align*}
    \mathbb{P}\!\left[\omega \in C_i^l \mid x\right] = \mathbb{P}\!\left[u \in \tilde{N}_{i}^{l} \mid \omega \in D\right] \cdot \prod_{j=1}^{n^{l\shortminus 1}} \Big( 1 - \mathbb{P}\!\left[C_j^{l\shortminus 1} \supseteq C_i^l\right] \cdot\big(1 - \mathbb{P}\!\left[\omega \in C_j^{l\shortminus 1} \mid x\right] \big) \Big)  \Big( 1 - \mathbb{P}\!\left[(C_j^{l\shortminus 1})^c \supseteq C_i^l\right] \cdot \mathbb{P}\!\left[\omega \in C_j^{l\shortminus 1} \mid x\right] \Big),
\end{align*}
}}

\noindent where we define the observed part of definition
\begin{align*}
    \omega \in D &:= \bigcap_{C \in \mathcal{C}_{\pm}^{l\shortminus 1}} \Big(\big( C \not\supseteq C_i^l \big) \,\cup\, \big(\omega \in C\big) \Big).
\end{align*}

To make this derivation easier to read, we will abuse notation in the following ways
\begin{align*}
    Y &:= C_i^l,\\
    C_j &:= C_{j}^{l\shortminus1}.
\end{align*}
We will also omit the conditioning on $x$ for the same reason. We remind the reader that the roles of necessary concepts $C_j \supseteq Y$ are independent of the input $x$.

We want to compute
\begin{align*}
    \mathbb{P}\left[\omega \in Y\right] = \mathbb{P}\left[(u \in \tilde{N}_{i}^{l}) \,\cap\, \bigcap_{C \in \mathcal{C}_{\pm}^{l\shortminus 1}} \Big(\big( C \not\supseteq Y \big) \,\cup\, \big(\omega \in C\big) \Big)\right].
\end{align*}
Let $D$ be the observed part of the definition
\begin{align*}
    \omega \in D &:= \bigcap_{C \in \mathcal{C}_{\pm}^{l\shortminus 1}} \Big(\big( C \not\supseteq Y \big) \,\cup\, \big(\omega \in C\big) \Big).
\end{align*}
Then $\mathbb{P}\left[\omega \in Y\right]$ can be written
\begin{align*}
    \mathbb{P}\left[\omega \in Y\right] = \mathbb{P}\left[\omega \in D\right] \cdot \mathbb{P}\left[u \in \tilde{N}_{i}^{l} \mid \omega \in D\right],
\end{align*}
and we only need to compute $\mathbb{P}\left[\omega \in D\right]$. We will derive it by induction by considering the contribution of every concept $C_j$ in the previous layer, one at a time. We begin with the law of total probability with respect to the role of $C_1$ or its negation as a necessary concept, of which there are three possible cases
\begin{align*}
    C_1 \supseteq Y,~ (C_1)^c \not\supseteq Y, \qquad \quad (C_1)^c \supseteq Y,~ C_1 \not\supseteq Y, \qquad \quad C_1 \not\supseteq Y,~ (C_1)^c \not\supseteq Y.
\end{align*}
Since we assume that at most one of $C_1$ and $(C_1)^c$ can be a necessary concept, we will note these three cases as
\begin{align*}
    C_1 \supseteq Y,\hphantom{~ (C_1)^c \not\supseteq Y,} \qquad \quad (C_1)^c \supseteq Y,\hphantom{~ C_1 \not\supseteq Y,} \qquad \quad C_1 \not\supseteq Y, (C_1)^c \not\supseteq Y.
\end{align*}
We obtain
\begin{align*}
    \mathbb{P}\left[\omega \in D\right] &= \mathbb{P}\left[C_1 \supseteq Y\right] \cdot \mathbb{P}\left[\omega \in D \mid C_1 \supseteq Y\right]\\[2pt]
    & ~~~~\, + \mathbb{P}\left[(C_1)^c \supseteq Y\right] \cdot \mathbb{P}\left[\omega \in D \mid (C_1)^c \supseteq Y\right]\\[2pt]
    & ~~~~\, + \mathbb{P}\left[ C_1 \not\supseteq Y, (C_1)^c \not\supseteq Y\right] \cdot \mathbb{P}\left[\omega \in D \mid C_1 \not\supseteq Y, (C_1)^c \not\supseteq Y\right].
\end{align*}
Expanding $\mathbb{P}\left[\omega \in D \mid C_1 \supseteq Y\right]$, we get
\begin{align*}
    \mathbb{P}\left[\omega \in D \mid C_1 \supseteq Y\right] &= \mathbb{P}\left[\omega \in C_1\mid C_1 \supseteq Y\right] \cdot \mathbb{P}\left[\omega \in D \mid \omega \in C_1, C_1 \supseteq Y\right] \\[-10pt]
    & ~~~~\, +  \big( 1 - \mathbb{P}\left[\omega \in C_1\mid C_1 \supseteq Y\right] \big) \cdot\cancelto{0}{\mathbb{P}\left[\omega \in D \mid \omega \in (C_1)^c, C_1 \supseteq Y\right]}\\[8pt]
    &= \mathbb{P}\left[\omega \in C_1\right] \cdot \mathbb{P}\left[\omega \in D \mid \omega \in C_1, C_1 \supseteq Y\right]
\end{align*}
where we assume \textbf{[1$^\textbf{st}$ assumption]} independence between the presence of concept $C_1$ and its role
\begin{align*}
\omega \in C_1 \indep C_1 \supseteq Y.
\end{align*}
Expanding $\mathbb{P}\left[\omega \in D \mid (C_1)^c \supseteq Y\right]$, we get
\begin{align*}
    \mathbb{P}\left[\omega \in D \mid (C_1)^c \supseteq Y\right] &= \mathbb{P}\left[\omega \in (C_1)^c\mid (C_1)^c \supseteq Y\right] \cdot \mathbb{P}\left[\omega \in D \mid \omega \in (C_1)^c, (C_1)^c \supseteq Y\right] \\[-10pt]
    & ~~~~\, + \big( 1 - \mathbb{P}\left[\omega \in (C_1)^c\mid (C_1)^c \supseteq Y\right] \big) \cdot\cancelto{0}{\mathbb{P}\left[\omega \in D \mid \omega \in C_1, (C_1)^c \supseteq Y\right]}\\[8pt]
    &= \mathbb{P}\left[\omega \in (C_1)^c\right] \cdot \mathbb{P}\left[\omega \in D \mid \omega \in (C_1)^c, (C_1)^c \supseteq Y\right]
\end{align*}
where we assume \textbf{[1$^\textbf{st}$ assumption]} independence between the presence of concept $C_1$ and its negated role
\begin{align*}
\omega \in (C_1)^c \indep (C_1)^c \supseteq Y.
\end{align*}
Combining the two previous cases, so far we have
\begin{align*}
    \mathbb{P}\left[\omega \in D\right] &= \mathbb{P}\left[C_1 \supseteq Y\right] \cdot \mathbb{P}\left[\omega \in C_1\right] \cdot \mathbb{P}\left[\omega \in D \mid \omega \in C_1, C_1 \supseteq Y\right]\\[2pt]
    & ~~~~\,  + \mathbb{P}\left[(C_1)^c \supseteq Y\right] \cdot \big(1 - \mathbb{P}\left[\omega \in C_1\right]\big) \cdot \mathbb{P}\left[\omega \in D \mid \omega \in (C_1)^c, (C_1)^c \supseteq Y\right]\\[2pt]
    & ~~~~\,  + \mathbb{P}\left[ C_1 \not\supseteq Y, (C_1)^c \not\supseteq Y\right] \cdot \mathbb{P}\left[\omega \in D \mid C_1 \not\supseteq Y, (C_1)^c \not\supseteq Y\right].
\end{align*}
By using the logical definition of $D$, we get

\hspace{0.075cm}\scalebox{0.93}{\parbox{1.\textwidth}{%
\begin{align*}
    \mathbb{P}\left[\omega \in D\right] &= \mathbb{P}\left[C_1 \supseteq Y\right] \cdot \mathbb{P}\left[\omega \in C_1\right] \cdot \mathbb{P}\left[\bigcap_{C \in \mathcal{C}_{\pm}^{l\shortminus 1}} \Big(\big( C \not\supseteq Y \big) \,\cup\, \big(\omega \in C\big) \Big) \mid \omega \in C_1, C_1 \supseteq Y\right]\\ 
    & ~~~~ + \mathbb{P}\left[(C_1)^c \supseteq Y\right] \cdot \big(1 - \mathbb{P}\left[\omega \in C_1\right]\big) \cdot \mathbb{P}\left[\bigcap_{C \in \mathcal{C}_{\pm}^{l\shortminus 1}} \Big(\big( C \not\supseteq Y \big) \,\cup\, \big(\omega \in C\big) \Big) \mid \omega \in (C_1)^c, (C_1)^c \supseteq Y\right]\\
    & ~~~~ + \mathbb{P}\left[ C_1 \not\supseteq Y, (C_1)^c \not\supseteq Y\right] \cdot \mathbb{P}\left[\bigcap_{C \in \mathcal{C}_{\pm}^{l\shortminus 1}} \Big(\big( C \not\supseteq Y \big) \,\cup\, \big(\omega \in C\big) \Big) \mid C_1 \not\supseteq Y, (C_1)^c \not\supseteq Y\right]\\
    &= \mathbb{P}\left[C_1 \supseteq Y\right] \cdot \mathbb{P}\left[\omega \in C_1\right] \cdot \mathbb{P}\left[\bigcap_{C \in (\mathcal{C}_{\pm}^{l\shortminus 1}\setminus \{C_1, \lnot C_1\})} \Big(\big( C \not\supseteq Y \big) \,\cup\, \big(\omega \in C\big) \Big) \mid \omega \in C_1, C_1 \supseteq Y\right]\\ 
    & ~~~~ + \mathbb{P}\left[(C_1)^c \supseteq Y\right] \cdot \big(1 - \mathbb{P}\left[\omega \in C_1\right]\big) \cdot \mathbb{P}\left[\bigcap_{C \in (\mathcal{C}_{\pm}^{l\shortminus 1}\setminus \{C_1, \lnot C_1\})} \Big(\big( C \not\supseteq Y \big) \,\cup\, \big(\omega \in C\big) \Big) \mid \omega \in (C_1)^c, (C_1)^c \supseteq Y\right]\\
    & ~~~~ + \mathbb{P}\left[ C_1 \not\supseteq Y, (C_1)^c \not\supseteq Y\right] \cdot \mathbb{P}\left[\bigcap_{C \in (\mathcal{C}_{\pm}^{l\shortminus 1}\setminus \{C_1, \lnot C_1\})} \Big(\big( C \not\supseteq Y \big) \,\cup\, \big(\omega \in C\big) \Big) \mid C_1 \not\supseteq Y, (C_1)^c \not\supseteq Y\right]
\end{align*}
}}

\noindent If we assume the following independences
\begin{align*}
    C_1 \supseteq Y &\;\indep\; \left.\bigcap_{C \in (\mathcal{C}_{\pm}^{l\shortminus 1}\setminus \{C_1, \lnot C_1\})} \Big(\big( C \not\supseteq Y \big) \,\cup\, \big(\omega \in C\big) \Big)~~\middle|~~ x ,\right.\\
    (C_1)^c \supseteq Y &\;\indep\; \left.\bigcap_{C \in (\mathcal{C}_{\pm}^{l\shortminus 1}\setminus \{C_1, \lnot C_1\})} \Big(\big( C \not\supseteq Y \big) \,\cup\, \big(\omega \in C\big) \Big)~~\middle|~~ x ,\right.\\
    \omega \in C_1 &\;\indep\; \left.\bigcap_{C \in (\mathcal{C}_{\pm}^{l\shortminus 1}\setminus \{C_1, \lnot C_1\})} \Big(\big( C \not\supseteq Y \big) \,\cup\, \big(\omega \in C\big) \Big) ~~\middle|~~ x ,\right. 
\end{align*}
\ie if we additionally assume
\begin{itemize}
    \item \textbf{[1$^\textbf{st}$ assumption]} The role of a concept $C_j$ as necessary to $Y$  ($C_j \supseteq Y$) is independent of every presence of concept $\omega \in C_{j'}$;

    \item \textbf{[2$^\textbf{nd}$ assumption]} The role of a concept $C_j$ as necessary to $Y$  ($C_j \supseteq Y$) is independent of every other concept $C_{j'}$'s role as necessary to $Y$ except for $(C_j)^c \supseteq Y$;

    \item \textbf{[3$^\textbf{rd}$ assumption]} The presence of concepts in the same layer $\omega \in C_j$ and $\omega \in C_{j'}$ (with $j \neq j'$) are conditionally independent given the input $x$;
\end{itemize}
we get
\begin{align*}
    \mathbb{P}\left[\omega \in D\right] &= \Big( \mathbb{P}\left[C_1 \supseteq Y\right] \cdot \mathbb{P}\left[\omega \in C_1\right] + \mathbb{P}\left[(C_1)^c \supseteq Y\right] \cdot \big(1 - \mathbb{P}\left[\omega \in C_1\right]\big) + \mathbb{P}\left[ C_1 \not\supseteq Y, (C_1)^c \not\supseteq Y\right] \Big) \\
    & ~~~~~\, \cdot \mathbb{P}\left[\bigcap_{C \in (\mathcal{C}_{\pm}^{l\shortminus 1}\setminus \{C_1, \lnot C_1\})} \Big(\big( C \not\supseteq Y \big) \,\cup\, \big(\omega \in C\big) \Big) \right]\\
    &= \prod_{j=1}^{n^{l\shortminus 1}} \underbrace{\Big( \mathbb{P}\left[C_j \supseteq Y\right] \cdot \mathbb{P}\left[\omega \in C_j\right] + \mathbb{P}\left[(C_j)^c \supseteq Y\right] \cdot \big(1 - \mathbb{P}\left[\omega \in C_j\right]\big) + \mathbb{P}\left[ C_j \not\supseteq Y, (C_j)^c \not\supseteq Y\right] \Big)}_{P_j}.
\end{align*}
We make one final assumption about the probability structure of $C_j \supseteq Y$ and $(C_j)^c \supseteq Y$. We assume that if the network believes one is possible (probability $> 0$), then it believes the other is impossible (probability $=0$) and vice versa. This is how we are able to model both $\mathbb{P}\left[C_j \supseteq Y\right]$ and $\mathbb{P}\left[(C_j)^c \supseteq Y\right]$ with a single variable
\begin{align*}
A_{i,j}^l = \underbrace{\mathbb{P}\left[C_j \supseteq Y\right]}_{[A_{i,j}^l]_+} - \underbrace{\mathbb{P}\left[(C_j)^c \supseteq Y\right]}_{[A_{i,j}^l]_-}.
\end{align*}
If we look at the term $P_j$ in the product on $j$, we can see that

\hspace{0.075cm}\scalebox{0.87}{\parbox{1.\textwidth}{%
\begin{align*}
P_j &= \left\{\begin{array}{ll}
    \mathbb{P}\left[C_j \supseteq Y\right] \cdot \mathbb{P}\left[\omega \in C_j\right] + \mathbb{P}\left[ C_j \not\supseteq Y, (C_j)^c \not\supseteq Y\right], & \textrm{if } \mathbb{P}\left[C_j \supseteq Y\right] > 0, \ \; \mathbb{P}\left[(C_j)^c \supseteq Y\right] = 0, \\[2pt]
    \mathbb{P}\left[(C_j)^c \supseteq Y\right] \cdot \big(1 - \mathbb{P}\left[\omega \in C_j\right]\big) + \mathbb{P}\left[ C_j \not\supseteq Y, (C_j)^c \not\supseteq Y\right], & \textrm{if } \mathbb{P}\left[C_j \supseteq Y\right] = 0, \ \; \mathbb{P}\left[(C_j)^c \supseteq Y\right] > 0, \\[2pt]
    \mathbb{P}\left[ C_j \not\supseteq Y, (C_j)^c \not\supseteq Y\right], & \textrm{if } \mathbb{P}\left[C_j \supseteq Y\right] = \mathbb{P}\left[(C_j)^c \supseteq Y\right] = 0.
\end{array}\right.\\[10pt]
    &= \left\{\begin{array}{ll}
    \mathbb{P}\left[C_j \supseteq Y\right] \cdot \mathbb{P}\left[\omega \in C_j\right] + \big( 1 - \mathbb{P}\left[C_j \supseteq Y\right] \big), & \textrm{if } \mathbb{P}\left[C_j \supseteq Y\right] > 0, \ \; \mathbb{P}\left[(C_j)^c \supseteq Y\right] = 0, \\[2pt]
    \mathbb{P}\left[(C_j)^c \supseteq Y\right] \cdot \big(1 - \mathbb{P}\left[\omega \in C_j\right]\big) + \big( 1 - \mathbb{P}\left[(C_j)^c \supseteq Y\right] \big), & \textrm{if } \mathbb{P}\left[C_j \supseteq Y\right] = 0, \ \; \mathbb{P}\left[(C_j)^c \supseteq Y\right] > 0, \\[2pt]
    1, & \textrm{if } \mathbb{P}\left[C_j \supseteq Y\right] = \mathbb{P}\left[(C_j)^c \supseteq Y\right] = 0.
\end{array}\right.\\[10pt]
    &= \left\{\begin{array}{ll}
    1 - \mathbb{P}\left[C_j \supseteq Y\right] \cdot \big(1 - \mathbb{P}\left[\omega \in C_j\right] \big), & \textrm{if } \mathbb{P}\left[C_j \supseteq Y\right] > 0, \ \; \mathbb{P}\left[(C_j)^c \supseteq Y\right] = 0, \\[2pt]
    1 - \mathbb{P}[(C_j)^c \supseteq Y] \cdot \mathbb{P}[\omega \in C_j], & \textrm{if } \mathbb{P}\left[C_j \supseteq Y\right] = 0, \ \; \mathbb{P}\left[(C_j)^c \supseteq Y\right] > 0, \\[2pt]
    1, & \textrm{if } \mathbb{P}\left[C_j \supseteq Y\right] = \mathbb{P}\left[(C_j)^c \supseteq Y\right] = 0.
\end{array}\right.\\[10pt]
    &= \Big( 1 - \mathbb{P}\left[C_j \supseteq Y\right] \cdot\big(1 - \mathbb{P}\left[\omega \in C_j\right] \big) \Big) \cdot \Big( 1 - \mathbb{P}[(C_j)^c \supseteq Y] \cdot \mathbb{P}[\omega \in C_j] \Big)\\
    &= 1 - \mathbb{P}\left[C_j \supseteq Y\right] \cdot\big(1 - \mathbb{P}\left[\omega \in C_j\right] \big) - \mathbb{P}[(C_j)^c \supseteq Y] \cdot \mathbb{P}[\omega \in C_j] + \cancelto{0}{\mathbb{P}\left[C_j \supseteq Y\right]{\cdot} \mathbb{P}[(C_j)^c \supseteq Y]} {\cdot}\big(1 - \mathbb{P}\left[\omega \in C_j\right] \big) {\cdot} \mathbb{P}[\omega \in C_j]\\[10pt]
    &= 1 - \mathbb{P}\left[C_j \supseteq Y\right] \cdot\big(1 - \mathbb{P}\left[\omega \in C_j\right] \big) - \mathbb{P}[(C_j)^c \supseteq Y] \cdot \mathbb{P}[\omega \in C_j]
\end{align*}
}}

\noindent since we only ever have at most one of $\mathbb{P}\left[C_j \supseteq Y\right]$ and $\mathbb{P}[(C_j)^c \supseteq Y]$ that is non-zero. We conclude
\begin{align*}
    \mathbb{P}\left[\omega \in Y\right] &= \mathbb{P}\left[u \in \tilde{N}_{i}^{l} \mid \omega \in D\right] \prod_{j=1}^{n^{l\shortminus 1}} \Big( 1 - \mathbb{P}\left[C_j \supseteq Y\right] \cdot\big(1 - \mathbb{P}\left[\omega \in C_j\right] \big) \Big)  \Big( 1 - \mathbb{P}[(C_j)^c \supseteq Y] \cdot \mathbb{P}[\omega \in C_j] \Big)\\
    &= \mathbb{P}\left[u \in \tilde{N}_{i}^{l} \mid \omega \in D\right] \prod_{j=1}^{n^{l\shortminus 1}} \Big( 1 - \mathbb{P}\left[C_j \supseteq Y\right] \cdot \big(1 - \mathbb{P}\left[\omega \in C_j\right] \big) - \mathbb{P}[(C_j)^c \supseteq Y] \cdot \mathbb{P}[\omega \in C_j]\Big),
\end{align*}
where the simplified last line is the formulation we use in our implementation.

\paragraph{OR (disjunction)}
\label{sec:app_prob_indep_OR}

We will derive the following formula for the probability of an OR node $C_j^l$ given $x$

\hspace{0.075cm}\scalebox{0.93}{\parbox{1.\textwidth}{%
\begin{align*}
    \scalebox{0.85}{$\displaystyle \mathbb{P}\!\left[\omega \in C_i^l \mid x\right] = 1 - \big( 1 - \mathbb{P}\!\left[u \in \tilde{S}_{i}^{l} \mid \big(\omega \in D\big)^c\right]\big) \prod_{j=1}^{n^{l\shortminus 1}} \Big( 1 - \mathbb{P}\!\left[C_j^{l\shortminus 1} \subseteq C_i^l\right] \cdot \mathbb{P}\!\left[\omega \in C_j^{l\shortminus 1} \mid x\right]\Big) \Big(1 - \mathbb{P}\!\left[(C_j^{l\shortminus 1})^c \subseteq C_i^l\right] \cdot\big(1 - \mathbb{P}\!\left[\omega \in C_j^{l\shortminus 1} \mid x\right] \big)\Big),$}
\end{align*}
}}

\noindent where we define the observed part of definition
\begin{align*}
    \omega \in D &:= \bigcup_{C \in \mathcal{C}_{\pm}^{l\shortminus 1}} \Big(\big(C \subseteq Y \big) \,\cap\, \big(\omega \in C\big) \Big).
\end{align*}

To make this derivation easier to read, we will again abuse notation in the following ways
\begin{align*}
    Y &:= C_i^l,\\
    C_j &:= C_{j}^{l\shortminus1}.
\end{align*}
We will also omit the conditioning on $x$ for the same reason. We remind the reader that the roles of sufficient concepts $C_j \subseteq Y$ are independent of the input $x$.

We want to compute
\begin{align*}
    \mathbb{P}\left[\omega \in Y\right] = \mathbb{P}\left[\big(u \in \tilde{S}_{i}^{l}\big) \,\cup\, \bigcup_{C \in \mathcal{C}_{\pm}^{l\shortminus 1}} \Big(\big(C \subseteq Y \big) \,\cap\, \big(\omega \in C\big) \Big)\right].
\end{align*}
Let $\omega \in D$ be the observed part of the definition
\begin{align*}
    \omega \in D &:= \bigcup_{C \in \mathcal{C}_{\pm}^{l\shortminus 1}} \Big(\big(C \subseteq Y \big) \,\cap\, \big(\omega \in C\big) \Big).
\end{align*}
Then $\mathbb{P}\left[\omega \in Y\right]$ can be rewritten
\begin{align*}
    \mathbb{P}\left[\omega \in Y\right] = \mathbb{P}\left[\big(u \in \tilde{S}_{i}^{l}\big) \;\cup\;  \big(\omega \in D\big)\right].
\end{align*}
Considering the complement of $\omega \in Y$ (its absence $(\omega \in Y)^c$), we have
\begin{align*}
    1 - \mathbb{P}\left[\omega \in Y\right] &= \mathbb{P}\left[(\omega \in Y)^c\right]\\[2pt]
    &= \mathbb{P}\left[(u \in \tilde{S}_{i}^{l})^c \;\cap\;  (\omega \in D)^c\right]\\[2pt]
    &= \mathbb{P}\left[(\omega \in D)^c\right] \cdot \mathbb{P}\left[(u \in \tilde{S}_{i}^{l})^c \mid (\omega \in D)^c\right]\\[2pt]
    &= \big(1 - \mathbb{P}\left[\omega \in D\right]\big) \cdot \big( 1 - \mathbb{P}\left[u \in \tilde{S}_{i}^{l} \mid (\omega \in D)^c\right]\big)\\[2pt]
    \mathbb{P}\left[\omega \in Y\right] &= 1 - \big(1 - \mathbb{P}\left[\omega \in D\right]\big) \cdot \big( 1 - \mathbb{P}\left[u \in \tilde{S}_{i}^{l} \mid (\omega \in D)^c\right]\big),
\end{align*}
and we only need to compute $\mathbb{P}\left[\omega \in D\right]$. Again, we will derive it by induction by considering the contribution of every concept $C_j$ in the previous layer, one at a time. We begin with the law of total probability with respect to the role of $C_1$ or its negation as a sufficient concept, of which there are three possible cases
\begin{align*}
    C_1 \subseteq Y,~ (C_1)^c \not\subseteq Y, \qquad \quad (C_1)^c \subseteq Y,~ C_1 \not\subseteq Y, \qquad \quad C_1 \not\subseteq Y,~ (C_1)^c \not\subseteq Y.
\end{align*}
Since we assume that at most one of $C_1$ and $(C_1)^c$ can be a sufficient concept, we will note these three cases as
\begin{align*}
    C_1 \subseteq Y,\hphantom{~ (C_1)^c \not\subseteq Y,} \qquad \quad (C_1)^c \subseteq Y,\hphantom{~ C_1 \not\subseteq Y,} \qquad \quad C_1 \not\subseteq Y, (C_1)^c \not\subseteq Y.
\end{align*}
We obtain
\begin{align*}
    \mathbb{P}\left[\omega \in D\right] &= \mathbb{P}\left[C_1 \subseteq Y\right] \cdot \mathbb{P}\left[\omega \in D \mid C_1 \subseteq Y\right]\\[2pt]
    & ~~~~\, + \mathbb{P}\left[(C_1)^c \subseteq Y\right] \cdot \mathbb{P}\left[\omega \in D \mid (C_1)^c \subseteq Y\right]\\[2pt]
    & ~~~~\, + \mathbb{P}\left[ C_1 \not\subseteq Y, (C_1)^c \not\subseteq Y\right] \cdot \mathbb{P}\left[\omega \in D \mid C_1 \not\subseteq Y, (C_1)^c \not\subseteq Y\right].
\end{align*} 
Expanding $\mathbb{P}\left[\omega \in D \mid C_1 \subseteq Y\right]$, we get
\begin{align*}
    \mathbb{P}\left[\omega \in D \mid C_1 \subseteq Y\right] &= \mathbb{P}\left[\omega \in C_1\mid C_1 \subseteq Y\right] \cdot \cancelto{1}{\mathbb{P}\left[\omega \in D \mid \omega \in C_1, C_1 \subseteq Y\right]} \\[5pt]
    & ~~~~\, + \big( 1 - \mathbb{P}\left[\omega \in C_1\mid C_1 \subseteq Y\right] \big) \cdot\mathbb{P}\left[\omega \in D \mid \omega \in (C_1)^c, C_1 \subseteq Y\right]\\[5pt]
    &= \mathbb{P}\left[\omega \in C_1\right] + \big( 1 - \mathbb{P}\left[\omega \in C_1\right] \big) \cdot\mathbb{P}\left[\omega \in D \mid \omega \in (C_1)^c, C_1 \subseteq Y\right]
\end{align*}
where we assume \textbf{[1$^\textbf{st}$ assumption]} independence between the presence of concept $C_1$ and its role
\begin{align*}
\omega \in C_{1} \indep C_1 \subseteq Y.
\end{align*}
Expanding $\mathbb{P}\left[\omega \in D \mid (C_1)^c \subseteq Y\right]$, we get
\begin{align*}
    \mathbb{P}\left[\omega \in D \mid (C_1)^c \subseteq Y\right] &= \mathbb{P}\left[\omega \in (C_1)^c\mid (C_1)^c \subseteq Y\right] \cdot \cancelto{1}{\mathbb{P}\left[\omega \in D \mid \omega \in (C_1)^c, (C_1)^c \subseteq Y\right]} \\[5pt]
    & ~~~~\, + \big( 1 - \mathbb{P}\left[\omega \in (C_1)^c\mid (C_1)^c \subseteq Y\right] \big) \cdot\mathbb{P}\left[\omega \in D \mid \omega \in C_1, (C_1)^c \subseteq Y\right]\\[5pt]
    &= \mathbb{P}\left[\omega \in (C_1)^c\right] + \big( 1 - \mathbb{P}\left[\omega \in (C_1)^c\right] \big) \cdot\mathbb{P}\left[\omega \in D \mid \omega \in C_1, (C_1)^c \subseteq Y\right]
\end{align*}
where we assume \textbf{[1$^\textbf{st}$ assumption]} independence between the presence of concept $C_1$ and its role
\begin{align*}
\omega \in (C_{1})^c \indep (C_1)^c \subseteq Y.
\end{align*}
Combining the two previous cases, so far we have
\begin{align*}
    \mathbb{P}\left[\omega \in D\right] &= \mathbb{P}\left[C_1 \subseteq Y\right] \cdot \big( \mathbb{P}\left[\omega \in C_1\right] + \big( 1 - \mathbb{P}\left[\omega \in C_1\right] \big) \cdot\mathbb{P}\left[\omega \in D \mid \omega \in (C_1)^c, C_1 \subseteq Y\right] \big)\\[2pt]
    & ~~~~\, + \mathbb{P}\left[(C_1)^c \subseteq Y\right] \cdot \big(\mathbb{P}\left[\omega \in (C_1)^c\right] + \big( 1 - \mathbb{P}\left[\omega \in (C_1)^c\right] \big) \cdot\mathbb{P}\left[\omega \in D \mid \omega \in C_1, (C_1)^c \subseteq Y\right]\big)\\[2pt]
    & ~~~~\, + \mathbb{P}\left[ C_1 \not\subseteq Y, (C_1)^c \not\subseteq Y\right] \cdot \mathbb{P}\left[\omega \in D \mid (C_1)^c \not\subseteq Y, C_1 \not\subseteq Y\right].
\end{align*}
By using the logical definition of $D$ and by denoting $\mathcal{C}_{\pm,-1}^{l\shortminus 1} := \mathcal{C}_{\pm}^{l\shortminus 1}\setminus \{C_1, \lnot C_1\}$, we get

\hspace{0.075cm}\scalebox{0.85}{\parbox{1.\textwidth}{%
\begin{align*}
    \mathbb{P}\left[\omega \in D\right] &= \mathbb{P}\left[C_1 \subseteq Y\right] \cdot \left( \mathbb{P}\left[\omega \in C_1\right] + \big( 1 - \mathbb{P}\left[\omega \in C_1\right] \big) \cdot \mathbb{P}\left[\bigcup_{C \in \mathcal{C}_{\pm}^{l\shortminus 1}} \Big(\big(C \subseteq Y \big) \,\cap\, \big(\omega \in C\big) \Big) \mid \omega \in (C_1)^c, C_1 \subseteq Y\right] \right)\\ 
    & ~~~~ + \mathbb{P}\left[(C_1)^c \subseteq Y\right] \cdot \left(\mathbb{P}\left[\omega \in (C_1)^c\right] + \big( 1 - \mathbb{P}\left[\omega \in (C_1)^c\right] \big) \cdot\mathbb{P}\left[\bigcup_{C \in \mathcal{C}_{\pm}^{l\shortminus 1}} \!\Big(\big(C \subseteq Y \big) {\cap} \big(\omega \in C\big) \Big) \mid \omega \in C_1, (C_1)^c \subseteq Y\right]\right)\\
    & ~~~~ + \mathbb{P}\left[ C_1 \not\subseteq Y, (C_1)^c \not\subseteq Y\right] \cdot \mathbb{P}\left[\bigcup_{C \in \mathcal{C}_{\pm}^{l\shortminus 1}} \Big(\big(C \subseteq Y \big) \,\cap\, \big(\omega \in C\big) \Big) \mid (C_1)^c \not\subseteq Y, C_1 \not\subseteq Y\right]\\
    &= \mathbb{P}\left[C_1 \subseteq Y\right] \cdot \left( \mathbb{P}\left[\omega \in C_1\right] + \big( 1 - \mathbb{P}\left[\omega \in C_1\right] \big) \cdot\mathbb{P}\left[\bigcup_{C \in \mathcal{C}_{\pm,-1}^{l\shortminus 1}} \Big(\big(C \subseteq Y \big) \,\cap\, \big(\omega \in C\big) \Big) \mid \omega \in (C_1)^c, C_1 \subseteq Y\right] \right)\\ 
    & ~~~~ + \mathbb{P}\left[(C_1)^c \subseteq Y\right] \cdot \left(\mathbb{P}\left[\omega \in (C_1)^c\right] + \big( 1 - \mathbb{P}\left[\omega \in (C_1)^c\right] \big) \cdot\mathbb{P}\left[\bigcup_{C \in \mathcal{C}_{\pm,-1}^{l\shortminus 1}} \!\!\Big(\big(C \subseteq Y \big) {\cap} \big(\omega \in C\big) \Big) \mid \omega \in C_1, (C_1)^c \subseteq Y\right]\right)\\
    & ~~~~ + \mathbb{P}\left[ C_1 \not\subseteq Y, (C_1)^c \not\subseteq Y\right] \cdot \mathbb{P}\left[\bigcup_{C \in \mathcal{C}_{\pm,-1}^{l\shortminus 1}} \Big(\big(C \subseteq Y \big) \,\cap\, \big(\omega \in C\big) \Big) \mid (C_1)^c \not\subseteq Y, C_1 \not\subseteq Y\right].
\end{align*}
}}

\noindent If we assume the following independences
\begin{align*}
    C_1 \subseteq Y &\;\indep\; \left.\bigcup_{C \in (\mathcal{C}_{\pm}^{l\shortminus 1}\setminus \{C_1, \lnot C_1\})} \Big(\big(C \subseteq Y \big) \,\cap\, \big(\omega \in C\big) \Big) ~~\middle|~~ x ,\right.\\
    (C_1)^c \subseteq Y &\;\indep\; \left.\bigcup_{C \in (\mathcal{C}_{\pm}^{l\shortminus 1}\setminus \{C_1, \lnot C_1\})} \Big(\big(C \subseteq Y \big) \,\cap\, \big(\omega \in C\big) \Big) ~~\middle|~~ x ,\right.\\
    \omega \in C_1 &\;\indep\; \left.\bigcup_{C \in (\mathcal{C}_{\pm}^{l\shortminus 1}\setminus \{C_1, \lnot C_1\})} \Big(\big(C \subseteq Y \big) \,\cap\, \big(\omega \in C\big) \Big) ~~\middle|~~ x ,\right.
\end{align*}
\ie if we additionally assume
\begin{itemize}
    \item \textbf{[1$^\textbf{st}$ assumption]} The role of a concept $C_j$ as sufficient to $Y$  ($C_j \subseteq Y$) is independent of every presence of concept $\omega \in C_{j'}$;

    \item \textbf{[2$^\textbf{nd}$ assumption]} The role of a concept $C_j$ as sufficient to $Y$  ($C_j \subseteq Y$) is independent of every other concept $C_{j'}$'s role as sufficient to $Y$ except for $(C_j)^c \subseteq Y$;

    \item \textbf{[3$^\textbf{rd}$ assumption]} The presence of concepts in the same layer $\omega \in C_j$ and $\omega \in C_{j'}$ (with $j \neq j'$) are conditionally independent given the input $x$;
\end{itemize}
we get
\begin{align*}
    \mathbb{P}\left[\omega \in D\right] &= \mathbb{P}\left[C_1 \subseteq Y\right] \cdot \mathbb{P}\left[\omega \in C_1\right] + \mathbb{P}\left[(C_1)^c \subseteq Y\right] \big(1 -  \mathbb{P}\left[\omega \in C_1\right]\big)\\[2pt]
    & ~~~~\, + \underbrace{\Big(\mathbb{P}\left[C_1 \subseteq Y\right] \big(1 - \mathbb{P}\left[\omega \in C_1\right]\big) + \mathbb{P}\left[(C_1)^c \subseteq Y\right] \cdot \mathbb{P}\left[\omega \in C_1\right] + \mathbb{P}\left[ C_1 \not\subseteq Y, (C_1)^c \not\subseteq Y\right]\Big)}_{R_1} \\[2pt]
    & ~~~~~~~~~\,\; \cdot \mathbb{P}\left[\bigcup_{C \in (\mathcal{C}_{\pm}^{l\shortminus 1}\setminus \{C_1, \lnot C_1\})} \Big(\big(C \subseteq Y \big) \,\cap\, \big(\omega \in C\big) \Big) \right].
\end{align*}
If we look at $R_1$, we can see that

\hspace{0.075cm}\scalebox{0.88}{\parbox{1.\textwidth}{%
\begin{align*}
R_1 &= \left\{\begin{array}{ll}
    \mathbb{P}\left[C_1 \subseteq Y\right] \cdot \big(1 - \mathbb{P}\left[\omega \in C_1\right]\big) + \mathbb{P}\left[ C_1 \not\subseteq Y, (C_1)^c \not\subseteq Y\right], & \textrm{if } \mathbb{P}\left[C_1 \subseteq Y\right] > 0, \ \; \mathbb{P}\left[(C_1)^c \subseteq Y\right] = 0, \\[2pt]
    \mathbb{P}\left[(C_1)^c \subseteq Y\right] \cdot \mathbb{P}\left[\omega \in C_1\right] + \mathbb{P}\left[ C_1 \not\subseteq Y, (C_1)^c \not\subseteq Y\right], & \textrm{if } \mathbb{P}\left[C_1 \subseteq Y\right] = 0, \ \; \mathbb{P}\left[(C_1)^c \subseteq Y\right] > 0, \\[2pt]
    \mathbb{P}\left[ C_1 \not\subseteq Y, (C_1)^c \not\subseteq Y\right], & \textrm{if } \mathbb{P}\left[C_1 \subseteq Y\right] = \mathbb{P}\left[(C_1)^c \subseteq Y\right] = 0.
\end{array}\right.\\[10pt]
    &= \left\{\begin{array}{ll}
    \mathbb{P}\left[C_1 \subseteq Y\right] \cdot \big(1 - \mathbb{P}\left[\omega \in C_1\right]\big) + \big( 1 - \mathbb{P}\left[C_1 \subseteq Y\right] \big), & \textrm{if } \mathbb{P}\left[C_1 \subseteq Y\right] > 0, \ \; \mathbb{P}\left[(C_1)^c \subseteq Y\right] = 0, \\[2pt]
    \mathbb{P}\left[(C_1)^c \subseteq Y\right] \cdot \mathbb{P}\left[\omega \in C_1\right] + \big( 1 - \mathbb{P}\left[(C_1)^c \subseteq Y\right] \big), & \textrm{if } \mathbb{P}\left[C_1 \subseteq Y\right] = 0, \ \; \mathbb{P}\left[(C_1)^c \subseteq Y\right] > 0, \\[2pt]
    1, & \textrm{if } \mathbb{P}\left[C_1 \subseteq Y\right] = \mathbb{P}\left[(C_1)^c \subseteq Y\right] = 0.
\end{array}\right.\\[10pt]
    &= \left\{\begin{array}{ll}
    1 - \mathbb{P}\left[C_1 \subseteq Y\right]\cdot \mathbb{P}[\omega \in C_1], & \textrm{if } \mathbb{P}\left[C_1 \subseteq Y\right] > 0, \ \; \mathbb{P}\left[(C_1)^c \subseteq Y\right] = 0, \\[2pt]
    1 - \mathbb{P}[(C_1)^c \subseteq Y] \cdot \big(1 - \mathbb{P}\left[\omega \in C_1\right] \big), & \textrm{if } \mathbb{P}\left[C_1 \subseteq Y\right] = 0, \ \; \mathbb{P}\left[(C_1)^c \subseteq Y\right] > 0, \\[2pt]
    1, & \textrm{if } \mathbb{P}\left[C_1 \subseteq Y\right] = \mathbb{P}\left[(C_1)^c \subseteq Y\right] = 0.
\end{array}\right.\\[10pt]
    &= \Big( 1 - \mathbb{P}\left[C_1 \subseteq Y\right] \cdot \mathbb{P}[\omega \in C_1] \Big) \cdot \Big( 1 - \mathbb{P}[(C_1)^c \subseteq Y] \cdot\big(1 - \mathbb{P}\left[\omega \in C_1\right] \big) \Big)\\
    &= 1 - \mathbb{P}\left[C_1 \subseteq Y\right] \cdot \mathbb{P}[\omega \in C_1] - \mathbb{P}[(C_1)^c \subseteq Y] \cdot\big(1 - \mathbb{P}\left[\omega \in C_1\right] \big) + \cancelto{0}{\mathbb{P}\left[C_1 \subseteq Y\right]{\cdot} \mathbb{P}[(C_1)^c \subseteq Y]}{\cdot} \mathbb{P}[\omega \in C_1] {\cdot}\big(1 - \mathbb{P}\left[\omega \in C_1\right] \big) \\[10pt]
    &= 1 - \mathbb{P}\left[C_1 \subseteq Y\right] \cdot \mathbb{P}[\omega \in C_1] - \mathbb{P}[(C_1)^c \subseteq Y] \cdot\big(1 - \mathbb{P}\left[\omega \in C_1\right] \big)
\end{align*}
}}

\noindent We conclude that

\hspace{0.075cm}\scalebox{0.93}{\parbox{1.\textwidth}{%
\begin{align*}
    \mathbb{P}\left[\omega \in Y\right] &= 1 - \big( 1 - \mathbb{P}\left[u \in \tilde{S}_{i}^{l} \mid (\omega \in D)^c\right]\big) \prod_{j=1}^{n^{l\shortminus 1}} \Big( 1 - \mathbb{P}\left[C_j \subseteq Y\right] \cdot \mathbb{P}[\omega \in C_j]\Big) \Big(1 - \mathbb{P}[(C_j)^c \subseteq Y] \cdot\big(1 - \mathbb{P}\left[\omega \in C_j\right] \big)\Big)\\
    &= 1 - \big( 1 - \mathbb{P}\left[u \in \tilde{S}_{i}^{l} \mid (\omega \in D)^c\right]\big) \prod_{j=1}^{n^{l\shortminus 1}} \Big( 1 - \mathbb{P}\left[C_j \subseteq Y\right] \cdot \mathbb{P}[\omega \in C_j] - \mathbb{P}[(C_j)^c \subseteq Y] \cdot\big(1 - \mathbb{P}\left[\omega \in C_j\right] \big)\Big),
\end{align*}
}}

\noindent where the simplified last line is the formulation we use in our implementation.

To get this final form, we proceed by induction to prove
\begin{align*}
    \mathbb{P}\left[\omega \in D\right] &= 1 - \prod_{j=1}^{n^{l\shortminus 1}} \Big( 1 - \mathbb{P}\left[C_j \subseteq Y\right] \cdot \mathbb{P}[\omega \in C_j] - \mathbb{P}[(C_j)^c \subseteq Y] \cdot\big(1 - \mathbb{P}\left[\omega \in C_j\right] \big)\Big).
\end{align*}
For $n^{l\shortminus 1} = 1$, we have
\begin{align*}
    \mathbb{P}\left[\omega \in D\right] &= \mathbb{P}\left[C_1 \subseteq Y\right] \cdot \mathbb{P}\left[\omega \in C_1\right] + \mathbb{P}\left[(C_1)^c \subseteq Y\right] \big(1 -  \mathbb{P}\left[\omega \in C_1\right]\big) \\[2pt]
    & ~~~~\, + \Big(1 - \mathbb{P}\left[C_1 \subseteq Y\right] \cdot \mathbb{P}[\omega \in C_1] - \mathbb{P}[(C_1)^c \subseteq Y]\big(1 - \mathbb{P}\left[\omega \in C_1\right] \big)\Big) \cdot 0\\[2pt]
    &= 1 - \Big( 1 - \mathbb{P}\left[C_1 \subseteq Y\right] \cdot \mathbb{P}[\omega \in C_1] - \mathbb{P}[(C_1)^c \subseteq Y]\big(1 - \mathbb{P}\left[\omega \in C_1\right] \big) \Big)\\
    &= 1 - \prod_{j=1}^{n^{l\shortminus 1}} \Big( 1 - \mathbb{P}\left[C_j \subseteq Y\right] \cdot \mathbb{P}[\omega \in C_j] - \mathbb{P}[C_j \not\subseteq Y]\big(1 - \mathbb{P}\left[\omega \in C_j\right] \big)\Big).
\end{align*}
If this expression is true for $n^l = k$, for $n^l = k+1$, we have
\begin{align*}
    \mathbb{P}\left[\omega \in D\right] &= \mathbb{P}\left[C_1 \subseteq Y\right] \cdot \mathbb{P}\left[\omega \in C_1\right] + \mathbb{P}\left[(C_1)^c \subseteq Y\right] \big(1 -  \mathbb{P}\left[\omega \in C_1\right]\big)\\
    & ~~~~\, + \Big(1 - \mathbb{P}\left[C_1 \subseteq Y\right] \cdot \mathbb{P}[\omega \in C_j] - \mathbb{P}[(C_j)^c \subseteq Y]\big(1 - \mathbb{P}\left[\omega \in C_1\right] \big)\Big) \\
    & ~~~~~~~~~\,\, \cdot \mathbb{P}\left[\bigcup_{C \in (\mathcal{C}_{\pm}^{l\shortminus 1}\setminus \{C_1 \lnot C_1\})} \Big(\big(C \subseteq Y \big) \,\cap\, \big(\omega \in C\big) \Big) \right]\\
    &= \mathbb{P}\left[C_1 \subseteq Y\right] \cdot \mathbb{P}\left[\omega \in C_1\right] + \mathbb{P}\left[(C_1)^c \subseteq Y\right] \big(1 -  \mathbb{P}\left[\omega \in C_1\right]\big)\\
    & ~~~~\, + \Big(1 - \mathbb{P}\left[C_1 \subseteq Y\right] \cdot \mathbb{P}[\omega \in C_j] - \mathbb{P}[(C_j)^c \subseteq Y]\big(1 - \mathbb{P}\left[\omega \in C_1\right] \big)\Big)  \\
    & ~~~~~~~~~\,\, \cdot\left( 1 - \prod_{j=2}^{n^{l\shortminus 1}} \Big( 1 - \mathbb{P}\left[C_j \subseteq Y\right] \cdot \mathbb{P}[\omega \in C_j] - \mathbb{P}[(C_j)^c \subseteq Y] \cdot\big(1 - \mathbb{P}\left[\omega \in C_j\right] \big)\Big)\right)\\
    &= \mathbb{P}\left[C_1 \subseteq Y\right] \cdot \mathbb{P}\left[\omega \in C_1\right] + \mathbb{P}\left[(C_1)^c \subseteq Y\right] \big(1 -  \mathbb{P}\left[\omega \in C_1\right]\big) \vphantom{\prod_{j=1}^{n^{l\shortminus 1}}}\\
    & ~~~~\, + \Big(1 - \mathbb{P}\left[C_1 \subseteq Y\right] \cdot \mathbb{P}[\omega \in C_j] - \mathbb{P}[(C_j)^c \subseteq Y]\big(1 - \mathbb{P}\left[\omega \in C_1\right] \big)\Big)\\
    & ~~~~\, - \prod_{j=1}^{n^{l\shortminus 1}} \Big( 1 - \mathbb{P}\left[C_j \subseteq Y\right] \cdot \mathbb{P}[\omega \in C_j] - \mathbb{P}[(C_j)^c \subseteq Y] \cdot\big(1 - \mathbb{P}\left[\omega \in C_j\right] \big)\Big)\\
    &= 1 - \prod_{j=1}^{n^{l\shortminus 1}} \Big( 1 - \mathbb{P}\left[C_j \subseteq Y\right] \cdot \mathbb{P}[\omega \in C_j] - \mathbb{P}[(C_j)^c \subseteq Y] \cdot\big(1 - \mathbb{P}\left[\omega \in C_j\right] \big)\Big).
\end{align*}

\subsubsection{Derivation without the 3$^\textrm{rd}$ assumption}
\label{sec:app_theory_prob_deriv_without_3}

The third assumption of conditional independence between concepts in the same layer is necessary for both AND and OR nodes to obtain an easily computable solution at the last step when we need, for AND,

\hspace{0.075cm}\scalebox{0.93}{\parbox{1.\textwidth}{%
\begin{align*}
    \mathbb{P}\left[\bigcap_{C \in (\mathcal{C}_{\pm}^{l\shortminus 1}\setminus \{C_1, \lnot C_1\})} \Big(\big( C \not\supseteq Y \big) \,\cup\, \big(\omega \in C\big) \Big)\right] &= \mathbb{P}\left[\bigcap_{C \in (\mathcal{C}_{\pm}^{l\shortminus 1}\setminus \{C_1, \lnot C_1\})} \Big(\big( C \not\supseteq Y \big) \,\cup\, \big(\omega \in C\big) \Big) \mid \omega \in C_1, C_1 \supseteq Y\right]\\ 
    &= \mathbb{P}\left[\bigcap_{C \in (\mathcal{C}_{\pm}^{l\shortminus 1}\setminus \{C_1, \lnot C_1\})} \Big(\big( C \not\supseteq Y \big) \,\cup\, \big(\omega \in C\big) \Big) \mid \omega \in (C_1)^c, (C_1)^c \supseteq Y\right]\\
    &= \mathbb{P}\left[\bigcap_{C \in (\mathcal{C}_{\pm}^{l\shortminus 1}\setminus \{C_1, \lnot C_1\})} \Big(\big( C \not\supseteq Y \big) \,\cup\, \big(\omega \in C\big) \Big) \mid (C_1)^c \not\supseteq Y, C_1 \not\supseteq Y\right]
\end{align*}
}}

\noindent and, for OR,

\hspace{0.075cm}\scalebox{0.93}{\parbox{1.\textwidth}{%
\begin{align*}
    \mathbb{P}\left[\bigcup_{C \in (\mathcal{C}_{\pm}^{l\shortminus 1}\setminus \{C_1, \lnot C_1\})} \Big(\big(C \subseteq Y \big) \,\cap\, \big(\omega \in C\big) \Big) \right] &= \mathbb{P}\left[\bigcup_{C \in (\mathcal{C}_{\pm}^{l\shortminus 1}\setminus \{C_1, \lnot C_1\})} \Big(\big(C \subseteq Y \big) \,\cap\, \big(\omega \in C\big) \Big) \mid \omega \in (C_1)^c, C_1 \subseteq Y\right]\\ 
    &= \mathbb{P}\left[\bigcup_{C \in (\mathcal{C}_{\pm}^{l\shortminus 1}\setminus \{C_1, \lnot C_1\})} \Big(\big(C \subseteq Y \big) \,\cap\, \big(\omega \in C\big) \Big) \mid \omega \in C_1, (C_1)^c \subseteq Y\right]\\
    &= \mathbb{P}\left[\bigcup_{C \in (\mathcal{C}_{\pm}^{l\shortminus 1}\setminus \{C_1, \lnot C_1\})} \Big(\big(C \subseteq Y \big) \,\cap\, \big(\omega \in C\big) \Big) \mid (C_1)^c \not\subseteq Y, C_1 \not\subseteq Y\right].
\end{align*}
}}

\noindent Given the other assumptions of independence between aleatoric and epistemic probabilities as well as between all epistemic probabilities in the same node, what remains is, for AND,
\begin{align*}
    \mathbb{P}\left[\bigcap_{C \in (\mathcal{C}_{\pm}^{l\shortminus 1}\setminus \{C_1 \lnot C_1\})} \Big(\big( C \not\supseteq Y \big) \,\cup\, \big(\omega \in C\big) \Big)\right] &= \mathbb{P}\left[\bigcap_{C \in (\mathcal{C}_{\pm}^{l\shortminus 1}\setminus \{C_1 \lnot C_1\})} \Big(\big( C \not\supseteq Y \big) \,\cup\, \big(\omega \in C\big) \Big) \mid \omega \in C_1\right]
\end{align*}
and, for OR,
\begin{align*}
    \mathbb{P}\left[\bigcup_{C \in (\mathcal{C}_{\pm}^{l\shortminus 1}\setminus \{C_1 \lnot C_1\})} \Big(\big(C \subseteq Y \big) \,\cap\, \big(\omega \in C\big) \Big) \right] &= \mathbb{P}\left[\bigcup_{C \in (\mathcal{C}_{\pm}^{l\shortminus 1}\setminus \{C_1 \lnot C_1\})} \Big(\big(C \subseteq Y \big) \,\cap\, \big(\omega \in C\big) \Big) \mid \omega \in C_1\right].
\end{align*}
To conclude, we need the missing assumption of conditional independence between all concepts in the same layer
\begin{equation*}
    \left.\omega \in C_i^{l} \indep \omega \in C_j^{l} ~~\middle|~~ x.\right.
\end{equation*}
This assumption seems improbable to say the least since, for $l>0$, the concepts $C_i^{l}$ and $C_j^{l}$ depend in general on the same upstream concepts, \ie on the concepts from the previous layers. However, if we condition on the concepts of the preceding layer $l-1$, the concepts in layer $l$ become independent by d-separation
\begin{align*}
\left.\omega \in C_i^{l} \indep \omega \in C_j^{l} ~~\mid~~ \textrm{C}^{l\shortminus 1}(\omega)\right.,
\end{align*}
where we introduce the notation $\textrm{C}^{l\shortminus 1}(\omega) = \mathbbm{1}\big(\omega \in C^{l\shortminus 1}\big) \in \{0,1\}^{n^l}$ to represent the vector of indicator binary random variables $\textrm{C}_i^{l\shortminus 1}(\omega) = \mathbbm{1}\big(\omega \in C_i^{l\shortminus 1}\big)$ such that the random event $\omega \in C_i^{l\shortminus 1}$ is equal to $\textrm{C}_i^{l\shortminus 1}(\omega)=1$ and its complement $\omega \notin C_i^{l\shortminus 1}$ is equal to $\textrm{C}_i^{l\shortminus 1}(\omega)=0$. The conditional probabilities $\mathbb{P}\left[ \omega \in C_j^l \mid \mathrm{C}^{l\shortminus 1}(\omega) = \mathrm{c}^{l\shortminus 1}\right]$ given the previous layer  for both AND and OR nodes can be easily computed with
\begin{align*}
\mathbb{P}\left[ \omega \in C_j^l \mid \mathrm{C}^{l\shortminus 1} = \mathrm{c}^{l\shortminus 1}\right] = a_j^{l} \prod_{i=1}^{n^{l\shortminus1}} \Big( 1 - [A_{i,j}^{l}]_+ \big( 1 - \mathrm{c}_i^{l\shortminus1} \big) \Big)\Big( 1 - [-A_{i,j}^{l}]_+ \; \mathrm{c}_i^{l\shortminus1} \Big),
\end{align*}
for AND and
\begin{align*}
\mathbb{P}\left[ \omega \in C_{j}^l \mid \mathrm{C}^{l\shortminus 1}(\omega) = \mathrm{c}^{l\shortminus 1}\right] = 1 - \big(1- o_j^{l}\big) \prod_{i=1}^{n^{l\shortminus1}} \Big( 1 - [O_{i,j}^{l}]_+ \; \mathrm{c}_i^{l\shortminus1}\Big) \Big(1 - [-O_{i,j}^{l}]_+ \big( 1 - \mathrm{c}_i^{l\shortminus1} \big) \Big),
\end{align*}
for OR. These are the same formulas as when there is full independence and we derive them in the same way. Moreover, because of their conditional independence, their joint conditional probability is given by, $\forall (\mathrm{c}^{l\shortminus 1},\mathrm{c}^l) \in \{0,1\}^{n^{l\shortminus 1} \times n^l}$,
\begin{align*}
    \mathbb{P}\left[\mathrm{C}^{l}(\omega) = \mathrm{c}^{l} \mid \mathrm{C}^{l\shortminus 1}(\omega) = \mathrm{c}^{l\shortminus 1}\right] \; = \prod_{i = 1}^{n^l} \Big(\mathrm{c}_i^l \, \cdot \, \mathbb{P}\left[ \omega \in C_i^l \mid \mathrm{C}^{l\shortminus 1} = \mathrm{c}^{l\shortminus 1}\right] + \big(1 - \mathrm{c}_i^l\big)\big(1 - \mathbb{P}\left[ \omega \in C_i^l \mid \mathrm{C}^{l\shortminus 1} = \mathrm{c}^{l\shortminus 1}\right]\big)  \Big).
\end{align*}

The input layer 0 has no preceding layer, but it does satisfy
\begin{align*}
\left.x \in C_i^{0} \indep x \in C_j^{0} ~~\middle|~~ x.\right.
\end{align*}
Its joint probability is thus given by
\begin{align*}
    \mathbb{P}[\mathrm{C}^0(x) = \mathrm{c}^0] \; &= \prod_{i=1}^{n^0} \Big(\mathrm{c}_i^0 \cdot \mathbb{P}[x \in C_i^0] + \big(1 - \mathrm{c}_i^0\big)\big(1 - \mathbb{P}[x \in C_i^0]\big)  \Big), \qquad \forall \mathrm{c}^0 \in \{0,1\}^{n^0}
\end{align*}
where the probabilities $\mathbb{P}[x \in C_i^0]$ are given.

Armed with these independences, we can take into account the dependences between concepts that are defined partially on the same concepts. We compute the probabilities layer by layer starting with the first logical layer. We can compute their joint probability with
\begin{align*}
    \mathbb{P}[\mathrm{C}^l(\omega) = \mathrm{c}^l] = \sum_{\mathrm{c}^{l\shortminus 1} \in \{0,1\}^{n^{l\shortminus 1}}} \mathbb{P}[\mathrm{C}^{l\shortminus 1}(\omega) = \mathrm{c}^{l\shortminus 1}] \cdot \mathbb{P}\left[ \mathrm{C}^l(\omega) = \mathrm{c}^l \mid \mathrm{C}^{l\shortminus 1}(\omega) = \mathrm{c}^{l\shortminus 1}\right]
\end{align*}
and their marginal probabilities with
\begin{align*}
    \mathbb{P}[\omega \in C_j^l] = \sum_{\mathrm{c}^{l\shortminus 1} \in \{0,1\}^{n^{l\shortminus 1}}} \mathbb{P}[\mathrm{C}^{l\shortminus 1}(\omega) = \mathrm{c}^{l\shortminus 1}] \cdot \mathbb{P}\left[ \omega \in C_j^l \mid \mathrm{C}^{l\shortminus 1}(\omega) = \mathrm{c}^{l\shortminus 1}\right].
\end{align*}

The issue with this approach is that we need full joint probability distributions
\begin{align*}
    \mathbb{P}\left[ \mathrm{C}^l(\omega) = \mathrm{c}^l \mid \mathrm{C}^{l\shortminus 1} = \mathrm{c}^{l\shortminus 1}\right],
\end{align*}
for every pair of joint values $(\mathrm{c}^{l\shortminus 1},\mathrm{c}^l) \in \{0,1\}^{n^{l\shortminus 1} \times n^l}$, for every layer. For each layer, a tensor of $2^{n^{l\shortminus 1} + n^l}$ entries taking values in $[0,1]$ would be needed for inference and would have to be re-computed after every learning step. This combinatorial explosion results in a exponential number of computations and memory that is only viable in applications with a very small number of nodes per layer.

Moreover, based on preliminary test results, it seems that this extended formulation does not improve the modeling ability of the framework meaningfully. We tested both formulations on randomly generated data that follows the assumptions of our model. We sampled random logical networks of 10 conjunction layers with
\begin{itemize}
    \item inputs modeled by independent Bernouillis with parameters uniformly sampled in $(0,1)$,

    \item weights between layers of either $A_{i,j}^l = 0$ with probability 1/2 or $A_{i,j}^l \in \{1,-1\}$ with probability 1/4 each,

    \item and unobserved concepts also modeled by independent Bernouillis of parameter $a_j^l$ uniformly sampled in $(0,1)$.
\end{itemize}
\begin{figure}[t!]
    \centering
    \includegraphics[width=0.75\textwidth]{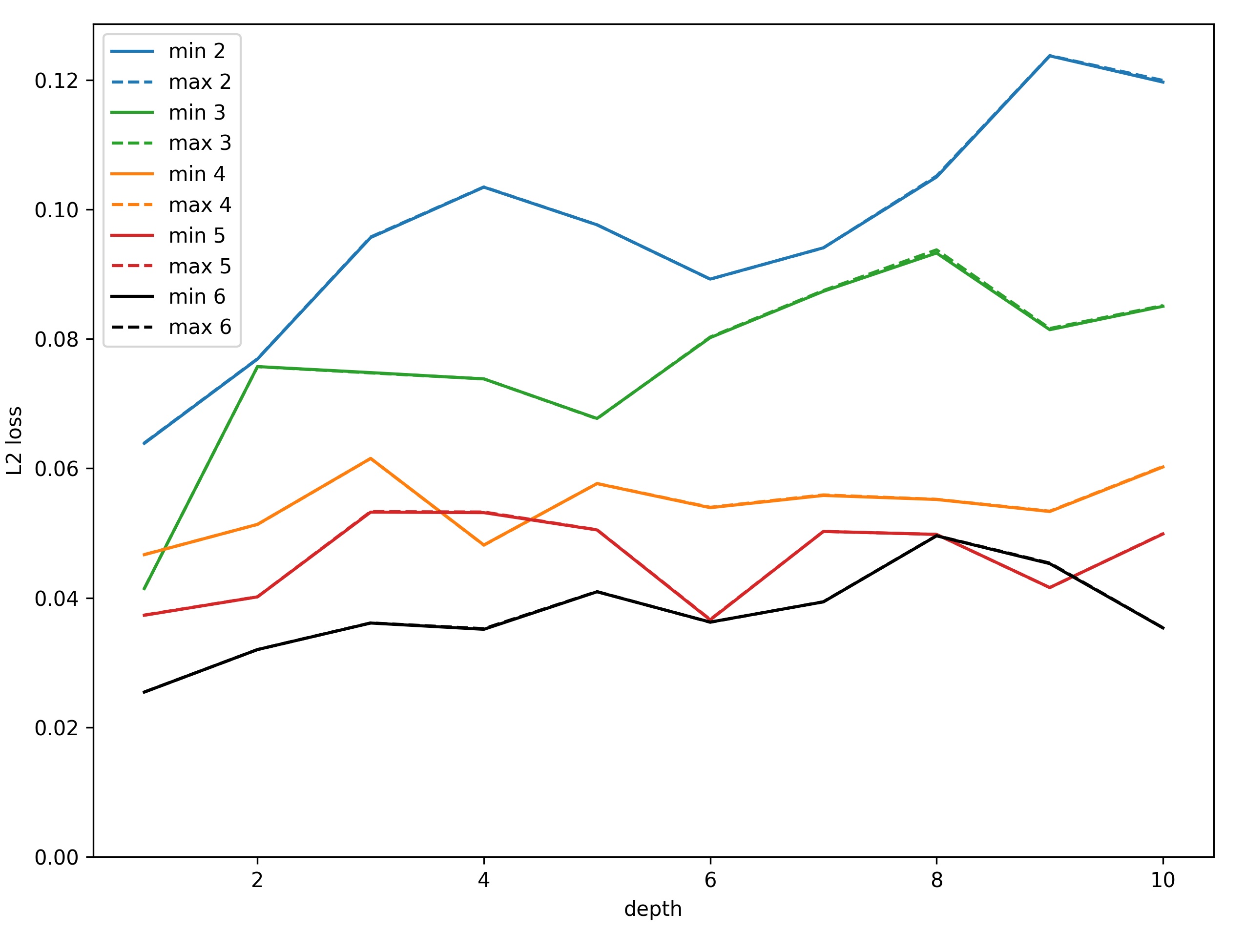}
    \caption{Preliminary comparison of independence hypotheses}
    \label{fig:hyp_comp}
\end{figure}
We additionally added weights of either $+1$ or $-1$ when a concept node had only incoming weights of $0$ (\ie an empty definition) or when a node had only outgoing weights of $0$ (\ie was unused in the next layer). We assumed a fixed width for each network that had the same number of inputs, outputs and concepts in all layers. We considered widths between 2 and 6 and, for each width, we sampled 30 such generating models, which produced data sets of 1000 points for each. We then measured the $L_2$ loss of both of our frameworks on these data sets. We computed this loss for all concepts at all depths, from layer $l=1$ to $l=L=10$. The results are given in Figure \ref{fig:hyp_comp} where the independent inputs are labeled ``min'' (for minimal independence hypothesis) and independent concepts in all layers are labeled ``max'' (for maximal independence hypothesis).
For $l=1$, the two frameworks are equivalent and their results are the same. Only for bigger depths do we start to see a small difference in performance, although it seems negligible in this preliminary testing.

% \newpage

\subsection{Logical modeling}

\subsubsection{De Morgan's laws for the AND/OR concepts}
\label{sec:app_theory_log_deMorgan}

\paragraph{Between (P-AND) and (P-OR)} Since (F-AND) and (F-OR) are equivalent rewritings and since product fuzzy logic's t-norm $\pand$ and t-conorm $\por$ follow de Morgan's laws with strong negation $\pnot$ \citep{Krieken22}, it is easier to show the De Morgan's laws between (F-AND) and (F-OR).

Let $C_i^l$ be an AND concept with
\begin{align*}
    c_i^{l}(x) &= a_i^{l} \ \pand\; \stackrel{{\textcolor{gray}{\scriptscriptstyle \textrm{P}}}}{\bigwedge_{j \in \{1, ..., n^{l\shortminus1}\}}} \Big( \pnot \pospart{A_{i,j}^{l}} \;\por\; c_j^{l\shortminus1}(x) \Big) \;\pand\; \Big( \pnot \negpart{A_{i,j}^{l}} \;\por\; \pnot\, c_j^{l\shortminus1}(x) \Big).
\end{align*}
We want to show that its opposite is an OR concept $C_{i'}^l$ with
\begin{align*}
    c_{i'}^{l}(x) &= o_{i'}^{l} \ \por\; \stackrel{{\textcolor{gray}{\scriptscriptstyle \textrm{P}}}}{\bigvee_{j \in \{1, ..., n^{l\shortminus1}\}}} \Big( \pospart{O_{i',j}^{l}} \;\pand\; c_j^{l\shortminus1}(x)\Big) \;\por\; \Big(\negpart{O_{i',j}^{l}} \;\pand\; \pnot\, c_j^{l\shortminus1}(x) \Big).
\end{align*}

We begin by taking the opposite of the AND concept $C_i^l$.
\begin{align*}
    \pnot\, c_i^{l}(x) &= \pnot \left(a_i^{l} \ \pand\; \stackrel{{\textcolor{gray}{\scriptscriptstyle \textrm{P}}}}{\bigwedge_{j \in \{1, ..., n^{l\shortminus1}\}}} \Big( \pnot \pospart{A_{i,j}^{l}} \;\por\; c_j^{l\shortminus1}(x) \Big) \;\pand\; \Big( \pnot \negpart{A_{i,j}^{l}} \;\por\; \pnot\, c_j^{l\shortminus1}(x) \Big)\right)\\
    &= \pnot\, a_i^{l} \ \por\; \stackrel{{\textcolor{gray}{\scriptscriptstyle \textrm{P}}}}{\bigvee_{j \in \{1, ..., n^{l\shortminus1}\}}} \pnot \Big( \pnot \pospart{A_{i,j}^{l}} \;\por\; c_j^{l\shortminus1}(x) \Big) \;\por\; \pnot \Big( \pnot \negpart{A_{i,j}^{l}} \;\por\; \pnot\, c_j^{l\shortminus1}(x) \Big)\\
    &= \pnot\, a_i^{l} \ \por\; \stackrel{{\textcolor{gray}{\scriptscriptstyle \textrm{P}}}}{\bigvee_{j \in \{1, ..., n^{l\shortminus1}\}}} \Big( \pospart{A_{i,j}^{l}} \;\pand\; \pnot\, c_j^{l\shortminus1}(x) \Big) \;\por\; \Big( \negpart{A_{i,j}^{l}} \;\pand\; c_j^{l\shortminus1}(x) \Big)
\end{align*}
By identification, we can see that
\begin{align*}
    o_{i'}^{l} = \pnot\, a_i^{l} = 1 - a_i^{l}, \qquad \textrm{and} \qquad O_{i',j}^{l} = - A_{i,j}^{l},
\end{align*}
\ie and AND concept can be converted to an OR concept, and vice versa, by taking the complement of its bias and flipping the signs of its incoming and outgoing weights.

\paragraph{Between (D-AND) and (D-OR)} Since there is a one-to-one translation from (D-AND) and (D-OR) to (L-AND) and (L-OR) and since both intersection ($\cap$) and union ($\cup$) as well as conjunction ($\land$) and disjunction ($\lor$) follow De Morgan's laws, it is enough to show that De Morgan's laws hold between (D-AND) and (D-OR) to have the same between (L-AND) and (L-OR).

Let $C_i^l$ be an AND concept with
\begin{align*}
    \omega \in C_i^l &= (u \in \tilde{N}_{i}^{l}) \,\cap\, \bigcap_{C \in \mathcal{C}_{\pm}^{l\shortminus 1}} \Big(\big( C \supseteq C_i^l \big)^c \,\cup\, \big(\omega \in C\big) \Big),
\end{align*}
We want to show that its opposite is an OR concept $C_{i'}^l$ with
\begin{align*}
    \omega \in C_{i'}^l &= (u \in \tilde{S}_{i'}^{l}) \,\cup\, \bigcup_{C \in \mathcal{C}_{\pm}^{l\shortminus 1}} \Big(\big(C \subseteq C_{i'}^l \big) \,\cap\, \big(\omega \in C\big) \Big),
\end{align*}

We begin by taking the opposite of the AND concept $C_i^l$.
\begin{align*}
    \omega \in (C_i^l)^c &= (\omega \in C_i^l)^c \\
    &= \left((u \in \tilde{N}_{i}^{l}) \,\cap\, \bigcap_{C \in \mathcal{C}_{\pm}^{l\shortminus 1}} \Big(\big( C \supseteq C_i^l \big)^c \,\cup\, \big(\omega \in C\big) \Big)\right)^c\\
    &= (u \in \tilde{N}_{i}^{l})^c \,\cup\, \bigcup_{C \in \mathcal{C}_{\pm}^{l\shortminus 1}} \Big(\big( C \supseteq C_i^l \big)^c \,\cup\, \big(\omega \in C\big) \Big)^c\\
    &= (u \in \tilde{N}_{i}^{l})^c \,\cup\, \bigcup_{C \in \mathcal{C}_{\pm}^{l\shortminus 1}} \Big(\big( C \supseteq C_i^l \big) \,\cap\, \big(\omega \in C\big)^c \Big)\\
    &= (u \in (\tilde{N}_{i}^{l})^c) \,\cup\, \bigcup_{C \in \mathcal{C}_{\pm}^{l\shortminus 1}} \Big(\big( C \supseteq C_i^l \big) \,\cap\, \big(\omega \in (C)^c\big) \Big)\\
    &= (u \in (\tilde{N}_{i}^{l})^c) \,\cup\, \bigcup_{C' \in \mathcal{C}_{\pm}^{l\shortminus 1}} \Big(\big( (C')^c \supseteq C_i^l \big) \,\cap\, \big(\omega \in C'\big) \Big)\\
    &= (u \in (\tilde{N}_{i}^{l})^c) \,\cup\, \bigcup_{C' \in \mathcal{C}_{\pm}^{l\shortminus 1}} \Big(\big( C' \subseteq (C_i^l)^c \big) \,\cap\, \big(\omega \in C'\big) \Big)\\
    &= (u \in \tilde{S}_{i'}^{l}) \,\cup\, \bigcup_{C' \in \mathcal{C}_{\pm}^{l\shortminus 1}} \Big(\big( C' \subseteq C_{i'}^l \big) \,\cap\, \big(\omega \in C'\big) \Big)\\
    &= \omega \in  C_{i'}^l
\end{align*}
where $C_{i'}^l = (C_i^l)^c$ is an OR concept. By identification, we can see that
\begin{align*}
    \tilde{S}_{i'}^{l} = (\tilde{N}_{i}^{l})^c,
\end{align*}
and that a concept $C$ in the previous layer is a sufficient concept $C \subseteq C_{i'}^l$ of OR concept $C_{i'}^l$ whenever its opposite $(C)^c$ was a necessary concept $(C)^c \supseteq C_i^l$ of AND concept $C_i^l$.

\subsection{Interpretation}
\label{sec:app_theory_interp}

Since any finite combination of AND (resp. OR) concepts can be represented by a single AND (resp. OR) concept, each type of concept can represent an infinite number of cases. Furthermore, in any such case, the missing necessary (resp. sufficient) concepts that are needed to determine the presence of the target concept $\omega \in C_i^l$ can all be absorbed into the unobserved concepts $u \in \tilde{N}_i^l$ (resp. $u \in \tilde{S}_i^l$), through the probability $a_{i}^l$ (resp. $o_i^l$). We give some intuitive and counter-intuitive examples below. 

In the following figures of causal structures, the circles are concepts, the arrows are cause-to-consequence relations (implications) and the bracket signifies a conjunction (AND) of concepts. The whites circles are the necessary or sufficient concepts, the black circle is the target concept and the gray circles are concepts that are not needed to determine the target's presence if given the white circles.

\subsubsection{Examples of causal structures that can be represented by an AND node}

\begin{figure}[h!]
    \centering
     \begin{subfigure}[b]{0.225\textwidth}
         \centering
         \includegraphics[width=\textwidth]{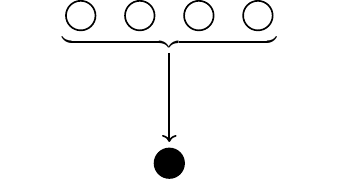}
         \caption{}
         \label{fig:interp_a}
     \end{subfigure}
     \hfill
     \begin{subfigure}[b]{0.225\textwidth}
         \centering
         \includegraphics[width=\textwidth]{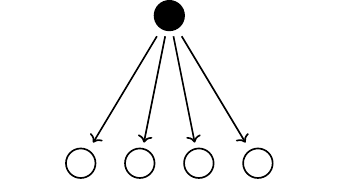}
         \caption{}
         \label{fig:interp_b}
     \end{subfigure}
     \hfill
     \begin{subfigure}[b]{0.225\textwidth}
         \centering
         \includegraphics[width=\textwidth]{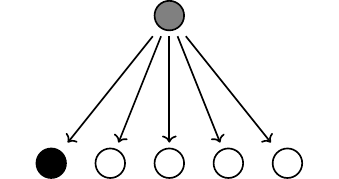}
         \caption{}
         \label{fig:interp_c}
     \end{subfigure}
     \hfill
     \begin{subfigure}[b]{0.225\textwidth}
         \centering
         \includegraphics[width=\textwidth]{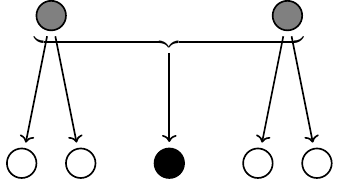}
         \caption{}
         \label{fig:interp_d}
     \end{subfigure}
    \caption{Examples of causal structures that can be represented by an AND node}
    \label{fig:interp}
\end{figure}

An AND node can represent a consequence of necessary causal ingredients (Figure \ref{fig:interp_a}). The same AND node can represent the opposite direction of causality, where the AND concept is the cause and the necessary concepts are its consequences (Figure \ref{fig:interp_b}). In this case, the unobserved concepts probability $a_{i}^l$ quantifies how often this common cause is what caused these consequences, when they are all present simultaneously. A similar case is when the AND node represents another one of these consequences from the same common cause (Figure \ref{fig:interp_c}). An AND node can even represent a consequence of causal ingredients which, themselves produce their own individual necessary consequences (Figure \ref{fig:interp_d}).

\subsubsection{Examples of causal structures that can be represented by an OR node}

\begin{figure}[h!]
    \centering
     \begin{subfigure}[b]{0.225\textwidth}
         \centering
         \includegraphics[width=\textwidth]{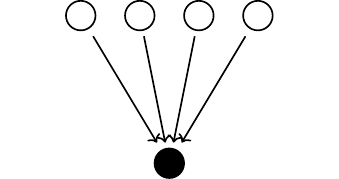}
         \caption{}
         \label{fig:interp_a2}
     \end{subfigure}
     \hfill
     \begin{subfigure}[b]{0.225\textwidth}
         \centering
         \includegraphics[width=\textwidth]{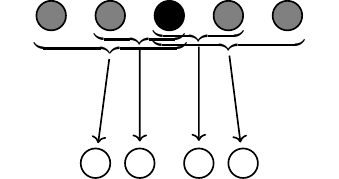}
         \caption{}
         \label{fig:interp_b2}
     \end{subfigure}
     \hfill
     \begin{subfigure}[b]{0.225\textwidth}
         \centering
         \includegraphics[width=\textwidth]{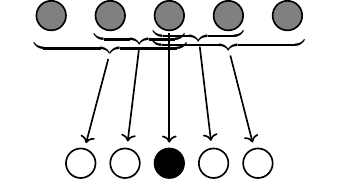}
         \caption{}
         \label{fig:interp_c2}
     \end{subfigure}
     \hfill
     \begin{subfigure}[b]{0.225\textwidth}
         \centering
         \includegraphics[width=\textwidth]{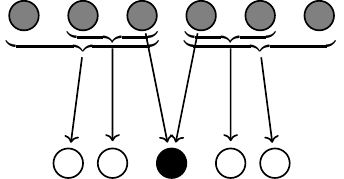}
         \caption{}
         \label{fig:interp_d2}
     \end{subfigure}
    \caption{Examples of causal structures that can be represented by an OR node}
    \label{fig:interp2}
\end{figure}
An OR node can represent a consequence of some sufficient concepts (Figure \ref{fig:interp_a2}). The same OR node can represent the opposite direction of causality, where the OR concept is a causal ingredient and the sufficient concepts are its possible consequences (Figure \ref{fig:interp_b2}). In this case, the unobserved sufficient concepts probability $o_{i}^l$ quantifies how often this common causal ingredient is still present even when none of these consequences are present. A similar case is when the OR node represents a direct consequence of this common causal ingredient (Figure \ref{fig:interp_c2}). An OR node can even represent a consequence of causes which, together with other causal ingredients produce their own consequences (Figure \ref{fig:interp_d2}).

\section{Machine Learning pipeline}

\subsection{Post-processing}

\subsubsection{Other weight discretization algorithms}
\label{sec:app_ML_post_discretiz}

~

\begin{figure}[h!]
    \centering
\scalebox{1}{\begin{algorithm}[H]
\SetKwIF{If}{ElseIf}{Else}{If}{$\!$,}{Else if}{Else}{}%
\SetKwFor{For}{For}{$\!$,}{}%
\SetKwFor{While}{While}{$\!$}{}%
\SetAlgoVlined
\DontPrintSemicolon
\For{{\upshape each layer }$l \in \{1,..,L\}${\upshape , starting from the last layer $L$}}{
    \For{{\upshape each weight }$A_{i,j}^l${\upshape ~(resp. }$O_{i,j}^l${\upshape ), in increasing likeliness }$\left|A_{i,j}^l\right|${\upshape ~(resp. }$\left|O_{i,j}^l\right|${\upshape )}}{
        \If{{\upshape it is non-zero}}{
            Compare the loss when we fix $A_{i,j}^l \in \left\{0,\, \textrm{sign}\left(A_{i,j}^l\right)\right\}$ (resp. $O_{i,j}^l \in \left\{0,\, \textrm{sign}\left(O_{i,j}^l\right)\right\}$).\;
            Commit to the best discretized value.\;
        }
    }
}
Do the same for the category and continuous input modules, one at a time.\;
\end{algorithm}}
    \caption*{Algorithm C.3: Ascending selection discretization algorithm}
    \label{fig:enter-label}
\end{figure}

\begin{figure}[h!]
    \centering
\scalebox{1}{\begin{algorithm}[H]
\SetKwIF{If}{ElseIf}{Else}{If}{$\!$,}{Else if}{Else}{}%
\SetKwFor{For}{For}{$\!$,}{}%
\SetKwFor{While}{While}{$\!$}{}%
\SetAlgoVlined
\DontPrintSemicolon
\For{{\upshape each layer }$l \in \{1,..,L\}${\upshape , starting from the last layer $L$}}{
    discretize all weights in the layer to their sign $A_{\cdot,\cdot}^l = \textrm{sign}(A_{\cdot,\cdot}^l)$ (resp. $O_{i,j}^l = \textrm{sign}\left(O_{i,j}^l\right)$) and keep original values in $\tilde{A}_{\cdot,\cdot}^l$ (resp. $\tilde{O}_{\cdot,\cdot}^l$).\;
    Compute the new best loss $e^*$.\;
    \For{{\upshape each weight }$A_{i,j}^l${\upshape ~(resp. }$O_{i,j}^l${\upshape ), in increasing likeliness }$\left|\tilde{A}_{i,j}^l\right|${\upshape ~(resp. }$\left|\tilde{O}_{i,j}^l\right|${\upshape )}}{
        \If{{\upshape it is non-zero}}{
            Compare the loss when we prune the weight $A_{i,j}^l = 0$ (resp. $O_{i,j}^l = 0$) to the best loss $e^*$.\;
            Commit to the pruning iff the loss is decreased and update the best loss $e^*$ in that case.\;
        }
    }
}
Do the same for the category and continuous input modules, one at a time.\;
\end{algorithm}}
    \caption*{Algorithm C.1: Subtractive discretization algorithm}
    \label{fig:enter-label}
\end{figure}

\begin{figure}[h!]
    \centering
\scalebox{1}{\begin{algorithm}[H]
\SetKwIF{If}{ElseIf}{Else}{If}{$\!$,}{Else if}{Else}{}%
\SetKwFor{For}{For}{$\!$,}{}%
\SetKwFor{While}{While}{$\!$}{}%
\SetAlgoVlined
\DontPrintSemicolon
\For{{\upshape each layer }$l \in \{1,..,L\}${\upshape , starting from the last layer $L$}}{
    Prune all weights in the layer $A_{\cdot,\cdot}^l = 0$ (resp. $O_{i,j}^l = 0$) and keep original values in $\tilde{A}_{\cdot,\cdot}^l$ (resp. $\tilde{O}_{\cdot,\cdot}^l$).\;
    Compute the new best loss $e^*$.\;
    \For{{\upshape each weight }$A_{i,j}^l${\upshape ~(resp. }$O_{i,j}^l${\upshape ), in decreasing likeliness }$\left|\tilde{A}_{i,j}^l\right|${\upshape ~(resp. }$\left|\tilde{O}_{i,j}^l\right|${\upshape )}}{
        \If{{\upshape it is non-zero}}{
            Compare the loss when we fix $A_{i,j}^l = \textrm{sign}(\tilde{A}_{i,j}^l)$ (resp. $O_{i,j}^l = \textrm{sign}(\tilde{O}_{i,j}^l)$) to the best loss $e^*$.\;
            Commit to the new value iff the loss is decreased and update the best loss $e^*$ in that case.\;
        }
    }
}
Do the same for the category and continuous input modules, one at a time.\;
\end{algorithm}}
    \caption*{Algorithm C.2: Additive discretization algorithm}
    \label{fig:enter-label}
\end{figure}

% \begin{table}[h!]
%     \centering
%     \caption{Comparison of five-fold cross-validation f1-score (\%)}
%     \label{tab:tab_res_discretiz}
%     \begin{tabular}{llllll}
%         \toprule
%         Data sets & Sel-Desc & Sel-Asc & Sub & Add & Thresh \\
%         \midrule
%         adult & ? & \textbf{?} & ? & ? & ?\\
%         balance & ? & \textbf{?} & ? & ? & ?\\
%         balance (cat.) & ? & ? & ? & \textbf{?} & ?\\
%         chess & ? & \textbf{?} & ? & ? & ?\\
%         monk2 & ? & ? & ? & \textbf{?} & ?\\
%         tictactoe & \textbf{100.00} & \textbf{100.00} & \textbf{100.00} & \textbf{100.00} & \textbf{100.00}\\
%         wine & ? & ? & \textbf{?} & ? & ?\\
%         \midrule
%         Average & \textbf{?} & ? & ? & ? & ?\\
%         \bottomrule
%     \end{tabular}
% \end{table}

% \newpage
\subsubsection{Data set coverage analysis and elimination of included rules}
\label{sec:app_ML_post_inclusion}

We say that a rule $i$ is \emph{included} in rule $j$ if $\mathbb{P} \! \left[ \tilde{C}_j^1 \mid \tilde{C}_i^1  \right] = 1$, where $\tilde{C}_i^1$ corresponds to the presence of rule $i$ without taking into account the bias $a_i$ from its unobserved necessary concepts. In other words, a rule $i$ is included in rule $j$ if all the data points that are covered by rule $i$ are also covered by rule $j$. We wish to estimate this probability with an expression of the form
\begin{align*}
    {\subseteq}(i,j) = \frac{\displaystyle \sum_{(x,y) \in \mathcal{D}} e_{i,j}(x)}{\displaystyle \sum_{(x,y) \in \mathcal{D}} \tilde{c}_i^1(x)},
\end{align*}
where $e_{i,j}(x) = \mathbb{P}\!\left[ \tilde{C}_i^1 \cap \tilde{C}_j^1 \mid x \right] \leq \mathbb{P}\!\left[ \tilde{C}_i^1 \mid x \right] = \tilde{c}_i^1(x)$ represents the probability that both rules $i$ and $j$ are activated for an input $x$ of the full training data set $\mathcal{D}$. If we simply used the product
\begin{align*}
    e_{i,j}(x) = \tilde{c}_i^1(x) \cdot \tilde{c}_j^1(x),
\end{align*}
as we do in our AND nodes, we could be end up in a situation where $e_{i,i}(x) < \tilde{c}_i^1(x)$ which would violate the reflexivity of inclusion, such that a rule is always included in itself ${\subseteq}(i,i) = 1$. Instead, to make sure that reflexivity is conserved, we use the following
\begin{align*}
    e_{i,j}(x) = \min\!\left\{\,\sqrt{\tilde{c}_i^1(x) \cdot \tilde{c}_j^1(x)} \,,\, \tilde{c}_i^1(x)\,\right\}.
\end{align*}
Using this definition, we remove a rule $i$ associated to an output target $k$ if it is included in another rule $j$ associated to the same output $k$ and $a_i < a_j$.

\section{Experiments}

\subsection{Boolean networks discovery}
\label{sec:app_exp_bool}

Boolean networks were introduced in \cite{Kauffman69} to model gene regulatory networks in biology. A boolean network models gene interactions where $n$ genes at a time step $t$ are either activated $A_i^t=1$ or not $A_i^t=0$. Given the activations $A_i^t$ at step $t$, their activations $A_i^{t+1}$ at the next time step $t+1$ are deterministically given by a logic program where each gene is activated if one of its rules are satisfied in the previous time step. For instance, we might have for gene 3 a simple logic program with only two rules
\begin{flalign*}
    \lnot A_4^t &\rightarrow A_3^{t+1}, \\
    A_1^t \land \lnot A_2^t &\rightarrow A_3^{t+1},
\end{flalign*}
\ie gene $3$ is activated at time step $t+1$ if at time step $t$, either gene 4 was not activated or gene 1 was activated and gene 2 was not, or both. For this task, we are provided a data set with all possible gene state transitions $(A_1^t,...,A_n^t,A_1^{t+1},...,A_n^{t+1})$ and we want to discover the ground-truth logic program by learning to predict the end state $(A_i^{t+1})_i$ from the start state $(A_i^{t})_i$. Since we almost never have access to this full distribution, we study the performance of our algorithm for partial data sets with ratios of the full data set ranging from 10\% to 100\%. We consider four data sets with known ground-truth logic programs: mammalian cell cycle regulation \citep{Faure06}, fission yeast cell cycle regulation \citep{Davidich08}, budding yeast cell cycle regulation \citep{Li04} and arabidopsis thaliana flower morphogenesis \citep{Chaos06}. We evaluate the performance of our algorithm according to its accuracy in two repeats of Five-Fold Cross-Validation (5F-CV).

For this task, we do not split the training set into training and validation sets. Since there is a ground-truth logic program in this task, we have observed in our experiments that NLNs do not tend to overfit and every additional data point is relevant to find all the correct rules. We compare our approach with four other methods that were used for this task. Two of them are neuro-symbolic in nature while the other two are purely symbolic. The first neuro-symbolic method is NN-LFIT \citep{Tourret17} which learns a neural network that is then approximated by a logic program. The second neuro-symbolic method is D-LFIT \citep{Gao21} which learns a logic program that is embedded in a set of matrices in a novel neural network structure. The purely symbolic methods are the Inductive Logic Programming method LF1T \citep{Inoue14} and the symbolic rule learner JRip \citep{Witten16}. The results are presented in Table \ref{tab:res_bool_net}, which is partially reproduced from \cite{Gao21}.
\begin{table}[h!]
    \centering
    \caption{Comparison of five-fold cross-validation accuracy (\%) on partial data sets with different split rates}
    \label{tab:res_bool_net}
    \scalebox{1}{\begin{tabular}{llllll}
        \toprule
        Data sets (variables, rules) & Model & \multicolumn{4}{l}{Ratio of the full distribution} \\ \cmidrule(r){3-6}
         &  & 8\% & 16\% & 40\% & 80\%\\%Split rates
        \midrule
        Mammalian (10, 23) & \textbf{NLN} & 92.42 & \textbf{98.46} & \textbf{100} & \textbf{100} \\
         & NN-LFIT & \textbf{96.60} &  94.35 &  99.89 &  99.91 \\
         & D-LFIT & 71.67 & 75.9 & 80.09 & 82.84 \\
         & LF1T & 76.01 & 76.48 & 76.73 & 91.56 \\
         & JRip & 77.84 & 75.44 & 76.41 & 74.66 \\[3pt]
        Fission (10, 24) & \textbf{NLN} & 91.59 & 97.69 & \textbf{100} & \textbf{100} \\
         & NN-LFIT & \textbf{98.80} & \textbf{99.80} & 99.92 & 99.87 \\
         & D-LFIT & 80.45 & 85.33 & 93.13 & 92.89 \\
         & LF1T & 76.85 & 77.03 & 77.15 & 100 \\
         & JRip & 79.14 & 78.05 & 80.04 & 78.47 \\[3pt]
        Budding (12, 54) & \textbf{NLN} & \textbf{97.03} & \textbf{99.57} & \textbf{100} & \textbf{100} \\
         & NN-LFIT & \textit{ROT} & \textit{ROT} & \textit{ROT} & \textit{ROT} \\
         & D-LFIT & 71.96 & 71.39 & 70.50 & 76.52 \\
         & LF1T & \textit{ROT} & \textit{ROT} & \textit{ROT} & \textit{ROT} \\
         & JRip & 67.97 & 68.55 & 67.91 & 68.35 \\[3pt]
        Arabidopsis (15, 28) & \textbf{NLN} & \textbf{100} & \textbf{100} & \textbf{100} & \textbf{100} \\
         & NN-LFIT & \textit{ROT} & \textit{ROT} & \textit{ROT} & \textit{ROT} \\
         & D-LFIT & 84.35 & 86.83 & 88.56 & 89.70 \\
         & LF1T & \textit{ROT} & \textit{ROT} & \textit{ROT} & \textit{ROT} \\
         & JRip & 68.84 & 69.00 & 68.79 & 68.67 \\
         \bottomrule
    \end{tabular}}
\end{table}

In all four data sets, our method achieves more than 97 \% accuracy with as little as 16 \% of the data. Moreover, it achieves perfect accuracy with only 40 \% of the data. In doing so, our method also discovers the ground-truth boolean networks by correctly identifying all of their necessary rules. In some cases, some of the ground-truth rules are subsumed by the disjunction of other ground-truth rules, making them redundant. In section \ref{sec:app_exp_bool_example}, we present examples of correctly discovered rules as well as incorrectly discovered rules when there is insufficient data. In contrast, the other methods never achieve perfect accuracy even with 80\% of the data. The biggest difficulty for these other methods seem to be the number of rules with the poorest performance being on the budding data set, which has the most rules. This is not an issue for our approach which, being a learning approach, is instead mostly concerned with the amount of data. It performs most poorly on data sets with lower dimensionality for which the same fraction of the full distribution represents a much smaller amount of data. For the arabidopsis data set, which has the largest solution space with 15 variables, our model is able to discover the ground-truth logic program with as little as 8 \% of the full distribution. This ground-truth rule discovery from small amounts of data is one of the strengths of neuro-symbolic methods as opposed to purely neural methods. On the other hand, the purely symbolic extraction step of NN-LFIT and the ILP LF1T method both Run Out of Time (ROT) (5 hours for this task) on the larger solution spaces of the budding and arabidopsis data sets.

\subsubsection{Examples of discovered logic programs}
\label{sec:app_exp_bool_example}

In this first example below, on the mammalian data set, the ground-truth logic program is discovered fully, except for one redundant rule that is subsumed by the disjunction of two other ground-truth rules.

\newpage
\vspace{.5cm}\begin{minipage}[t]{.0\textwidth}\centering
~
\end{minipage}%
\begin{minipage}[t]{.45\textwidth}\centering
Ground-truth logic program
\begin{flalign*}
    A_1^{t} &\rightarrow A_1^{t+1},\\
    \lnot A_3^{t}\land A_4^{t} &\rightarrow A_2^{t+1},\\
    \lnot A_1^{t}\land A_6^{t}\land \lnot A_{10}^{t} &\rightarrow A_3^{t+1},\\
    \lnot A_1^{t}\land \lnot A_2^{t}\land \lnot A_5^{t}\land \lnot A_{10}^{t} &\rightarrow A_3^{t+1},\\
    \lnot A_3^{t}\land A_6^{t}\land \lnot A_{10}^{t} &\rightarrow A_4^{t+1},\\
    \lnot A_3^{t}\land \lnot A_5^{t}\land \lnot A_{10}^{t} &\rightarrow A_4^{t+1},\\
    \lnot A_3^{t}\land A_5^{t}\land \lnot A_7^{t}\land \lnot A_8^{t} &\rightarrow A_5^{t+1},\\
    \lnot A_3^{t}\land A_4^{t}\land \lnot A_7^{t}\land \lnot A_9^{t} &\rightarrow A_5^{t+1},\\
    \lnot A_3^{t}\land A_5^{t}\land \lnot A_7^{t}\land \lnot A_9^{t} &\rightarrow A_5^{t+1},\\
    \lnot A_3^{t}\land A_4^{t}\land \lnot A_7^{t}\land \lnot A_8^{t} &\rightarrow A_5^{t+1},\\
    \lnot A_1^{t}\land \lnot A_2^{t}\land \lnot A_5^{t}\land \lnot A_{10}^{t} &\rightarrow A_6^{t+1},\\
    \lnot A_1^{t}\land \lnot A_5^{t}\land A_6^{t}\land \lnot A_{10}^{t} &\rightarrow A_6^{t+1},\\
    \lnot A_1^{t}\land \lnot A_2^{t}\land A_6^{t}\land \lnot A_{10}^{t} &\rightarrow A_6^{t+1},\\
    A_{10}^{t} &\rightarrow A_7^{t+1},\\
    \lnot A_9^{t} &\rightarrow A_8^{t+1},\\
    A_7^{t}\land A_8^{t} &\rightarrow A_8^{t+1},\\
    A_8^{t}\land A_{10}^{t} &\rightarrow A_8^{t+1},\\
    A_5^{t}\land A_8^{t} &\rightarrow A_8^{t+1},\\
    A_7^{t} &\rightarrow A_9^{t+1},\\
    \lnot A_5^{t}\land \lnot A_{10}^{t} &\rightarrow A_9^{t+1},\\
    A_6^{t}\land A_{10}^{t} &\rightarrow A_9^{t+1},\\
    \lnot A_5^{t}\land A_6^{t} &\rightarrow A_9^{t+1},\\
    \lnot A_7^{t}\land \lnot A_9^{t} &\rightarrow A_{10}^{t+1},\\
\end{flalign*}
\end{minipage}%
\begin{minipage}[t]{.45\textwidth}\centering
Discovered logic program
\begin{flalign*}
    A_1^{t} &\rightarrow A_1^{t+1},\\
    \lnot A_3^{t}\land A_4^{t} &\rightarrow A_2^{t+1},\\
    \lnot A_1^{t}\land A_6^{t}\land \lnot A_{10}^{t} &\rightarrow A_3^{t+1},\\
    \lnot A_1^{t}\land \lnot A_2^{t}\land \lnot A_5^{t}\land \lnot A_{10}^{t} &\rightarrow A_3^{t+1},\\
    \lnot A_3^{t}\land A_6^{t}\land \lnot A_{10}^{t} &\rightarrow A_4^{t+1},\\
    \lnot A_3^{t}\land \lnot A_5^{t}\land \lnot A_{10}^{t} &\rightarrow A_4^{t+1},\\
    \lnot A_3^{t}\land A_5^{t}\land \lnot A_7^{t}\land \lnot A_8^{t} &\rightarrow A_5^{t+1},\\
    \lnot A_3^{t}\land A_4^{t}\land \lnot A_7^{t}\land \lnot A_9^{t} &\rightarrow A_5^{t+1},\\
    \lnot A_3^{t}\land A_5^{t}\land \lnot A_7^{t}\land \lnot A_9^{t} &\rightarrow A_5^{t+1},\\
    \lnot A_3^{t}\land A_4^{t}\land \lnot A_7^{t}\land \lnot A_8^{t} &\rightarrow A_5^{t+1},\\
    \lnot A_1^{t}\land \lnot A_2^{t}\land \lnot A_5^{t}\land \lnot A_{10}^{t} &\rightarrow A_6^{t+1},\\
    \lnot A_1^{t}\land \lnot A_5^{t}\land A_6^{t}\land \lnot A_{10}^{t} &\rightarrow A_6^{t+1},\\
    \lnot A_1^{t}\land \lnot A_2^{t}\land A_6^{t}\land \lnot A_{10}^{t} &\rightarrow A_6^{t+1},\\
    A_{10}^{t} &\rightarrow A_7^{t+1},\\
    \lnot A_9^{t} &\rightarrow A_8^{t+1},\\
    A_7^{t}\land A_8^{t} &\rightarrow A_8^{t+1},\\
    A_8^{t}\land A_{10}^{t} &\rightarrow A_8^{t+1},\\
    A_5^{t}\land A_8^{t} &\rightarrow A_8^{t+1},\\
    A_7^{t} &\rightarrow A_9^{t+1},\\
    \lnot A_5^{t}\land \lnot A_{10}^{t} &\rightarrow A_9^{t+1},\\
    A_6^{t}\land A_{10}^{t} &\rightarrow A_9^{t+1},\\
    \vphantom{\lnot A_5^{t}\land A_6^{t}} &\vphantom{\rightarrow A_9^{t+1},}\\
    \lnot A_7^{t}\land \lnot A_9^{t} &\rightarrow A_{10}^{t+1},\\
\end{flalign*}
\end{minipage}%
\begin{minipage}[t]{.1\textwidth}\centering
~
\end{minipage}
where $\lnot A_5^{t}\land A_6^{t} \rightarrow A_9^{t+1}$ is subsumed by the disjunction of $\lnot A_5^{t}\land \lnot A_{10}^{t} \rightarrow A_9^{t+1}$  and  $A_6^{t}\land A_{10}^{t} \rightarrow A_9^{t+1}$, \ie whenever the first rule should be activated, either the second or the third rule is activated, thus making the first rule redundant.

In this second example below, on the arabidopsis data set, the ground-truth logic program is discovered fully, except for one rule, which was discovered as two rules that imply it by their resolution.

\newpage
\vspace{.5cm}\begin{minipage}[t]{.0\textwidth}\centering
~
\end{minipage}%
\begin{minipage}[t]{.45\textwidth}\centering
Ground-truth logic program
\scalebox{.94}{\parbox{\linewidth}{\begin{flalign*}
    A_2^{t}\land A_7^{t} &\rightarrow A_1^{t+1},\\
    A_1^{t}\land A_5^{t}\land A_{14}^{t}\land A_{15}^{t} &\rightarrow A_1^{t+1},\\
    A_1^{t}\land A_{10}^{t}\land A_{14}^{t}\land A_{15}^{t} &\rightarrow A_1^{t+1},\\
    A_2^{t} &\rightarrow A_2^{t+1},\\
    \lnot A_5^{t}\land \lnot A_{13}^{t} &\rightarrow A_3^{t+1},\\
    \lnot A_6^{t} &\rightarrow A_4^{t+1},\\
    A_4^{t}\land \lnot A_{10}^{t} &\rightarrow A_5^{t+1},\\
    \lnot A_{10}^{t}\land \lnot A_{13}^{t} &\rightarrow A_5^{t+1},\\
    A_7^{t}\land \lnot A_{10}^{t} &\rightarrow A_5^{t+1},\\
    \lnot A_7^{t} &\rightarrow A_6^{t+1},\\
    \lnot A_6^{t} &\rightarrow A_7^{t+1},\\
    \lnot A_{13}^{t} &\rightarrow A_7^{t+1},\\
    \lnot A_{13}^{t} &\rightarrow A_8^{t+1},\\
    A_9^{t}\land \lnot A_{15}^{t} &\rightarrow A_9^{t+1},\\
    A_9^{t}\land \lnot A_{10}^{t} &\rightarrow A_9^{t+1},\\
    \lnot A_5^{t}\land A_7^{t} &\rightarrow A_{10}^{t+1},\\
    A_7^{t}\land \lnot A_{11}^{t} &\rightarrow A_{10}^{t+1},\\
    A_7^{t}\land A_{10}^{t}\land A_{15}^{t} &\rightarrow A_{10}^{t+1},\\
    \lnot A_8^{t}\land \lnot A_{13}^{t} &\rightarrow A_{10}^{t+1},\\
    A_7^{t}\land \lnot A_8^{t} &\rightarrow A_{10}^{t+1},\\
    A_7^{t}\land A_9^{t} &\rightarrow A_{10}^{t+1},\\
    A_7^{t}\land \lnot A_{12}^{t} &\rightarrow A_{10}^{t+1},\\
    \lnot A_5^{t}\land A_6^{t}\land \lnot A_7^{t} &\rightarrow A_{13}^{t+1},\\
    A_1^{t}\land A_5^{t}\land A_{14}^{t}\land A_{15}^{t} &\rightarrow A_{14}^{t+1},\\
    A_1^{t}\land A_{10}^{t}\land A_{14}^{t}\land A_{15}^{t} &\rightarrow A_{14}^{t+1},\\
    A_1^{t}\land A_7^{t} &\rightarrow A_{14}^{t+1},\\
    A_7^{t}\land A_{10}^{t} &\rightarrow A_{14}^{t+1},\\
    A_7^{t} &\rightarrow A_{15}^{t+1},\\
    \vphantom{A_7^{t}\land A_{10}^{t}} &\vphantom{\rightarrow A_{15}^{t+1},}\\
\end{flalign*}}}
\end{minipage}%
\begin{minipage}[t]{.45\textwidth}\centering
Discovered logic program
\scalebox{.94}{\parbox{\linewidth}{\begin{flalign*}
    A_2^{t}\land A_7^{t} &\rightarrow A_1^{t+1},\\
    A_1^{t}\land A_5^{t}\land A_{14}^{t}\land A_{15}^{t} &\rightarrow A_1^{t+1},\\
    A_1^{t}\land A_{10}^{t}\land A_{14}^{t}\land A_{15}^{t} &\rightarrow A_1^{t+1},\\
    A_2^{t} &\rightarrow A_2^{t+1},\\
    \lnot A_5^{t}\land \lnot A_{13}^{t} &\rightarrow A_3^{t+1},\\
    \lnot A_6^{t} &\rightarrow A_4^{t+1},\\
    A_4^{t}\land \lnot A_{10}^{t} &\rightarrow A_5^{t+1},\\
    \lnot A_{10}^{t}\land \lnot A_{13}^{t} &\rightarrow A_5^{t+1},\\
    A_7^{t}\land \lnot A_{10}^{t} &\rightarrow A_5^{t+1},\\
    \lnot A_7^{t} &\rightarrow A_6^{t+1},\\
    \lnot A_6^{t} &\rightarrow A_7^{t+1},\\
    \lnot A_{13}^{t} &\rightarrow A_7^{t+1},\\
    \lnot A_{13}^{t} &\rightarrow A_8^{t+1},\\
    A_9^{t}\land \lnot A_{15}^{t} &\rightarrow A_9^{t+1},\\
    A_9^{t}\land \lnot A_{10}^{t} &\rightarrow A_9^{t+1},\\
    \lnot A_5^{t}\land A_7^{t} &\rightarrow A_{10}^{t+1},\\
    A_7^{t}\land \lnot A_{11}^{t} &\rightarrow A_{10}^{t+1},\\
    A_7^{t}\land A_{10}^{t}\land A_{15}^{t} &\rightarrow A_{10}^{t+1},\\
    \lnot A_8^{t}\land \lnot A_{13}^{t} &\rightarrow A_{10}^{t+1},\\
    A_7^{t}\land \lnot A_8^{t} &\rightarrow A_{10}^{t+1},\\
    A_7^{t}\land A_9^{t} &\rightarrow A_{10}^{t+1},\\
    A_7^{t}\land \lnot A_{12}^{t} &\rightarrow A_{10}^{t+1},\\
    \lnot A_5^{t}\land A_6^{t}\land \lnot A_7^{t} &\rightarrow A_{13}^{t+1},\\
    A_1^{t}\land A_5^{t}\land A_{14}^{t}\land A_{15}^{t} &\rightarrow A_{14}^{t+1},\\
    A_1^{t}\land A_{10}^{t}\land A_{14}^{t}\land A_{15}^{t} &\rightarrow A_{14}^{t+1},\\
    A_1^{t}\land A_7^{t} &\rightarrow A_{14}^{t+1},\\
    A_7^{t}\land A_{10}^{t} &\rightarrow A_{14}^{t+1},\\
    A_7^{t}\land \lnot A_{10}^{t} &\rightarrow A_{15}^{t+1},\\
    A_7^{t}\land A_{10}^{t} &\rightarrow A_{15}^{t+1},\\
\end{flalign*}}}
\end{minipage}%
\begin{minipage}[t]{.1\textwidth}\centering
~
\end{minipage}
where $A_7^{t} \rightarrow A_{15}^{t+1}$ is implied by the resolution of $A_7^{t}\land \lnot A_{10}^{t} \rightarrow A_{15}^{t+1}$  and  $A_7^{t}\land A_{10}^{t} \rightarrow A_{15}^{t+1}$, \ie the first rule is a direct consequence of the second and third rules.

In this third example below, on the budding data set, we again have redundant ground-truth rules which are subsumed by the disjunction of two other discovered rules. These have no impact on the predictive accuracy of the model. In this case however, we also have two missing ground-truth rules that are instead incorrectly discovered as a single more general rule.
\vspace{.5cm}\begin{minipage}[t]{.03\textwidth}\centering
~
\end{minipage}%
\begin{minipage}[t]{.45\textwidth}\centering
Ground-truth logic program
\scalebox{.59}{\parbox{\linewidth}{\begin{flalign*}
    A_1^{t} &\rightarrow A_2^{t+1},\\
    A_2^{t}\land A_3^{t} &\rightarrow A_3^{t+1},\\
    A_3^{t}\land \lnot A_9^{t} &\rightarrow A_3^{t+1},\\
    A_2^{t}\land \lnot A_9^{t} &\rightarrow A_3^{t+1},\\
    A_2^{t}\land \lnot A_9^{t} &\rightarrow A_4^{t+1},\\
    A_4^{t}\land \lnot A_9^{t} &\rightarrow A_4^{t+1},\\
    A_2^{t}\land A_4^{t} &\rightarrow A_4^{t+1},\\
    A_3^{t} &\rightarrow A_5^{t+1},\\
    \lnot A_5^{t}\land A_6^{t}\land A_{11}^{t}\land A_{12}^{t} &\rightarrow A_6^{t+1},\\
    \lnot A_5^{t}\land A_6^{t}\land \lnot A_7^{t}\land A_{11}^{t} &\rightarrow A_6^{t+1},\\
    \lnot A_5^{t}\land A_6^{t}\land \lnot A_9^{t}\land A_{11}^{t} &\rightarrow A_6^{t+1},\\
    A_6^{t}\land \lnot A_7^{t}\land \lnot A_9^{t}\land A_{11}^{t} &\rightarrow A_6^{t+1},\\
    A_6^{t}\land \lnot A_9^{t}\land A_{11}^{t}\land A_{12}^{t} &\rightarrow A_6^{t+1},\\
    \lnot A_7^{t}\land \lnot A_9^{t}\land A_{11}^{t}\land A_{12}^{t} &\rightarrow A_6^{t+1},\\
    \lnot A_5^{t}\land A_6^{t}\land \lnot A_7^{t}\land A_{12}^{t} &\rightarrow A_6^{t+1},\\
    A_6^{t}\land \lnot A_7^{t}\land A_{11}^{t}\land A_{12}^{t} &\rightarrow A_6^{t+1},\\
    A_6^{t}\land \lnot A_7^{t}\land \lnot A_9^{t}\land A_{12}^{t} &\rightarrow A_6^{t+1},\\
    \lnot A_5^{t}\land \lnot A_7^{t}\land \lnot A_9^{t}\land A_{11}^{t} &\rightarrow A_6^{t+1},\\
    \lnot A_5^{t}\land A_6^{t}\land \lnot A_7^{t}\land \lnot A_9^{t} &\rightarrow A_6^{t+1},\\
    \lnot A_5^{t}\land \lnot A_7^{t}\land A_{11}^{t}\land A_{12}^{t} &\rightarrow A_6^{t+1},\\
    \lnot A_5^{t}\land \lnot A_9^{t}\land \lnot A_{11}^{t}\land A_{12}^{t} &\rightarrow A_6^{t+1},\\
    \lnot A_5^{t}\land \lnot A_7^{t}\land \lnot A_9^{t}\land A_{12}^{t} &\rightarrow A_6^{t+1},\\
    \lnot A_5^{t}\land A_6^{t}\land \lnot A_9^{t}\land A_{12}^{t} &\rightarrow A_6^{t+1},\\
    A_4^{t}\land A_7^{t}\land \lnot A_{11}^{t} &\rightarrow A_7^{t+1},\\
    \lnot A_6^{t}\land A_7^{t}\land \lnot A_{11}^{t} &\rightarrow A_7^{t+1},\\
    A_4^{t}\land \lnot A_6^{t}\land A_7^{t} &\rightarrow A_7^{t+1},\\
    A_4^{t}\land \lnot A_6^{t}\land \lnot A_{11}^{t} &\rightarrow A_7^{t+1},\\
    \lnot A_5^{t}\land \lnot A_7^{t}\land A_8^{t}\land A_{11}^{t} &\rightarrow A_8^{t+1},\\
    \lnot A_5^{t}\land A_8^{t}\land \lnot A_9^{t}\land A_{11}^{t} &\rightarrow A_8^{t+1},\\
    \lnot A_7^{t}\land A_8^{t}\land \lnot A_9^{t}\land A_{11}^{t} &\rightarrow A_8^{t+1},\\
    \lnot A_5^{t}\land \lnot A_7^{t}\land A_8^{t}\land \lnot A_9^{t} &\rightarrow A_8^{t+1},\\
    \lnot A_5^{t}\land \lnot A_7^{t}\land \lnot A_9^{t}\land A_{11}^{t} &\rightarrow A_8^{t+1},\\
    \lnot A_6^{t}\land A_7^{t}\land \lnot A_8^{t}\land \lnot A_{11}^{t} &\rightarrow A_9^{t+1},\\
    A_7^{t}\land A_9^{t}\land A_{10}^{t}\land \lnot A_{11}^{t} &\rightarrow A_9^{t+1},\\
    \lnot A_8^{t}\land A_9^{t}\land A_{10}^{t}\land \lnot A_{11}^{t} &\rightarrow A_9^{t+1},\\
    \lnot A_6^{t}\land A_7^{t}\land A_{10}^{t}\land \lnot A_{11}^{t} &\rightarrow A_9^{t+1},\\
    A_7^{t}\land \lnot A_8^{t}\land A_9^{t}\land A_{10}^{t} &\rightarrow A_9^{t+1},\\
    A_7^{t}\land \lnot A_8^{t}\land A_9^{t}\land \lnot A_{11}^{t} &\rightarrow A_9^{t+1},\\
    \lnot A_6^{t}\land A_7^{t}\land \lnot A_8^{t}\land A_9^{t} &\rightarrow A_9^{t+1},\\
    \lnot A_6^{t}\land \lnot A_8^{t}\land A_{10}^{t}\land \lnot A_{11}^{t} &\rightarrow A_9^{t+1},\\
    \lnot A_6^{t}\land A_7^{t}\land \lnot A_8^{t}\land A_{10}^{t} &\rightarrow A_9^{t+1},\\
    A_7^{t}\land \lnot A_8^{t}\land A_{10}^{t}\land \lnot A_{11}^{t} &\rightarrow A_9^{t+1},\\
    \lnot A_6^{t}\land A_9^{t}\land A_{10}^{t}\land \lnot A_{11}^{t} &\rightarrow A_9^{t+1},\\
    \lnot A_6^{t}\land \lnot A_8^{t}\land A_9^{t}\land A_{10}^{t} &\rightarrow A_9^{t+1},\\
    \lnot A_6^{t}\land \lnot A_8^{t}\land A_9^{t}\land \lnot A_{11}^{t} &\rightarrow A_9^{t+1},\\
    \lnot A_6^{t}\land A_7^{t}\land A_9^{t}\land A_{10}^{t} &\rightarrow A_9^{t+1},\\
    \lnot A_6^{t}\land A_7^{t}\land A_9^{t}\land \lnot A_{11}^{t} &\rightarrow A_9^{t+1},\\
    A_7^{t} &\rightarrow A_{10}^{t+1},\\
    A_9^{t} &\rightarrow A_{10}^{t+1},\\
    A_9^{t} &\rightarrow A_{11}^{t+1},\\
    A_{10}^{t} &\rightarrow A_{11}^{t+1},\\
    \lnot A_9^{t}\land A_{11}^{t} &\rightarrow A_{12}^{t+1},\\
    A_{10}^{t}\land A_{11}^{t} &\rightarrow A_{12}^{t+1},\\
    \lnot A_9^{t}\land A_{10}^{t} &\rightarrow A_{12}^{t+1},\\
\end{flalign*}}}
\end{minipage}%
\begin{minipage}[t]{.45\textwidth}\centering
Discovered logic program
\scalebox{.59}{\parbox{\linewidth}{\begin{flalign*}
    A_1^{t} &\rightarrow A_2^{t+1},\\
    A_2^{t}\land A_3^{t} &\rightarrow A_3^{t+1},\\
    A_3^{t}\land \lnot A_9^{t} &\rightarrow A_3^{t+1},\\
    A_2^{t}\land \lnot A_9^{t} &\rightarrow A_3^{t+1},\\
    A_2^{t}\land \lnot A_9^{t} &\rightarrow A_4^{t+1},\\
    A_4^{t}\land \lnot A_9^{t} &\rightarrow A_4^{t+1},\\
    A_2^{t}\land A_4^{t} &\rightarrow A_4^{t+1},\\
    A_3^{t} &\rightarrow A_5^{t+1},\\
    \lnot A_5^{t}\land A_6^{t}\land A_{11}^{t}\land A_{12}^{t} &\rightarrow A_6^{t+1},\\
    \lnot A_5^{t}\land A_6^{t}\land \lnot A_7^{t}\land A_{11}^{t} &\rightarrow A_6^{t+1},\\
    \lnot A_5^{t}\land A_6^{t}\land \lnot A_9^{t}\land A_{11}^{t} &\rightarrow A_6^{t+1},\\
    A_6^{t}\land \lnot A_7^{t}\land \lnot A_9^{t}\land A_{11}^{t} &\rightarrow A_6^{t+1},\\
    A_6^{t}\land \lnot A_9^{t}\land A_{11}^{t}\land A_{12}^{t} &\rightarrow A_6^{t+1},\\
    \lnot A_7^{t}\land \lnot A_9^{t}\land A_{11}^{t}\land A_{12}^{t} &\rightarrow A_6^{t+1},\\
    \lnot A_5^{t}\land A_6^{t}\land \lnot A_7^{t}\land A_{12}^{t} &\rightarrow A_6^{t+1},\\
    A_6^{t}\land \lnot A_7^{t}\land A_{11}^{t}\land A_{12}^{t} &\rightarrow A_6^{t+1},\\
    A_6^{t}\land \lnot A_7^{t}\land \lnot A_9^{t}\land A_{12}^{t} &\rightarrow A_6^{t+1},\\
    \lnot A_5^{t}\land \lnot A_7^{t}\land \lnot A_9^{t}\land A_{11}^{t} &\rightarrow A_6^{t+1},\\
    \lnot A_5^{t}\land A_6^{t}\land \lnot A_7^{t}\land \lnot A_9^{t} &\rightarrow A_6^{t+1},\\
    \lnot A_5^{t}\land \lnot A_7^{t}\land A_{11}^{t}\land A_{12}^{t} &\rightarrow A_6^{t+1},\\
    \lnot A_5^{t}\land \lnot A_9^{t}\land \lnot A_{11}^{t}\land A_{12}^{t} &\rightarrow A_6^{t+1},\\
    \vphantom{\lnot A_5^{t}\land \lnot A_7^{t}\land \lnot A_9^{t}\land A_{12}^{t}} &\vphantom{\rightarrow A_6^{t+1},}\\
    \vphantom{\lnot A_5^{t}\land A_6^{t}\land \lnot A_9^{t}\land A_{12}^{t}} &\vphantom{\rightarrow A_6^{t+1},}\\
    A_4^{t}\land A_7^{t}\land \lnot A_{11}^{t} &\rightarrow A_7^{t+1},\\
    \lnot A_6^{t}\land A_7^{t}\land \lnot A_{11}^{t} &\rightarrow A_7^{t+1},\\
    A_4^{t}\land \lnot A_6^{t}\land A_7^{t} &\rightarrow A_7^{t+1},\\
    A_4^{t}\land \lnot A_6^{t}\land \lnot A_{11}^{t} &\rightarrow A_7^{t+1},\\
    \lnot A_5^{t}\land \lnot A_7^{t}\land A_8^{t}\land A_{11}^{t} &\rightarrow A_8^{t+1},\\
    \lnot A_5^{t}\land A_8^{t}\land \lnot A_9^{t}\land A_{11}^{t} &\rightarrow A_8^{t+1},\\
    \lnot A_7^{t}\land A_8^{t}\land \lnot A_9^{t}\land A_{11}^{t} &\rightarrow A_8^{t+1},\\
    \lnot A_5^{t}\land \lnot A_7^{t}\land A_8^{t}\land \lnot A_9^{t} &\rightarrow A_8^{t+1},\\
    \lnot A_5^{t}\land \lnot A_7^{t}\land \lnot A_9^{t}\land A_{11}^{t} &\rightarrow A_8^{t+1},\\
    \lnot A_6^{t}\land A_7^{t}\land \lnot A_8^{t}\land \lnot A_{11}^{t} &\rightarrow A_9^{t+1},\\
    A_7^{t}\land A_9^{t}\land A_{10}^{t}\land \lnot A_{11}^{t} &\rightarrow A_9^{t+1},\\
    \lnot A_8^{t}\land A_9^{t}\land A_{10}^{t}\land \lnot A_{11}^{t} &\rightarrow A_9^{t+1},\\
    \lnot A_6^{t}\land A_7^{t}\land A_{10}^{t}\land \lnot A_{11}^{t} &\rightarrow A_9^{t+1},\\
    A_7^{t}\land \lnot A_8^{t}\land A_9^{t}\land A_{10}^{t} &\rightarrow A_9^{t+1},\\
    A_7^{t}\land \lnot A_8^{t}\land A_9^{t}\land \lnot A_{11}^{t} &\rightarrow A_9^{t+1},\\
    \lnot A_6^{t}\land A_7^{t}\land \lnot A_8^{t}\land A_9^{t} &\rightarrow A_9^{t+1},\\
    \lnot A_6^{t}\land \lnot A_8^{t}\land A_{10}^{t}\land \lnot A_{11}^{t} &\rightarrow A_9^{t+1},\\
    \lnot A_6^{t}\land A_7^{t}\land \lnot A_8^{t}\land A_{10}^{t} &\rightarrow A_9^{t+1},\\
    A_7^{t}\land \lnot A_8^{t}\land A_{10}^{t}\land \lnot A_{11}^{t} &\rightarrow A_9^{t+1},\\
    \lnot A_6^{t}\land A_9^{t}\land A_{10}^{t}\land \lnot A_{11}^{t} &\rightarrow A_9^{t+1},\\
    \lnot A_6^{t}\land \lnot A_8^{t}\land A_9^{t}\land A_{10}^{t} &\rightarrow A_9^{t+1},\\
    \lnot A_6^{t}\land \lnot A_8^{t}\land A_9^{t}\land \lnot A_{11}^{t} &\rightarrow A_9^{t+1},\\
    A_6^{t}\land A_7^{t}\land A_9^{t} &\rightarrow A_9^{t+1},\\
    \vphantom{A_6^{t}\land A_7^{t}\land A_9^{t}\land \lnot A_{11}^{t}} &\vphantom{\rightarrow A_9^{t+1},}\\
    A_7^{t} &\rightarrow A_{10}^{t+1},\\
    A_9^{t} &\rightarrow A_{10}^{t+1},\\
    A_9^{t} &\rightarrow A_{11}^{t+1},\\
    A_{10}^{t} &\rightarrow A_{11}^{t+1},\\
    \lnot A_9^{t}\land A_{11}^{t} &\rightarrow A_{12}^{t+1},\\
    A_{10}^{t}\land A_{11}^{t} &\rightarrow A_{12}^{t+1},\\
    \lnot A_9^{t}\land A_{10}^{t} &\rightarrow A_{12}^{t+1},\\
\end{flalign*}}}
\end{minipage}%
\begin{minipage}[t]{.07\textwidth}\centering
~
\end{minipage}
where
\begin{itemize}
    \item $\lnot A_5^{t}\land \lnot A_7^{t}\land \lnot A_9^{t}\land A_{12}^{t} \rightarrow A_6^{t+1}$ is correctly subsumed by  $\lnot A_5^{t}\land \lnot A_7^{t}\land A_{11}^{t}\land A_{12}^{t} \rightarrow A_6^{t+1}$  and  $\lnot A_5^{t}\land \lnot A_9^{t}\land \lnot A_{11}^{t}\land A_{12}^{t} \rightarrow A_6^{t+1}$,
    
    \item $\lnot A_5^{t}\land A_6^{t}\land \lnot A_9^{t}\land A_{12}^{t} \rightarrow A_6^{t+1}$ is correctly subsumed by  $\lnot A_5^{t}\land A_6^{t}\land \lnot A_9^{t}\land A_{11}^{t} \rightarrow A_6^{t+1}$  and  $\lnot A_5^{t}\land \lnot A_9^{t}\land \lnot A_{11}^{t}\land A_{12}^{t} \rightarrow A_6^{t+1}$,
    
    \item but $\lnot A_6^{t}\land A_7^{t}\land A_9^{t}\land A_{10}^{t} \rightarrow A_9^{t+1}$ and $\lnot A_6^{t}\land A_7^{t}\land A_9^{t}\land \lnot A_{11}^{t} \rightarrow A_9^{t+1}$ are incorrectly discovered as the more general $\lnot A_6^{t}\land A_7^{t}\land A_9^{t} \rightarrow A_9^{t+1}$, \ie the third rule covers all the cases when the first two rules are activated, but it is also activated in other cases where it should not.
\end{itemize}

\subsection{Medical application}

\subsubsection{Details on the merging and and finding of the minimal RRL network}
\label{sec:app_exp_med_RRL}

Multiple RRL models, like with NLNs, can be merged into a single model. To do so, we concatenated the rules of the five models into a single layer and copied their respective weights in the linear output layer. The biases of the linear output layer were averaged over the five input models. We then pruned the rules of this merged model by using the same procedure than the NLNs but only on the linear output layer. By doing this the initial 175 rules were reduced to 62 rules.

With 62 rules, finding the smallest sub-combination of rules that perfectly classifies the data set exhaustively proved to be impossible in practice. We opted to prune the model again, but, instead of using the loss to evaluate if a parameter was necessary, we directly used the f1 classification score. This reduced the model to 16 rules, of which 13 were distinct from one another.

A small caveat, there were in fact $2\,838$ rules in the five initial models, but $2\,663$ of them were empty. In the pruning on the loss, 284 of these empty rules were kept. After the final pruning on the f1 score, 20 empty rules remained. These empty rules (specifically the AND rules), although they were empty, still contributed to the biases of the linear layer with their weight. As such, in the final shown model, the biases were updated accordingly and then the linear layer was normalized.

\subsection{Industrial application}

\subsubsection{Rules found for the NSL-KDD data set}
\label{sec:app_exp_KDD}

% \newpage
\begin{figure}[p]
    \centering
    \parbox[c]{0.475\textwidth}{
     \begin{subfigure}[b]{0.475\textwidth}
         \centering
         \includegraphics[width=\textwidth]{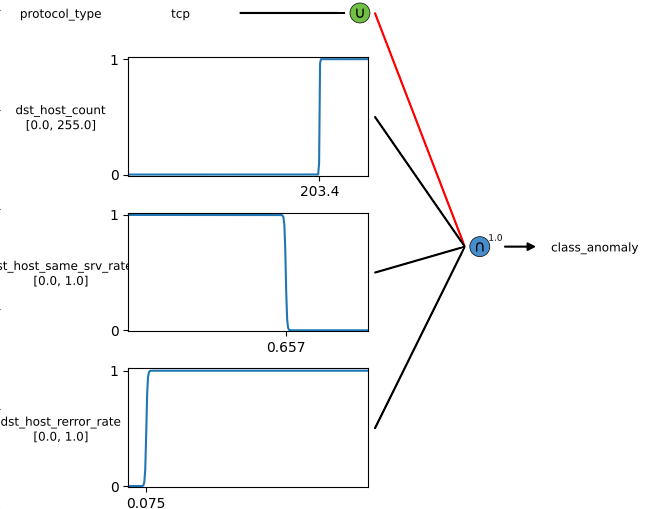}
         \caption{1$^{\text{st}}$ rule (covers 2 \% of anomalies)}
         \label{fig:interp_OR1}
     \end{subfigure}

     \vspace{12pt}
     
     \begin{subfigure}[b]{0.39\textwidth}
         \centering
         \includegraphics[width=\textwidth]{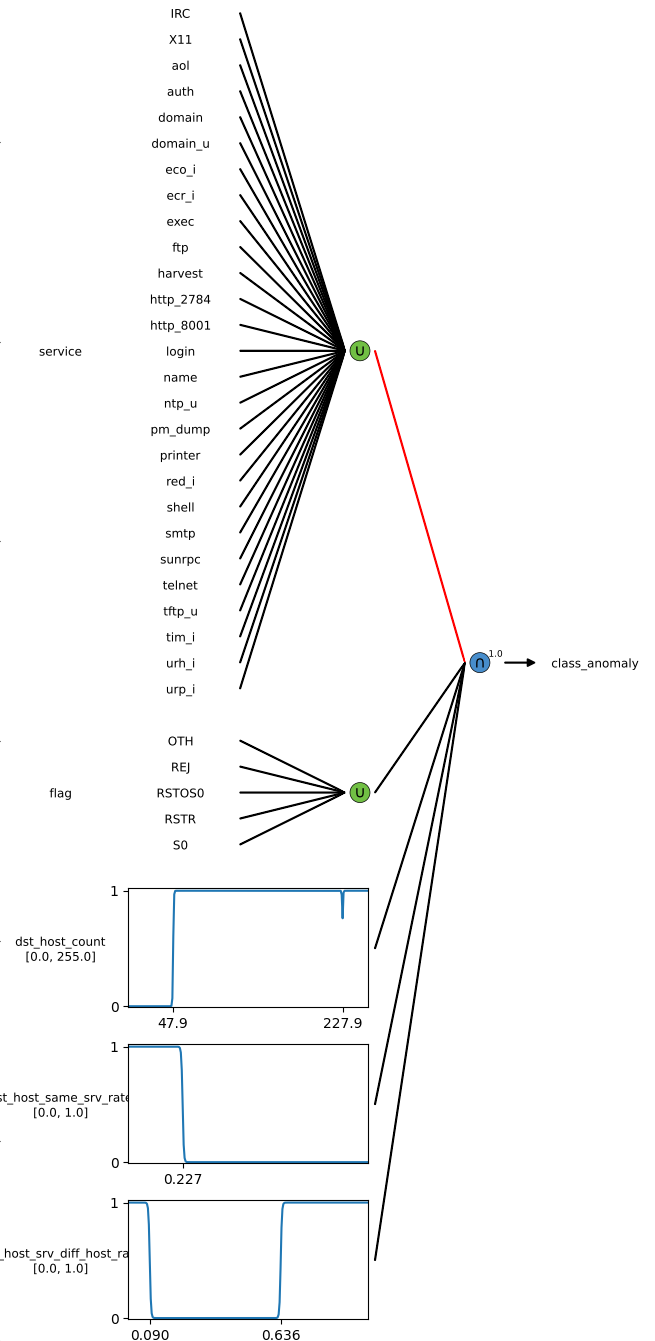}
         \caption{2$^{\text{nd}}$ rule (covers 61 \% of anomalies)}
         \label{fig:interp_AND1}
     \end{subfigure}
     }
     \hfill
     \parbox[c]{0.475\textwidth}{
     \begin{subfigure}[b]{0.45\textwidth}
         \centering
         \includegraphics[width=\textwidth]{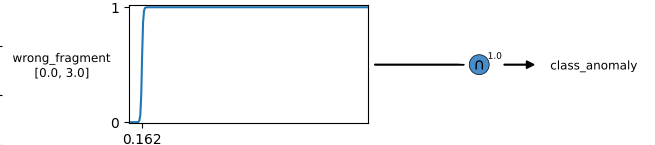}
         \caption{3$^{\text{rd}}$ rule (covers 2 \% of anomalies)}
         \label{fig:interp_AND1}
     \end{subfigure}

     \vspace{12pt}
     
     \begin{subfigure}[b]{0.475\textwidth}
         \centering
         \includegraphics[width=\textwidth]{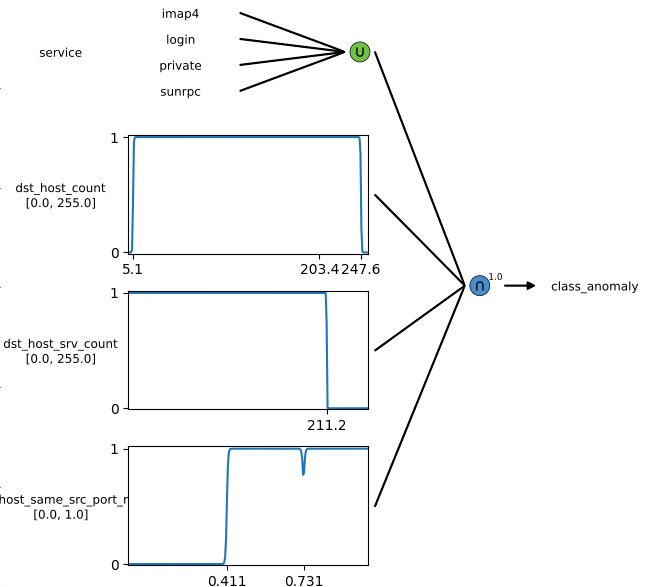}
         \caption{4$^{\text{th}}$ rule (covers 1 \% of anomalies)}
         \label{fig:interp_AND1}
     \end{subfigure}

     \vspace{12pt}
     
     \begin{subfigure}[b]{0.475\textwidth}
         \centering
         \includegraphics[width=\textwidth]{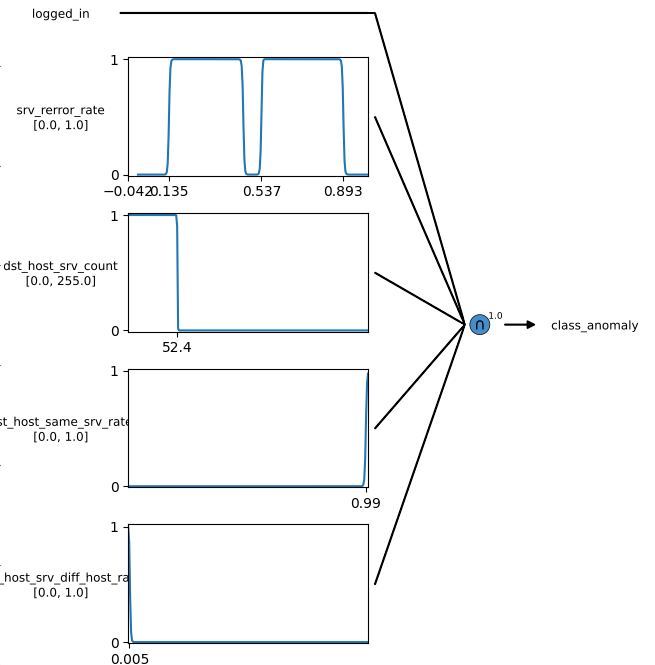}
         \caption{5$^{\text{th}}$ rule (covers <1 \% of anomalies)}
         \label{fig:interp_AND1}
     \end{subfigure}
     }
\end{figure}

% \newpage
\begin{figure}[p]
    \centering
    \parbox[c]{0.475\textwidth}{
     \begin{subfigure}[b]{0.45\textwidth}
         \centering
         \includegraphics[width=\textwidth]{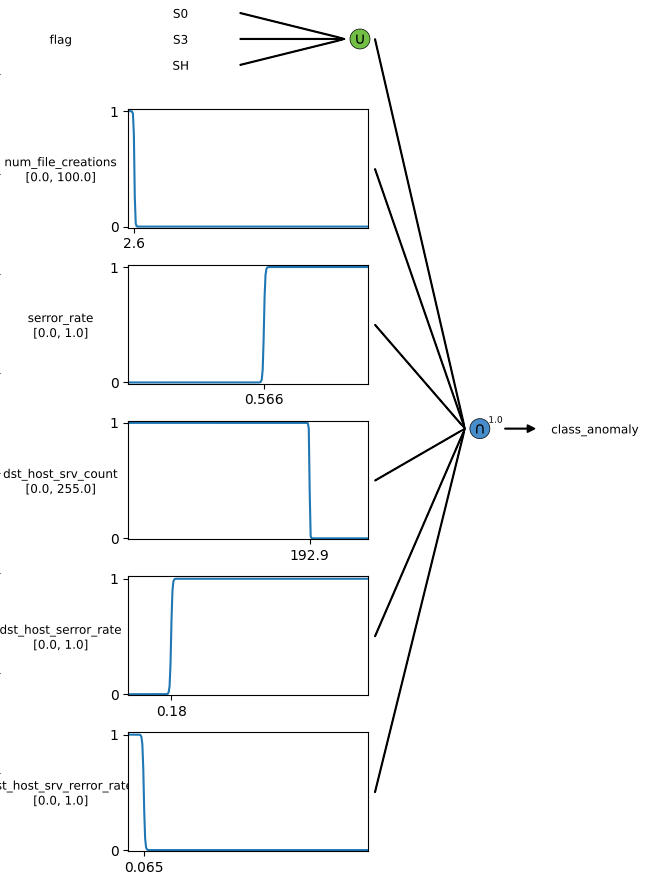}
         \caption{6$^{\text{th}}$ rule (covers 51 \% of anomalies)}
         \label{fig:interp_OR1}
     \end{subfigure}

     \vspace{12pt}
     
     \begin{subfigure}[b]{0.45\textwidth}
         \centering
         \includegraphics[width=\textwidth]{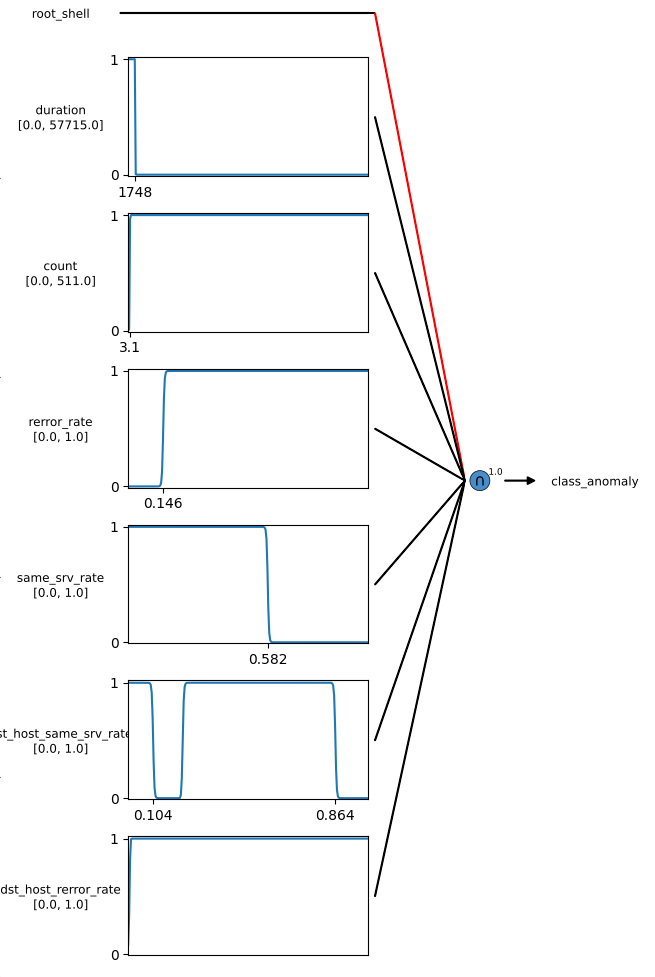}
         \caption{7$^{\text{th}}$ rule (covers 19 \% of anomalies)}
         \label{fig:interp_OR1}
     \end{subfigure}
     }
     \hfill
     \parbox[c]{0.475\textwidth}{
     \begin{subfigure}[b]{0.475\textwidth}
         \centering
         \includegraphics[width=\textwidth]{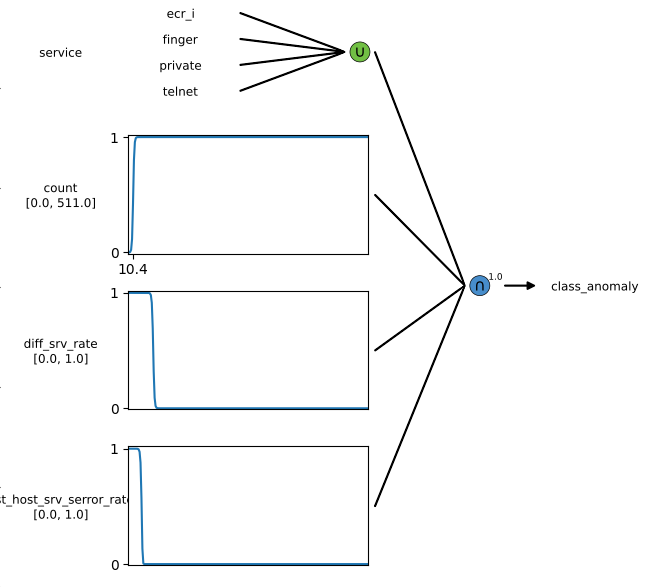}
         \caption{8$^{\text{th}}$ rule (covers 13 \% of anomalies)}
         \label{fig:interp_OR1}
     \end{subfigure}

     \vspace{12pt}
     
     \begin{subfigure}[b]{0.475\textwidth}
         \centering
         \includegraphics[width=\textwidth]{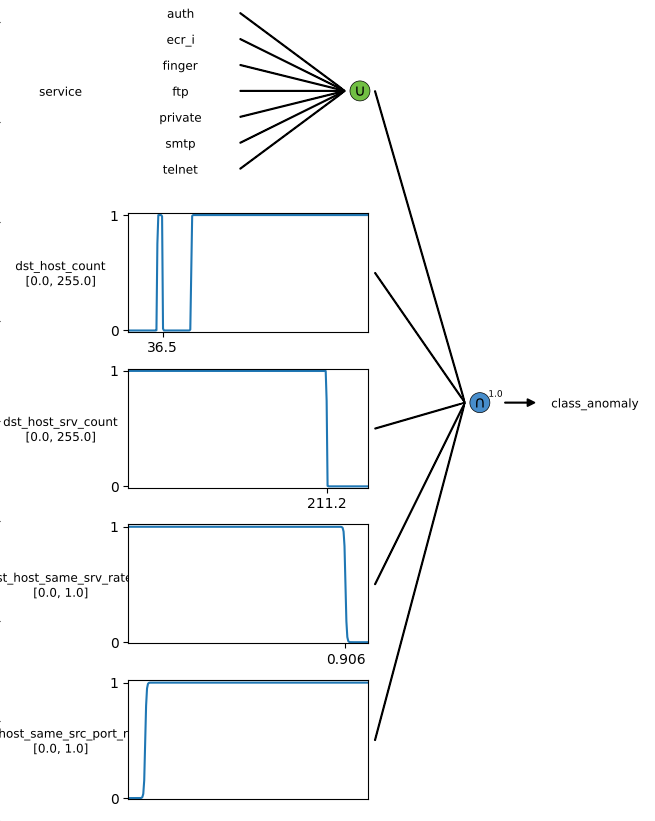}
         \caption{9$^{\text{th}}$ rule (covers 8 \% of anomalies)}
         \label{fig:interp_OR1}
     \end{subfigure}
     }
\end{figure}

% \newpage
\begin{figure}[p]
    \centering
    \parbox[c]{0.475\textwidth}{
     \begin{subfigure}[b]{0.4\textwidth}
         \centering
         \includegraphics[width=\textwidth]{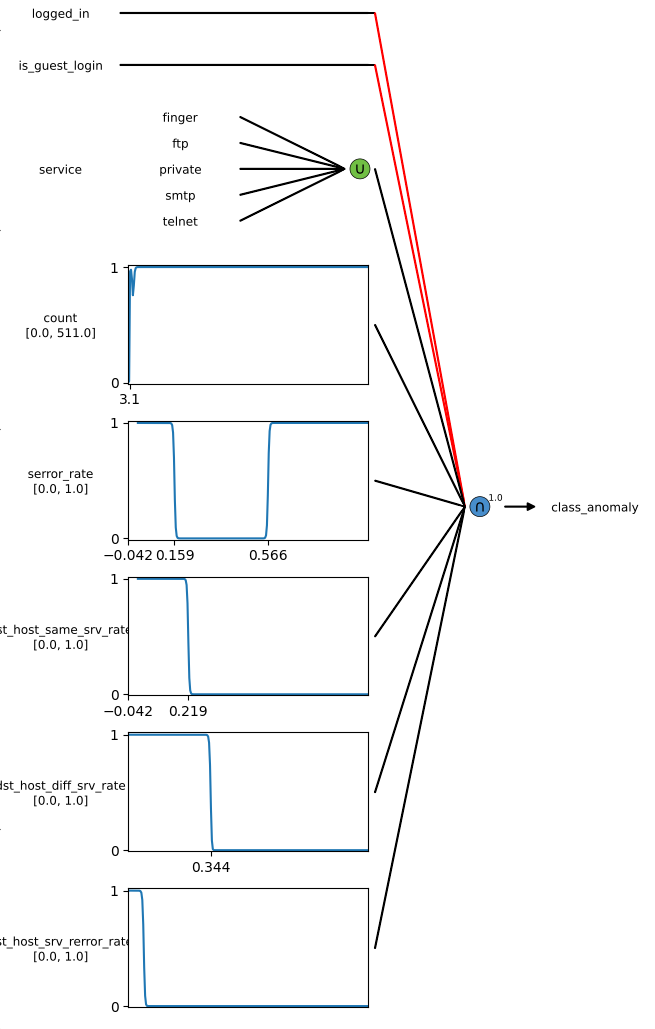}
         \caption{10$^{\text{th}}$ rule (covers 22 \% of anomalies)}
         \label{fig:interp_OR1}
     \end{subfigure}

     \vspace{12pt}
     
     \begin{subfigure}[b]{0.4\textwidth}
         \centering
         \includegraphics[width=\textwidth]{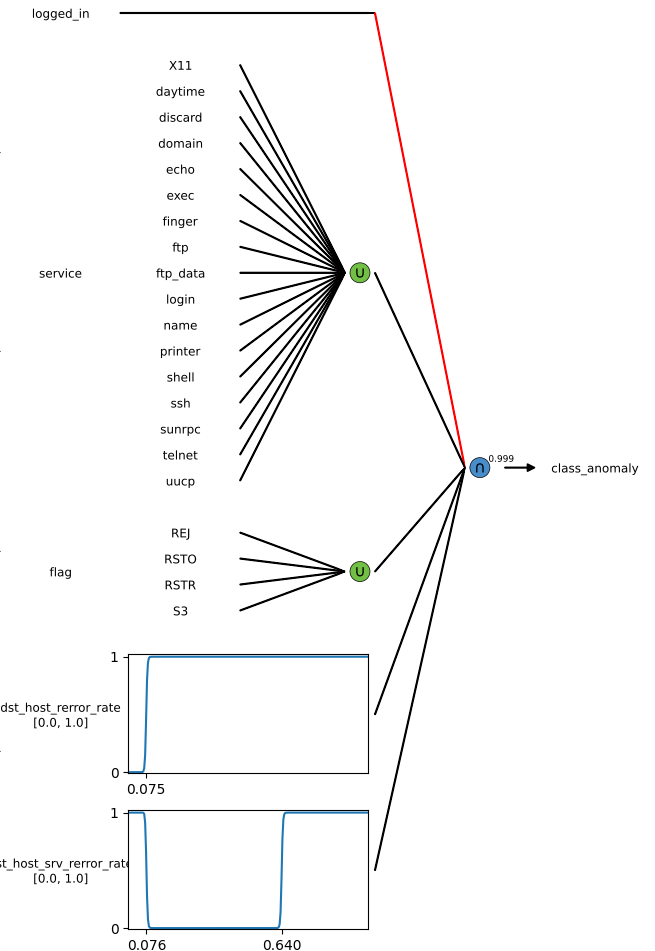}
         \caption{11$^{\text{th}}$ rule (covers 3 \% of anomalies)}
         \label{fig:interp_OR1}
     \end{subfigure}
     }
     \hfill
     \parbox[c]{0.475\textwidth}{
     \begin{subfigure}[b]{0.4\textwidth}
         \centering
         \includegraphics[width=\textwidth]{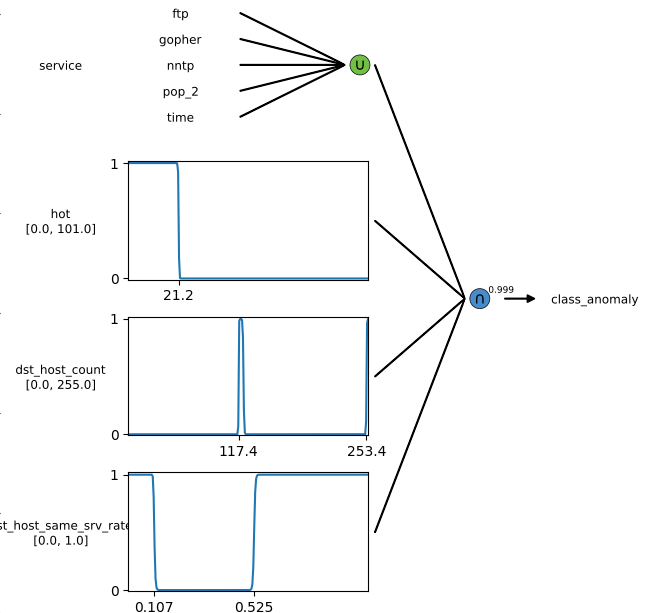}
         \caption{12$^{\text{th}}$ rule (covers 3 \% of anomalies)}
         \label{fig:interp_OR1}
     \end{subfigure}

     \vspace{12pt}
     
     \begin{subfigure}[b]{0.4\textwidth}
         \centering
         \includegraphics[width=\textwidth]{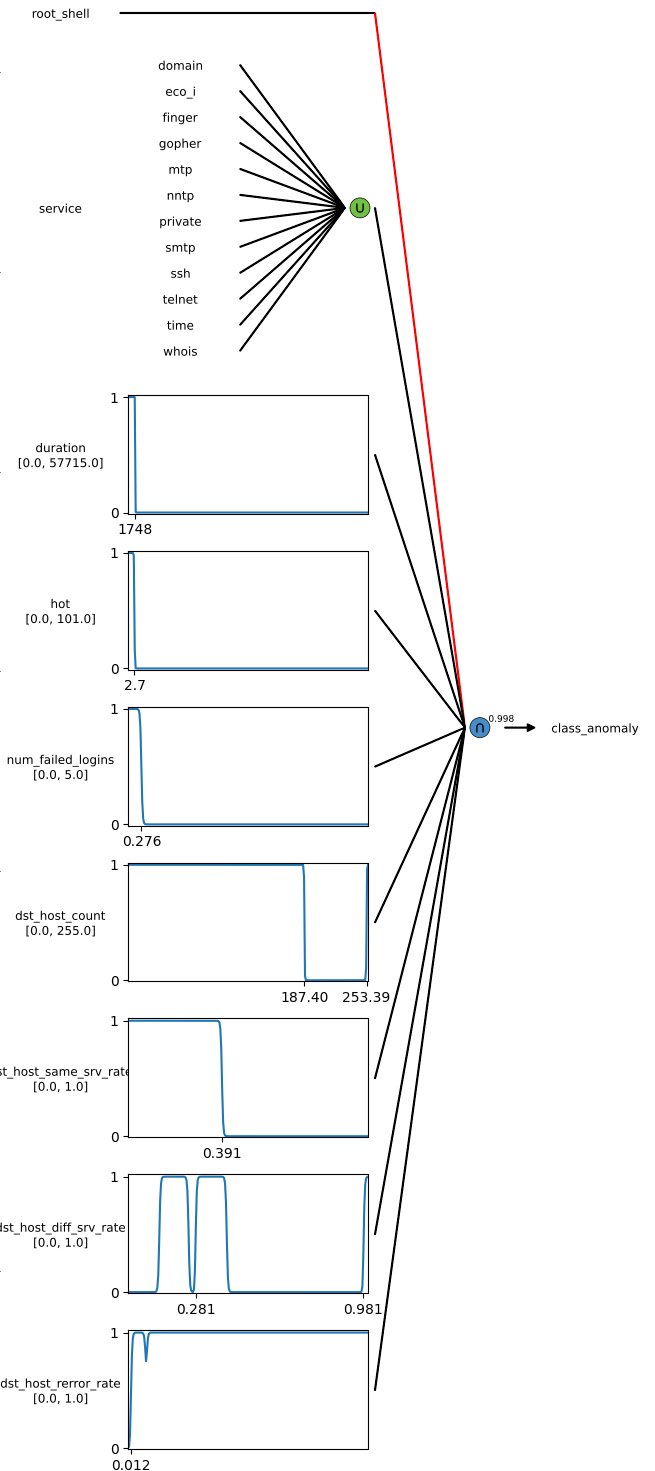}
         \caption{13$^{\text{th}}$ rule (covers 3 \% of anomalies)}
         \label{fig:interp_OR1}
     \end{subfigure}
     }
\end{figure}

% \newpage
\begin{figure}[p]
    \centering
    \parbox[c]{0.475\textwidth}{
     \begin{subfigure}[b]{0.475\textwidth}
         \centering
         \includegraphics[width=\textwidth]{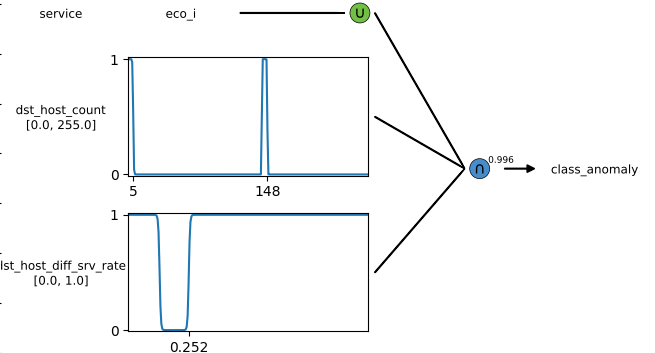}
         \caption{14$^{\text{th}}$ rule (covers 6 \% of anomalies)}
         \label{fig:interp_OR1}
     \end{subfigure}

     \vspace{12pt}
     
     \begin{subfigure}[b]{0.475\textwidth}
         \centering
         \includegraphics[width=\textwidth]{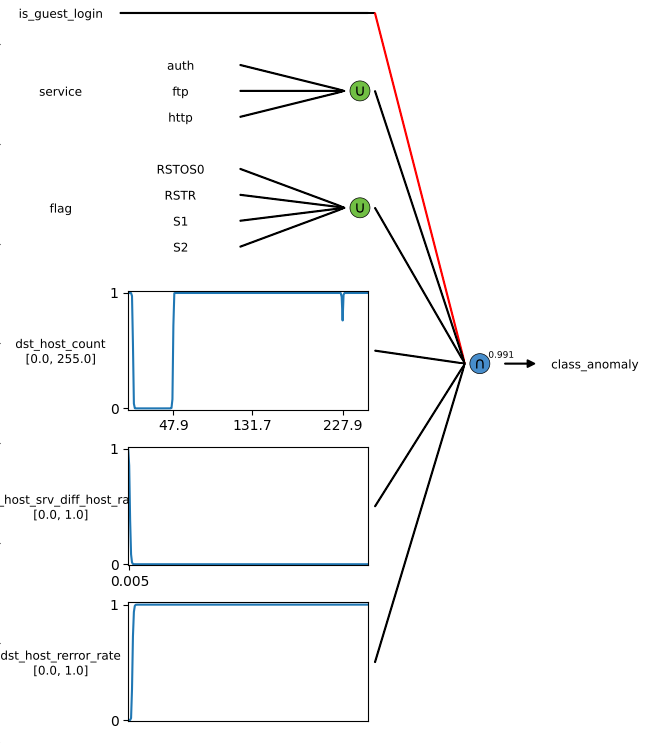}
         \caption{15$^{\text{th}}$ rule (covers 1 \% of anomalies)}
         \label{fig:interp_OR1}
     \end{subfigure}
     }
     \hfill
     \parbox[c]{0.475\textwidth}{
     \begin{subfigure}[b]{0.475\textwidth}
         \centering
         \includegraphics[width=\textwidth]{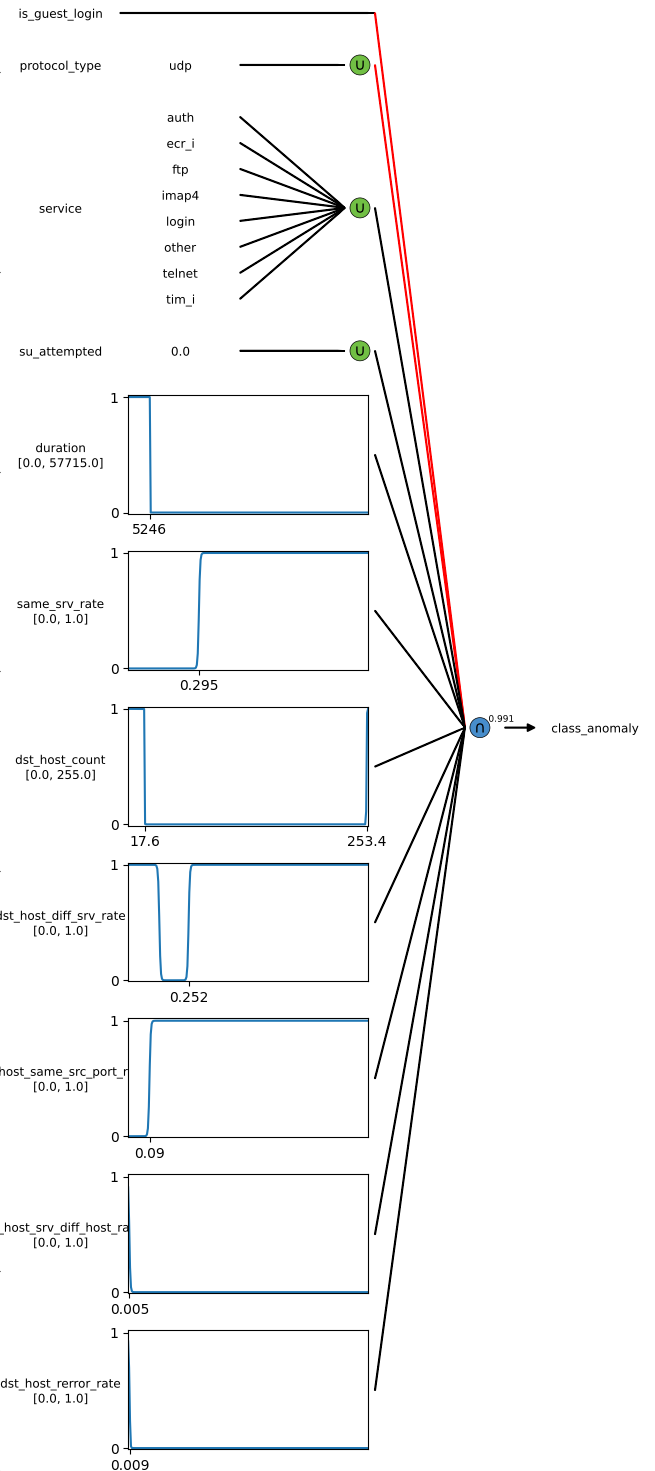}
         \caption{16$^{\text{th}}$ rule (covers 3 \% of anomalies)}
         \label{fig:interp_OR1}
     \end{subfigure}
     }
\end{figure}

% \newpage
\begin{figure}[p]
    \centering
    \parbox[c]{0.475\textwidth}{
     \begin{subfigure}[b]{0.475\textwidth}
         \centering
         \includegraphics[width=\textwidth]{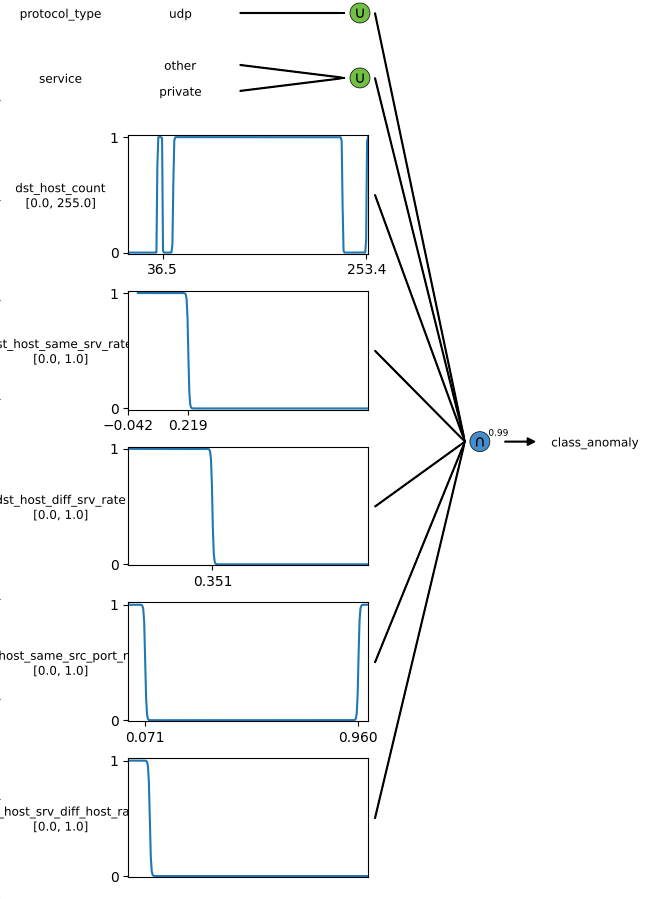}
         \caption{17$^{\text{th}}$ rule (covers 1 \% of anomalies)}
         \label{fig:interp_OR1}
     \end{subfigure}

     \vspace{12pt}
     
     \begin{subfigure}[b]{0.475\textwidth}
         \centering
         \includegraphics[width=\textwidth]{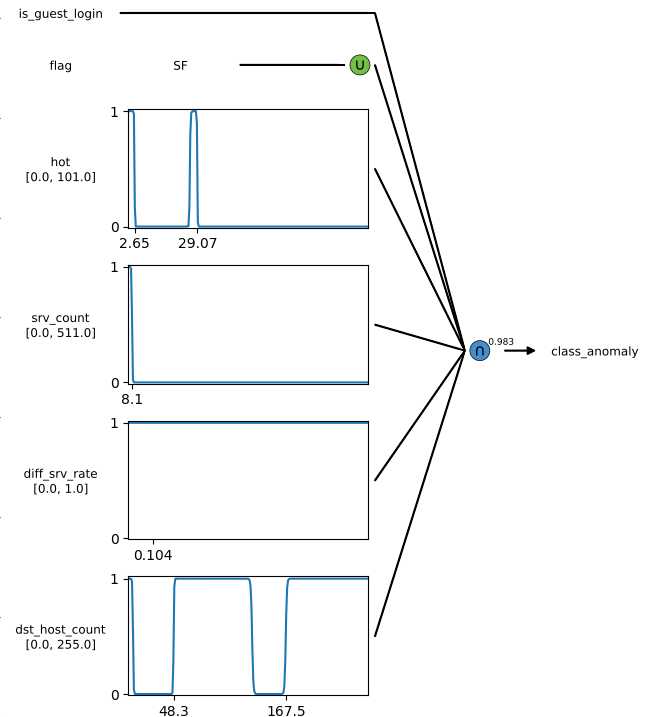}
         \caption{18$^{\text{th}}$ rule (covers 1 \% of anomalies)}
         \label{fig:interp_OR1}
     \end{subfigure}
     }
     \hfill
     \parbox[c]{0.475\textwidth}{
     \begin{subfigure}[b]{0.475\textwidth}
         \centering
         \includegraphics[width=\textwidth]{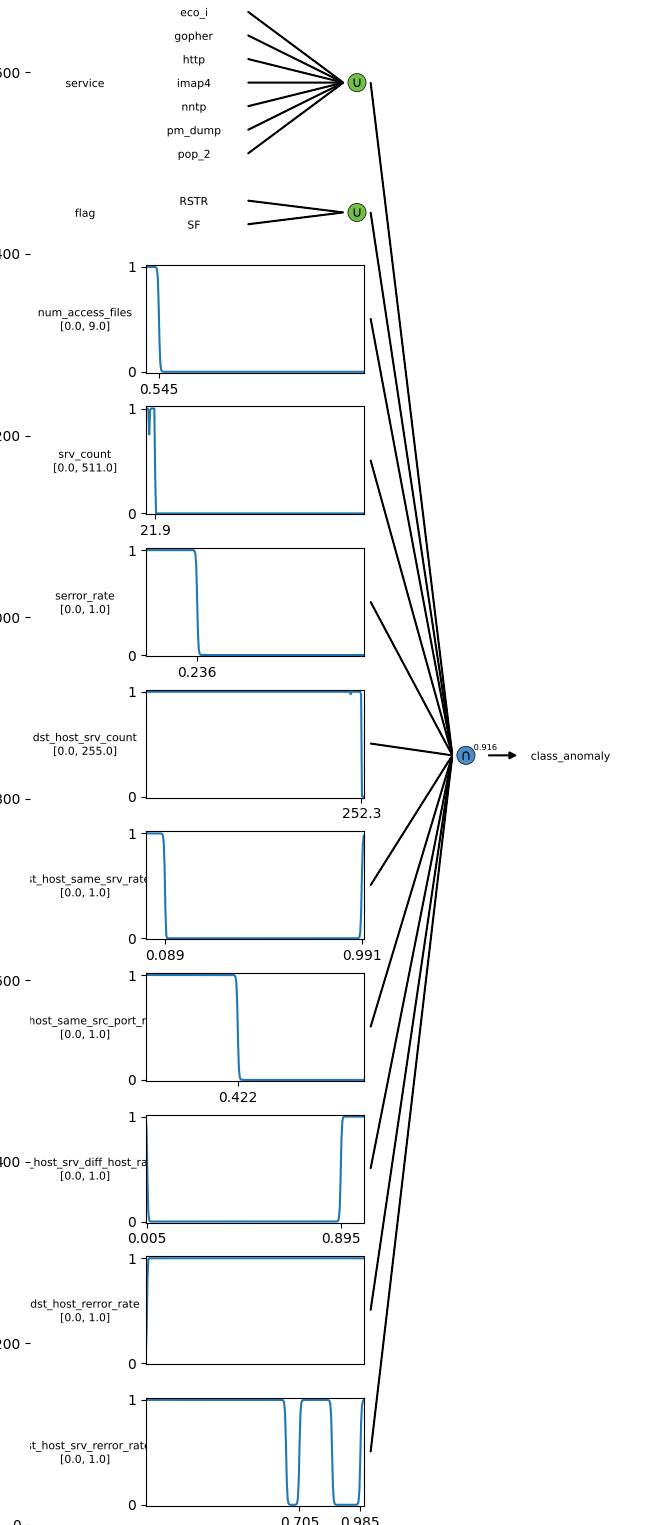}
         \caption{20$^{\text{th}}$ rule (covers 1 \% of anomalies)}
         \label{fig:interp_OR1}
     \end{subfigure}
     }
\end{figure}

% \newpage
\begin{figure}[p]
    \centering
    \parbox[c]{0.475\textwidth}{
     \begin{subfigure}[b]{0.475\textwidth}
         \centering
         \includegraphics[width=\textwidth]{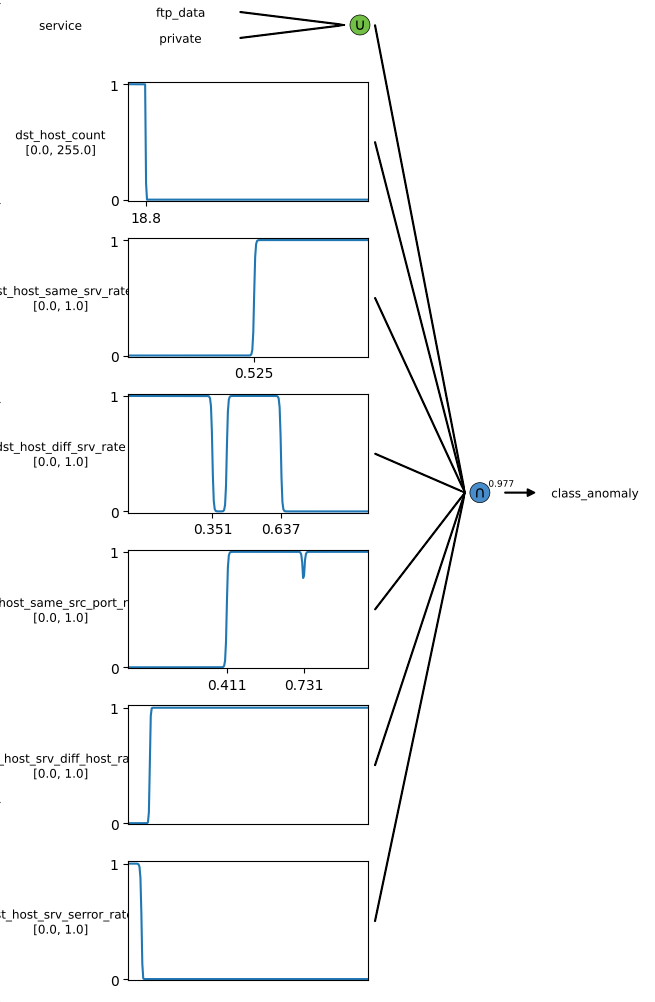}
         \caption{19$^{\text{th}}$ rule (covers 1 \% of anomalies)}
         \label{fig:interp_OR1}
     \end{subfigure}

     \vspace{12pt}
     
     \begin{subfigure}[b]{0.475\textwidth}
         \centering
         \includegraphics[width=\textwidth]{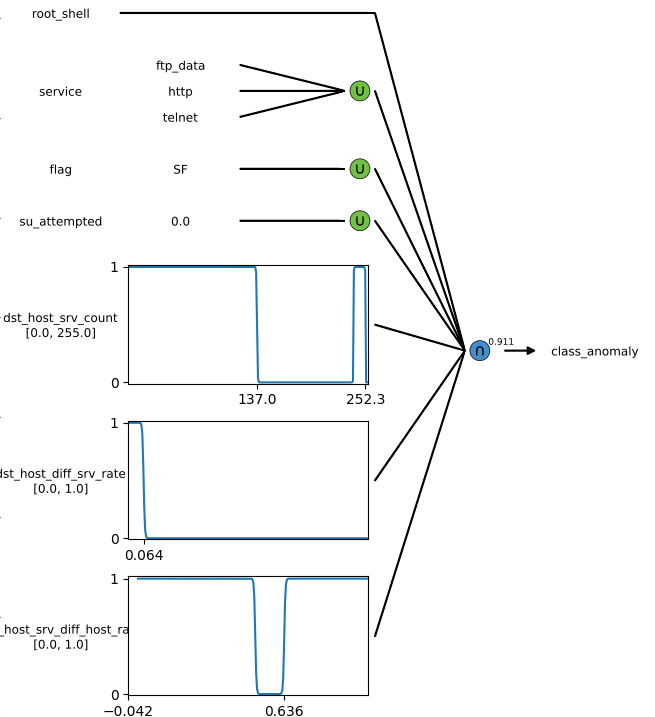}
         \caption{21$^{\text{st}}$ rule (covers <1 \% of anomalies)}
         \label{fig:interp_OR1}
     \end{subfigure}
     }
     \hfill
     \parbox[c]{0.475\textwidth}{
     \begin{subfigure}[b]{0.475\textwidth}
         \centering
         \includegraphics[width=\textwidth]{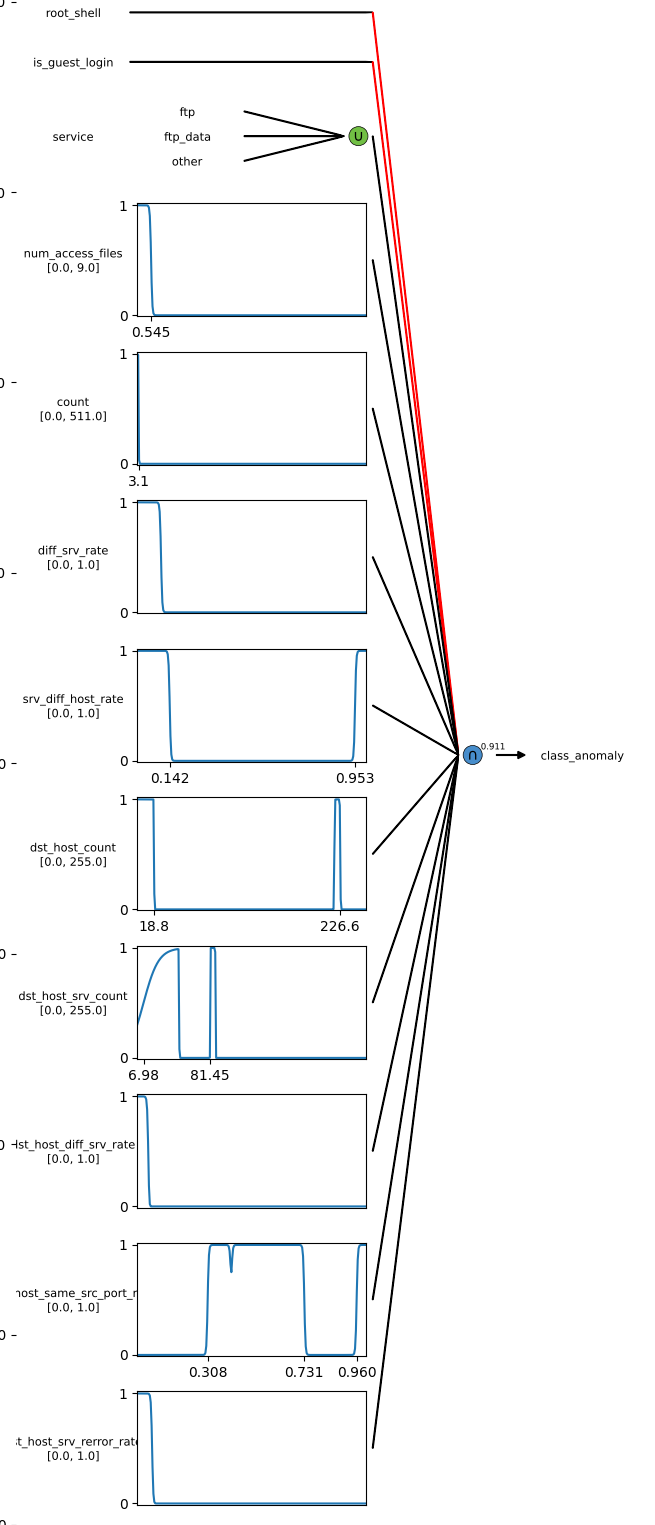}
         \caption{22$^{\text{nd}}$ rule (covers 1 \% of anomalies)}
         \label{fig:interp_OR1}
     \end{subfigure}
     }
\end{figure}

% \newpage
\begin{figure}[p]
    \centering
    \parbox[c]{0.475\textwidth}{
     \begin{subfigure}[b]{0.475\textwidth}
         \centering
         \includegraphics[width=\textwidth]{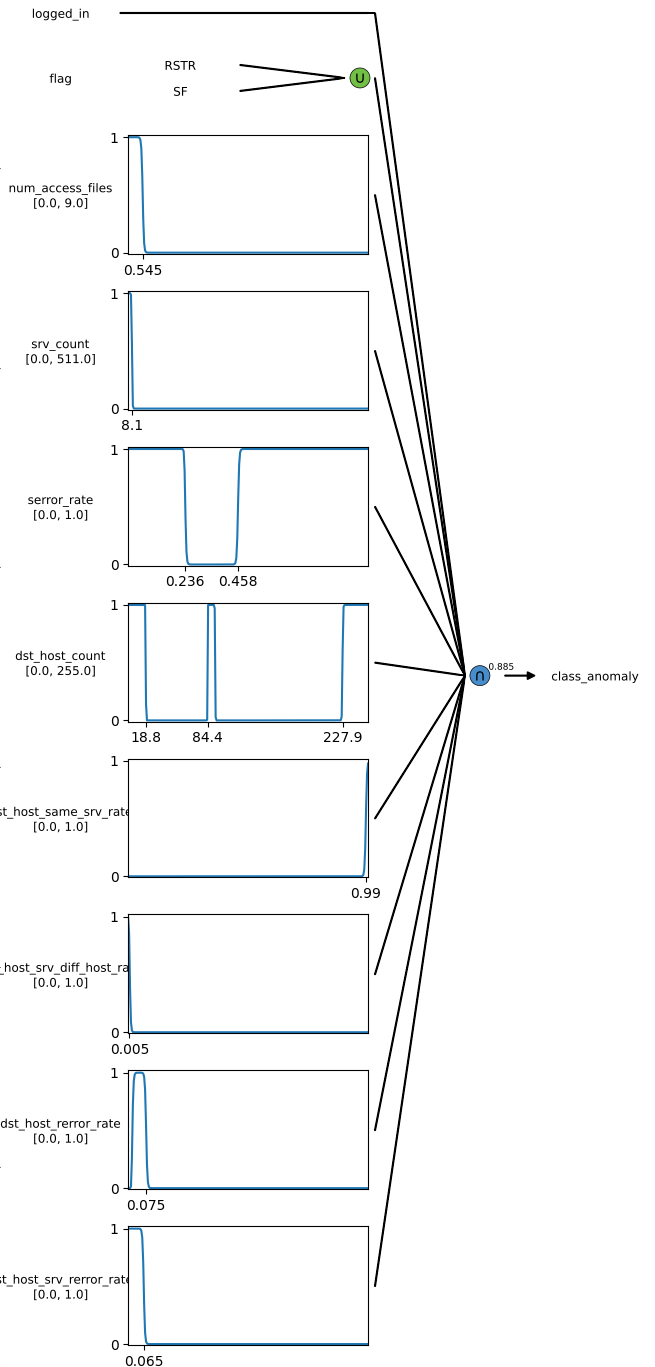}
         \caption{23$^{\text{rd}}$ rule (covers 1 \% of anomalies)}
         \label{fig:interp_OR1}
     \end{subfigure}
     }
     \hfill
     \parbox[c]{0.475\textwidth}{
     \begin{subfigure}[b]{0.475\textwidth}
         \centering
         \includegraphics[width=\textwidth]{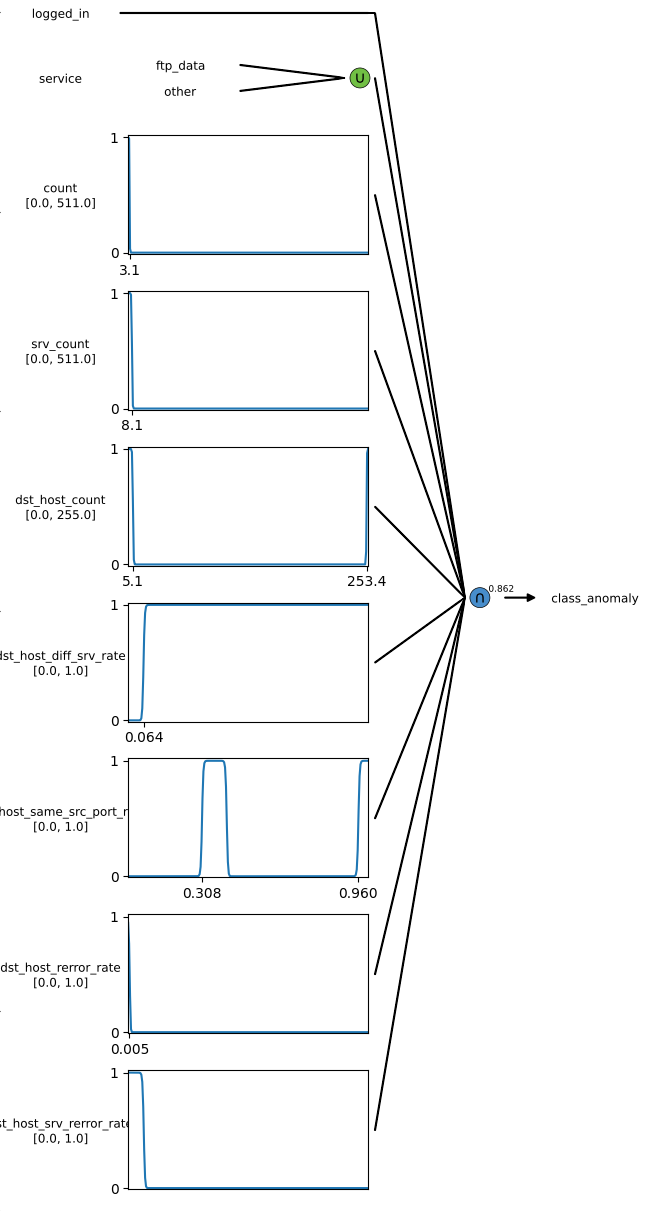}
         \caption{24$^{\text{th}}$ rule (covers <1 \% of anomalies)}
         \label{fig:interp_OR1}
     \end{subfigure}
     }
\end{figure}

\newpage
\subsection{Ablations and statistical validation}
\label{sec:app_exp_abl}

We performed an ablation study on our discretization heuristics and our rule reset mechanism. To do so, we considered the five datasets which were our NLNs performed best (tic-tac-toe, chess, monk2, kidney and wine). In order to take into account the stochastic nature of our algorithm, we performed 5-fold cross-validation (5F-CV) for each configuration with 10 different random seeds. The table below presents the averages of the 5F-CV f1 scores and the standard deviations of those 5F-CV f1 scores. In other words, the standard deviations in the table below reflect the impact of the seed on the 5F-CV f1 score.

\begin{table}[h!]
    \centering
    \caption{Ablation study of discretization heuristic and rule reset mechanism}
    \label{tab:res_bool_net}
  {\scalebox{0.85}{\begin{tabular}{llrclrclrclrclrclrcl}
\toprule
 & & \multicolumn{3}{c}{Tic-tac-toe} & \multicolumn{3}{c}{Chess} & \multicolumn{3}{c}{Monk2} & \multicolumn{3}{c}{Kidney} & \multicolumn{3}{c}{Wine} & \multicolumn{3}{c}{Average}\\\cmidrule(r){3-5} \cmidrule(r){6-8} \cmidrule(r){9-11} \cmidrule(r){12-14} \cmidrule(r){15-17} \cmidrule(r){18-20}
Reset & Discr. & f1 & $\!\!\!\!\!\!\pm\!\!\!\!\!\!$ & $\Delta$ (\%) & f1 & $\!\!\!\!\!\!\pm\!\!\!\!\!\!$ & $\Delta$ (\%) & f1 & $\!\!\!\!\!\!\pm\!\!\!\!\!\!$ & $\Delta$ (\%) & f1 & $\!\!\!\!\!\!\pm\!\!\!\!\!\!$ & $\Delta$ (\%) & f1 & $\!\!\!\!\!\!\pm\!\!\!\!\!\!$ & $\Delta$ (\%) & f1 & $\!\!\!\!\!\!\pm\!\!\!\!\!\!$ & $\Delta$ (\%) \\
\midrule
\multirow{ 6}{*}{\textbf{Yes}} & \textit{None} & \textit{99.74} & $\!\!\!\!\!\!\pm\!\!\!\!\!\!$ & \textit{0.30} & \textit{99.66} & $\!\!\!\!\!\!\pm\!\!\!\!\!\!$ & \textit{0.07}    & \textit{79.09} & $\!\!\!\!\!\!\pm\!\!\!\!\!\!$ & \textit{4.05} & \textit{99.15} & $\!\!\!\!\!\!\pm\!\!\!\!\!\!$ & \textit{0.24} & \textit{95.11} & $\!\!\!\!\!\!\pm\!\!\!\!\!\!$ & \textit{1.00} & \textit{94.55} & $\!\!\!\!\!\!\pm\!\!\!\!\!\!$ & \textit{1.13}\\
\cmidrule{2-20}
 & Thresh. & 99.95 & $\!\!\!\!\!\!\pm\!\!\!\!\!\!$ & 0.16 & 99.40 & $\!\!\!\!\!\!\pm\!\!\!\!\!\!$ & 0.26    & 78.68 & $\!\!\!\!\!\!\pm\!\!\!\!\!\!$ & 4.47 & 98.39 & $\!\!\!\!\!\!\pm\!\!\!\!\!\!$ & 0.67 & 94.05 & $\!\!\!\!\!\!\pm\!\!\!\!\!\!$ & 1.91 & 94.09 & $\!\!\!\!\!\!\pm\!\!\!\!\!\!$ & 1.49\\
\cmidrule{2-20}
 & \textbf{Sel. Desc.} & \textbf{100} & $\!\!\!\!\!\!\pm\!\!\!\!\!\!$ & \textbf{0} & \textbf{99.53} & $\!\!\!\!\!\!\pm\!\!\!\!\!\!$ & 0.09    & 81.27 & $\!\!\!\!\!\!\pm\!\!\!\!\!\!$ & 2.61 & 98.60 & $\!\!\!\!\!\!\pm\!\!\!\!\!\!$ & 0.54 & 94.85 & $\!\!\!\!\!\!\pm\!\!\!\!\!\!$ & 1.59 & 94.85 & $\!\!\!\!\!\!\pm\!\!\!\!\!\!$ & 0.96\\
 & Sel Asc. & \textbf{100} & $\!\!\!\!\!\!\pm\!\!\!\!\!\!$ & \textbf{0} & 99.49 & $\!\!\!\!\!\!\pm\!\!\!\!\!\!$ & \textbf{0.08}    & 81.67 & $\!\!\!\!\!\!\pm\!\!\!\!\!\!$ & 3.38 & \textbf{98.68} & $\!\!\!\!\!\!\pm\!\!\!\!\!\!$ & 0.68 & 94.62 & $\!\!\!\!\!\!\pm\!\!\!\!\!\!$ & 1.56 & 94.89 & $\!\!\!\!\!\!\pm\!\!\!\!\!\!$ & 1.14\\
 & Add. & \textbf{100} & $\!\!\!\!\!\!\pm\!\!\!\!\!\!$ & \textbf{0} & 99.47 & $\!\!\!\!\!\!\pm\!\!\!\!\!\!$ & 0.09    & \textbf{85.91} & $\!\!\!\!\!\!\pm\!\!\!\!\!\!$ & \textbf{2.53} & 98.37 & $\!\!\!\!\!\!\pm\!\!\!\!\!\!$ & 0.58 & \textbf{95.08} & $\!\!\!\!\!\!\pm\!\!\!\!\!\!$ & \textbf{1.18} & \textbf{95.77} & $\!\!\!\!\!\!\pm\!\!\!\!\!\!$ & \textbf{0.87}\\
 & Sub. & \textbf{100} & $\!\!\!\!\!\!\pm\!\!\!\!\!\!$ & \textbf{0} & 99.51 & $\!\!\!\!\!\!\pm\!\!\!\!\!\!$ & \textbf{0.08}    & 80.69 & $\!\!\!\!\!\!\pm\!\!\!\!\!\!$ & 3.00 & 98.51 & $\!\!\!\!\!\!\pm\!\!\!\!\!\!$ & \textbf{0.42} & 94.52 & $\!\!\!\!\!\!\pm\!\!\!\!\!\!$ & 1.32 & 94.65 & $\!\!\!\!\!\!\pm\!\!\!\!\!\!$ & 0.96\\
\midrule
\multirow{ 2}{*}{No} & \textit{None} & \textit{99.73} & $\!\!\!\!\!\!\pm\!\!\!\!\!\!$ & \textit{0.31} & \textit{99.65} & $\!\!\!\!\!\!\pm\!\!\!\!\!\!$ & \textit{0.08}    & \textit{78.73} & $\!\!\!\!\!\!\pm\!\!\!\!\!\!$ & \textit{4.84} & \textit{99.13} & $\!\!\!\!\!\!\pm\!\!\!\!\!\!$ & \textit{0.24} & \textit{95.11} & $\!\!\!\!\!\!\pm\!\!\!\!\!\!$ & \textit{1.00} & \textit{94.47} & $\!\!\!\!\!\!\pm\!\!\!\!\!\!$ & \textit{1.29}\\
\cmidrule{2-20}
 & Sel. Desc. & \textbf{100} & $\!\!\!\!\!\!\pm\!\!\!\!\!\!$ & \textbf{0} & 99.51 & $\!\!\!\!\!\!\pm\!\!\!\!\!\!$ & \textbf{0.08}    & 81.15 & $\!\!\!\!\!\!\pm\!\!\!\!\!\!$ & 3.46 & 98.34 & $\!\!\!\!\!\!\pm\!\!\!\!\!\!$ & 0.51 & 94.84 & $\!\!\!\!\!\!\pm\!\!\!\!\!\!$ & 1.49 & 94.77 & $\!\!\!\!\!\!\pm\!\!\!\!\!\!$ & 1.11\\
\bottomrule
\end{tabular}}}
\end{table}

All of our discretization heuristics have a similar performance profile, beating the instantaneous Threshold heuristic in all datasets and gaining on average between 0.5 \% and 1.5 \%. The biggest gain comes from the Additive heuristic in the Monk2 dataset with an additional 7 \%. However, these gains come at the cost of a very time-intensive discretization algorithm, especially as the model and/or the dataset grows in size. Whether these gains are worth the additional wait time and compute resources will be at the discretion of its users. The other experimental results in this paper use the Descending Selecion heuristic, which is slightly worse than the Additive heuristic on average. However, this small difference does not seem to be consequential enough to justify redoing all of the other experiments.

Our rule reset mechanism improves the performance on average systematically in all datasets, although this difference in performance is extremely small, being at most 0.25 \%. This puts into question the usefulness of this rule reset mechanism. It is clearly not necessary for a good performance, but it can help marginally at no significant additional cost in computation.

The standard deviations of the 5f-CV f1 scores are often bigger than the difference between configurations. This suggest that the initialization seed is just as if not more consequent than the configuration. However, it is interesting to note that the best configurations on average often also have some of the smallest standard deviations. The most adapted configuration for a dataset could then also be more robust to the initialization seed.

\end{document}